\documentclass[fleqn,format=acmsmall, review=false, screen=true]{acmart}
\usepackage{amsmath}
\usepackage{graphicx}

\usepackage{graphicx,calc}
\newlength\myheight
\newlength\mydepth
\settototalheight\myheight{Xygp}
\newcommand*\inlinegraphics[1]{
\settototalheight\myheight{Xygp}
\settodepth\mydepth{Xygp}
\raisebox{-\mydepth}{\includegraphics[height=\myheight]{#1}}
}

\begin{document}

\title{Structural analysis of Hindi online handwritten characters for character recognition}

\author{Anand Sharma}
\email{anand.sharma@miet.ac.in}
\affiliation{
\institution{MIET}
\city{Meerut}
\country{India}}
\author{A. G. Ramakrishnan}
\email{agr@iisc.ac.in}
\affiliation{
\institution{Indian Institute of Science}
\city{Bangaluru}
\country{India}}

\begin{abstract}
\indent Direction properties of online strokes are used to analyze them in terms of homogeneous regions or sub-strokes with points satisfying common geometric properties. Such sub-strokes are called sub-units. These properties are used to extract sub-units from Hindi ideal online characters. These properties along with some heuristics are used to extract sub-units from Hindi online handwritten characters.\\
\indent A method is developed to extract point stroke, clockwise curve stroke, counter-clockwise curve stroke and loop stroke segments as sub-units from Hindi online handwritten characters. These extracted sub-units are close in structure to the sub-units of the corresponding Hindi online ideal characters.\\
\indent Importance of local representation of online handwritten characters in terms of sub-units is assessed by training a classifier with sub-unit level local and character level global features extracted from characters for character recognition. The classifier has the recognition accuracy of 93.5\% on the testing set. This accuracy is the highest when compared with that of the classifiers trained only with global features extracted from characters in the same training set and evaluated on the same testing set.\\
\indent Sub-unit extraction algorithm and the sub-unit based character classifier are tested on Hindi online handwritten character dataset.  This dataset consists of samples from 96 different characters. There are 12832 and 2821 samples in the training and testing sets, respectively.\\ 

\end{abstract}




\maketitle

\section{Introduction}
\indent Online handwritten character is generated as a sequence of strokes. A stroke is a unit of online handwriting. There are various studies that use stroke level or sub-stroke level features for training classifier for character recognition. Swethalakshmi et al. \cite{troaq} use stroke level features for stroke classification and then use script dependent rules for character recognition. Characters are modeled in terms of sub-strokes called ballistic strokes by Teja et al. \cite{sua} which are extracted based on curvature regions in strokes. Euclidean distance between consecutive points and direction between consecutive points are used by Tokuno et al. \cite{sub}, Nakai et al. \cite{suc}, Shimodaira et al. \cite{sud} and nakai et al. \cite{suf} as sub-stroke level features for training sub-stroke HMM models that are used for character recognition. Sub-strokes that contribute towards the discrimination between online strokes are identified by Alahari et al. \cite{sue} and are used for character and numeral recognition. Concatenated nebulous stroke HMM models are used to build character models for character recognition by Hu et al. \cite{sug}. Study done by Biswas et al. \cite{suh} estimates probability distributions of stroke classes based on the stroke features and build HMM based classifier using each stroke class as state for character recognition. Parui et al. \cite{sui} use sub-stroke level features to train HMM model for stroke classification that are then used for character classification. Fundamental character structure components called radicals which correspond to character strokes or sub-strokes are used to model character for character recognition in the studies done by Ma et al.\cite{suj}, Ma et al. \cite{suk} and Zhang et al. \cite{sul}. Stroke is represented as a string of shape features by Shankar et al. \cite{sum} to identify an unknown stroke. Study done  by Bercu et al. \cite{sun} classifies successive sub-strokes in a character as loops, humps and cusps and uses these as features to build HMM based recognition system.  \\
\indent This study uses direction properties of an online stroke in an ideal character to analyze it by identifying homogeneous sub-structures or sub-strokes called sub-units in the stroke. A method is developed for extraction of sub-units from strokes in samples of Hindi online handwritten characters. The extracted sub-units are close in structure to the sub-units of the corresponding ideal character. A classifier using the sub-unit level local and character level global representation of a character is used for character recognition to assess the importance of local representation of characters in terms of sub-units.\\
\indent This paper is organized into seven Sections. Section 2 describes Hindi character set and the corresponding character dataset. Structure of online handwritten characters is described in Section 3. Sub-unit extraction methods for the online ideal characters and the corresponding online handwritten characters are developed in sections 4 and 5, repectively. Experiments and results are given in Section 6. Section 7 summarizes the contribution of this work. 

\section{Hindi characters}
\indent Hindi language is written using Devanagari script \cite{troiscii} and so Hindi characters are subset of Devanagari characters. Online handwritten samples of some of the the characters used for writing Hindi language have been collected for assessing the sub-unit extraction algorithm and to check the importance of local representation of a character in terms of sub-units towards character recognition. 
\subsection{Hindi character set}
\indent Hindi characters considered for sub-unit extraction and character classification are vowels, consonants, pure consonants, nasalization sign, vowel omission sign, vowel signs, consonant with vowel sign, and conjuncts. Strokes with different shapes are also considered in addition to these characters. Figure 1 shows subset of Hindi vowels considered for analysis. The other vowels can be obtained by a combination of these vowels and one or more vowel signs. For example, the vowel {\inlinegraphics{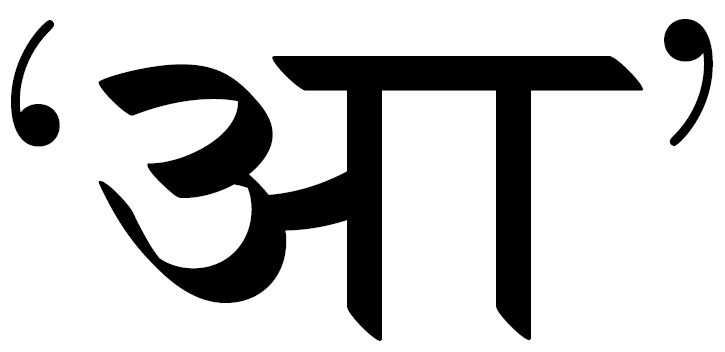}} can be obtained as a combination of the symbol 1 in Fig. 1 and the symbol 67 in Fig. 4.
\begin{figure}[ht]
\begin{center}
$\begin{array}{cccccc}
\includegraphics[width=0.4 in]{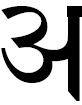}&
\includegraphics[width=0.4 in]{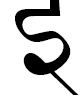}&
\includegraphics[width=0.4 in]{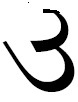}&
\includegraphics[width=0.5 in]{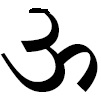}&
\includegraphics[width=0.6 in]{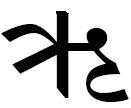}&
\includegraphics[width=0.35 in]{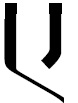}\\
1&2&3&4&5&6\\
\end{array}$
\caption{Basic Hindi vowels used as recognition primitives.}
\end{center}
\end{figure}
All the consonants have an implicit vowel `a' \cite{troiscii}. Figure 2  shows the subset of Hindi consonants considered.
\begin{figure}[ht]
\begin{center}
$\begin{array}{cccccccccc}
\includegraphics[width=0.45 in]{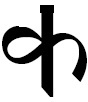}&
\includegraphics[width=0.45 in]{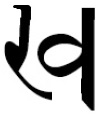}&
\includegraphics[width=0.4 in]{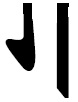}&
\includegraphics[width=0.4 in]{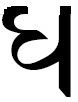}&
\includegraphics[width=0.4 in]{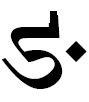}&
\includegraphics[width=0.4 in]{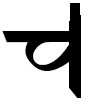}&
\includegraphics[width=0.4 in]{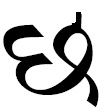}&
\includegraphics[width=0.4 in]{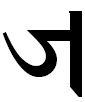}&
\includegraphics[width=0.4 in]{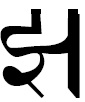}&
\includegraphics[width=0.4 in]{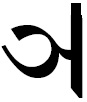}\\
7&8&9&10&11&12&13&14&15&16\\
\includegraphics[width=0.4 in]{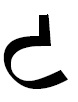}&
\includegraphics[width=0.4 in]{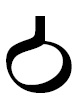}&
\includegraphics[width=0.4 in]{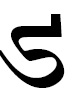}&
\includegraphics[width=0.4 in]{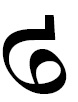}&
\includegraphics[width=0.4 in]{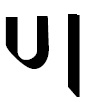}&
\includegraphics[width=0.4 in]{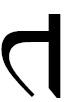}&
\includegraphics[width=0.4 in]{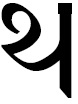}&
\includegraphics[width=0.4 in]{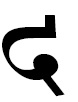}&
\includegraphics[width=0.4 in]{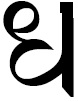}&
\includegraphics[width=0.4 in]{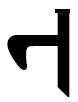}\\
17&18&19&20&21&22&23&24&25&26\\
\includegraphics[width=0.4 in]{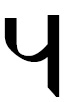}&
\includegraphics[width=0.4 in]{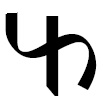}&
\includegraphics[width=0.4 in]{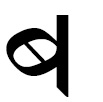}&
\includegraphics[width=0.4 in]{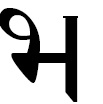}&
\includegraphics[width=0.4 in]{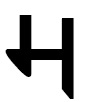}&
\includegraphics[width=0.4 in]{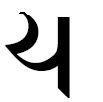}&
\includegraphics[width=0.4 in]{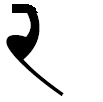}&
\includegraphics[width=0.4 in]{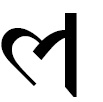}&
\includegraphics[width=0.4 in]{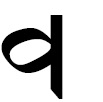}&
\includegraphics[width=0.4 in]{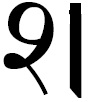}\\
27&28&29&30&31&32&33&34&35&36\\
\includegraphics[width=0.4 in]{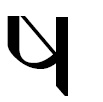}&
\includegraphics[width=0.4 in]{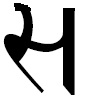}&
\includegraphics[width=0.4 in]{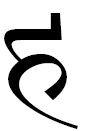}\\
37&38&39&&&&&&&\\
\end{array}$
\caption{Hindi consonants included as part of the recognized classes.}
\end{center}
\end{figure}
Pure consonants or half consonants are the corresponding consonants with their implicit vowel muted. Figure 3 shows subset of Hindi pure consonants considered.
\begin{figure}[ht]
\begin{center}
$\begin{array}{cccccccccc}
\includegraphics[width=0.45 in]{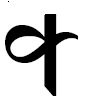}&
\includegraphics[width=0.45 in]{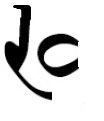}&
\includegraphics[width=0.4 in]{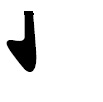}&
\includegraphics[width=0.4 in]{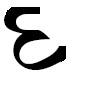}&
\includegraphics[width=0.4 in]{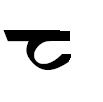}&
\includegraphics[width=0.4 in]{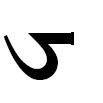}&
\includegraphics[width=0.4 in]{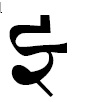}&
\includegraphics[width=0.4 in]{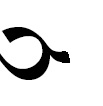}&
\includegraphics[width=0.4 in]{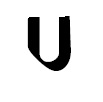}&
\includegraphics[width=0.4 in]{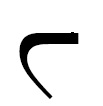}\\
40&41&42&43&44&45&46&47&48&49\\
\includegraphics[width=0.4 in]{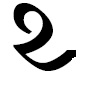}&
\includegraphics[width=0.4 in]{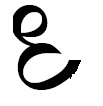}&
\includegraphics[width=0.4 in]{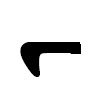}&
\includegraphics[width=0.4 in]{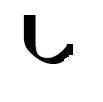}&
\includegraphics[width=0.4 in]{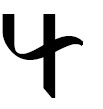}&
\includegraphics[width=0.4 in]{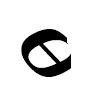}&
\includegraphics[width=0.4 in]{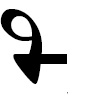}&
\includegraphics[width=0.4 in]{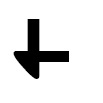}&
\includegraphics[width=0.4 in]{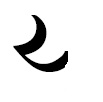}&
\includegraphics[width=0.4 in]{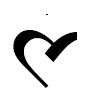}\\
50&51&52&53&54&55&56&57&58&59\\
\includegraphics[width=0.4 in]{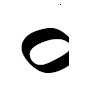}&
\includegraphics[width=0.4 in]{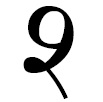}&
\includegraphics[width=0.4 in]{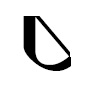}&
\includegraphics[width=0.4 in]{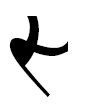}&
\includegraphics[width=0.4 in]{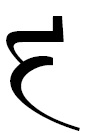}\\
60&61&62&63&64&&&&&\\
\end{array}$
\caption{Hindi half consonants considered for recognition.}
\end{center}
\end{figure}

The nasalization sign indicates nasalization of the character the sign is written over. The vowel omission sign is used to mute the implicit vowel of a consonant. The vowel sign is used to modify the implicit vowel of a consonant. Figure 4(65) shows the nasalization sign, Figure 4(66) shows the vowel omission sign, Figures 4(67)-(72) show vowel signs and Figure 4(73) shows a consonant modified by a vowel sign.
\begin{figure}[ht!]
\begin{center}
$\begin{array}{ccccccccc}
\includegraphics[width=0.45 in]{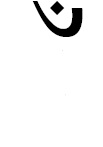}&
\includegraphics[width=0.45 in]{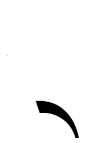}&
\includegraphics[width=0.4 in]{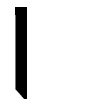}&
\includegraphics[width=0.4 in]{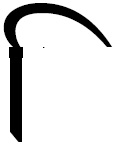}&
\includegraphics[width=0.4 in]{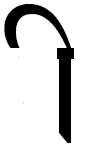}&
\includegraphics[width=0.4 in]{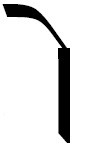}&
\includegraphics[width=0.4 in]{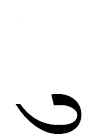}&
\includegraphics[width=0.4 in]{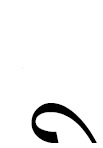}&
\includegraphics[width=0.4 in]{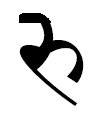}\\
65&66&67&68&69&70&71&72&73\\
\end{array}$
\caption{Some signs and consonant with vowel sign considered for recognition. (65) Nasalization sign. (66) Vowel omission sign. (67)-(72) Vowel signs. (73) Consonant with vowel sign.}
\end{center}
\end{figure}
Conjuncts are cluster of consonants where implicit vowels of all but the last consonant of the cluster are muted. Some conjuncts considered are shown in Figure 5.
\begin{figure}[ht!]
\begin{center}
$\begin{array}{cccccccccc}
\includegraphics[width=0.4 in]{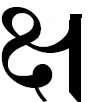}&
\includegraphics[width=0.4 in]{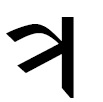}&
\includegraphics[width=0.4 in]{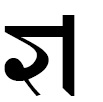}&
\includegraphics[width=0.4 in]{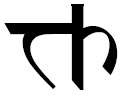}&
\includegraphics[width=0.4 in]{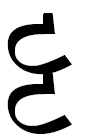}&
\includegraphics[width=0.4 in]{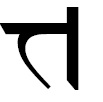}&
\includegraphics[width=0.4 in]{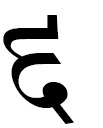}&
\includegraphics[width=0.4 in]{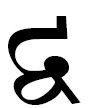}&
\includegraphics[width=0.4 in]{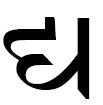}&
\includegraphics[width=0.4 in]{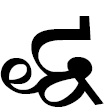}\\
74&75&76&77&78&79&80&81&82&83\\
\includegraphics[width=0.4 in]{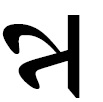}&
\includegraphics[width=0.4 in]{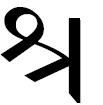}&
\includegraphics[width=0.4 in]{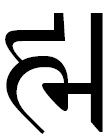}&
\includegraphics[width=0.4 in]{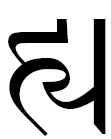}&
\includegraphics[width=0.4 in]{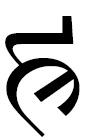}&
\includegraphics[width=0.4 in]{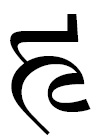}&
\includegraphics[width=0.4 in]{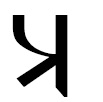}&
\includegraphics[width=0.4 in]{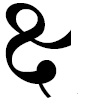}&
\includegraphics[width=0.4 in]{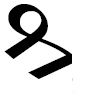}&
\includegraphics[width=0.4 in]{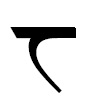}\\
84&85&86&87&88&89&90&91&92&93\\
\end{array}$
\caption{Conjuncts and consonant clusters included as part of recognized classes. Conjuncts are clusters of consonants, where only the implicit vowel of the last consonant is not muted.(74)-(90) Frequently used conjuncts. Symbols, where the implicit vowel of the last consonant of a consonant cluster is muted, are also used. Symbols 91 to 93 belong to this category.}
\label{conjunc}
\end{center}
\end{figure}
Some strokes with different shapes are shown in Figure 6.  
\begin{figure}[ht]
\begin{center}
$\begin{array}{ccc}
\includegraphics[width=0.45 in]{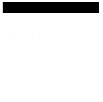}&
\includegraphics[width=0.45 in]{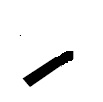}&
\includegraphics[width=0.4 in]{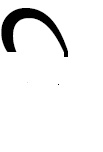}\\
94&95&96\\
\end{array}$
\caption{Other strokes with different shapes considered for recognition.}
\end{center}
\end{figure}
\subsection{Hindi character dataset}
\indent Samples of Hindi online handwritten characters have been collected from Center for Development of Advanced Computation (CDAC) and HP datasets \cite{trohp}. Characters that are not cursive and have been written correctly are collected. Figures 7(a) and 7(b) show samples of Hindi online handwritten characters obtained from the constructed dataset. These characters are produced as a sequence of strokes in a particular order. The strokes are produced in a particular direction. The header lines over all the characters have been removed because they do not contribute to the structure of the characters.\\
\indent The total number of samples of characters in the dataset is 15653. The dataset is divided to form training and testing datasets. The training and testing datasets consist of 12832 and 2821 samples, respectively. These datasets consist of samples from 96 different characters with an average of 133 and 29 samples per character class in the training and testing datasets, respectively.
\begin{figure}[ht]
\begin{center}
$\begin{array}{cc}
\includegraphics[width=2.5 in]{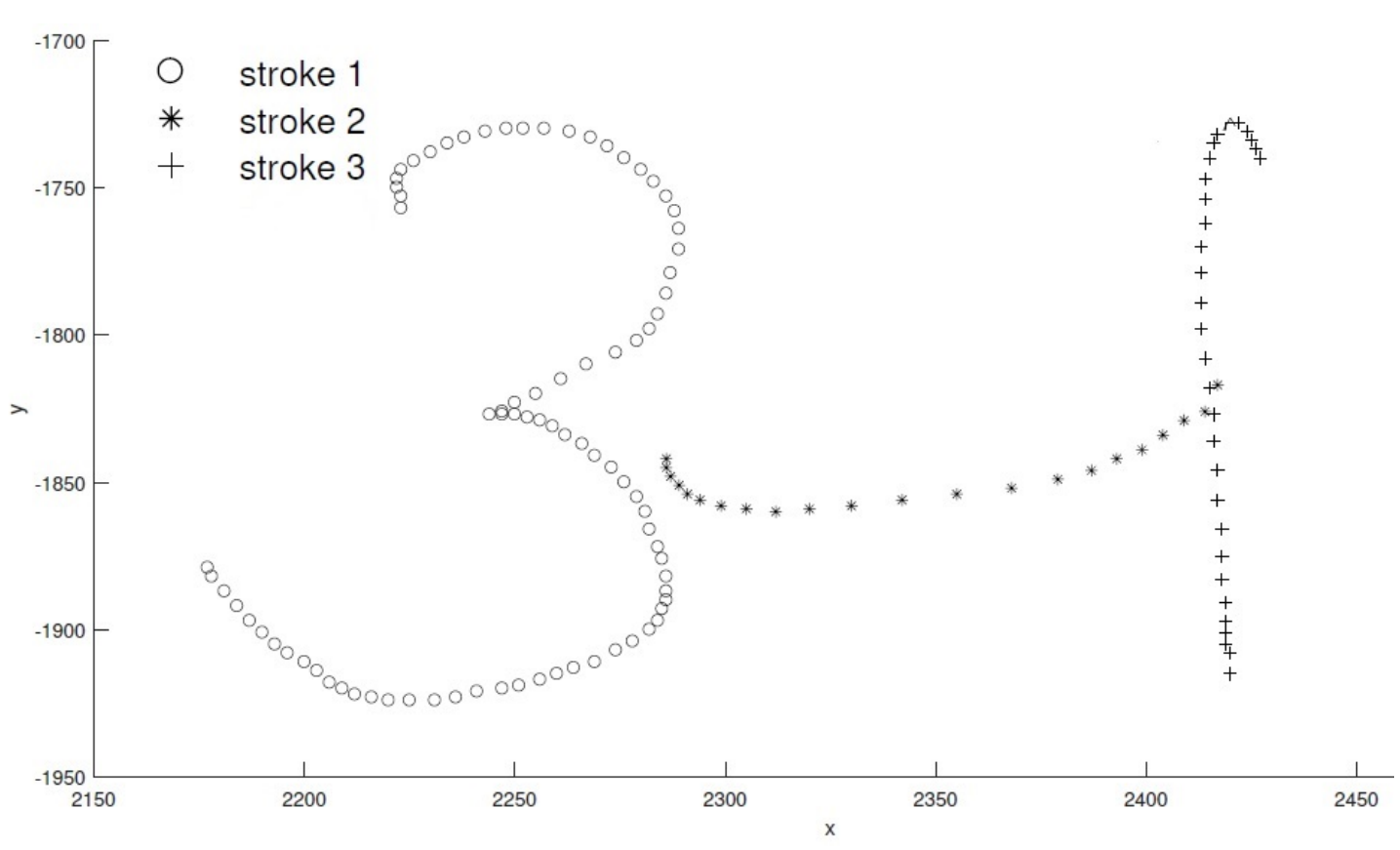}&
\includegraphics[width=2.5 in]{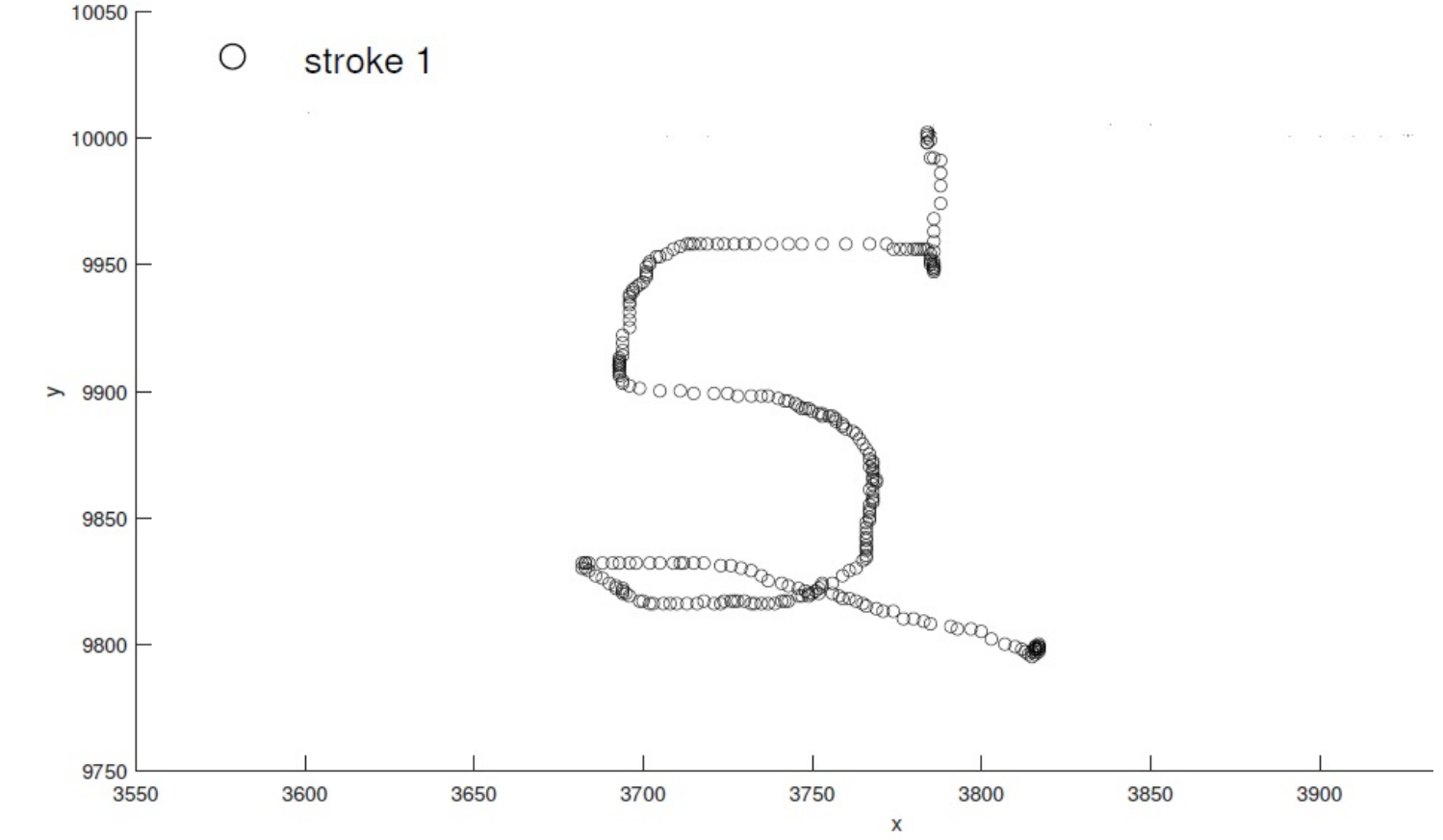}\\

(a)&(b)\\
\end{array}$
\end{center}
\caption{Samples of Hindi online handwritten characters. (a) Sample character produced using three strokes. (b) Sample character produced using one stroke.}
%\caption{Hindi online handwritten character.}
\end{figure}
\subsection{Preprocessing of samples in the dataset}
\indent The samples of online handwritten characters in the constructed datasets have got variations because they have been produced by different individuals at different times. Variations among samples of characters can be classified as external variations and internal variations. External variations are variations that can be removed by preprocessing or feature design. Internal variations are caused by deviation of handwritten character structure from the ideal character structure and are very difficult to remove. It is important to preprocess the  characters so that feature extraction from characters can be done in a uniform fashion.\\
\indent Some of the external variations removed from handwritten characters in the datasets are as follows. Repeated points are consecutive points in a character that have the same co-ordinate values and do not contribute to the structure of the character and so are detected and removed. Co-ordinate values of points in characters in both the $x$ and $y$ directions are mapped to the range $[0\,\,1]$ by linear transformation. This removes variations in location and size of characters. Distance between all the consecutive points in characters are made equal to a constant $\Delta$ thus removing variations in speed at which handwritten characters have been produced. Character stroke trace roughness is removed by linear filtering the sequences of x-co-ordinates and y-co-ordinates of points in the characters.\\
\indent After preprocessing, an online handwritten character is represented as a sequence of strokes, where a stroke is represented as a sequence of $x$-co-ordinates and $y$-co-ordinates of points in the stroke. Figure 8 shows samples of online handwritten characters before and after preprocessing.  
\begin{figure}[ht]
\begin{center}
$\begin{array}{cc}
\includegraphics[width=2.5 in]{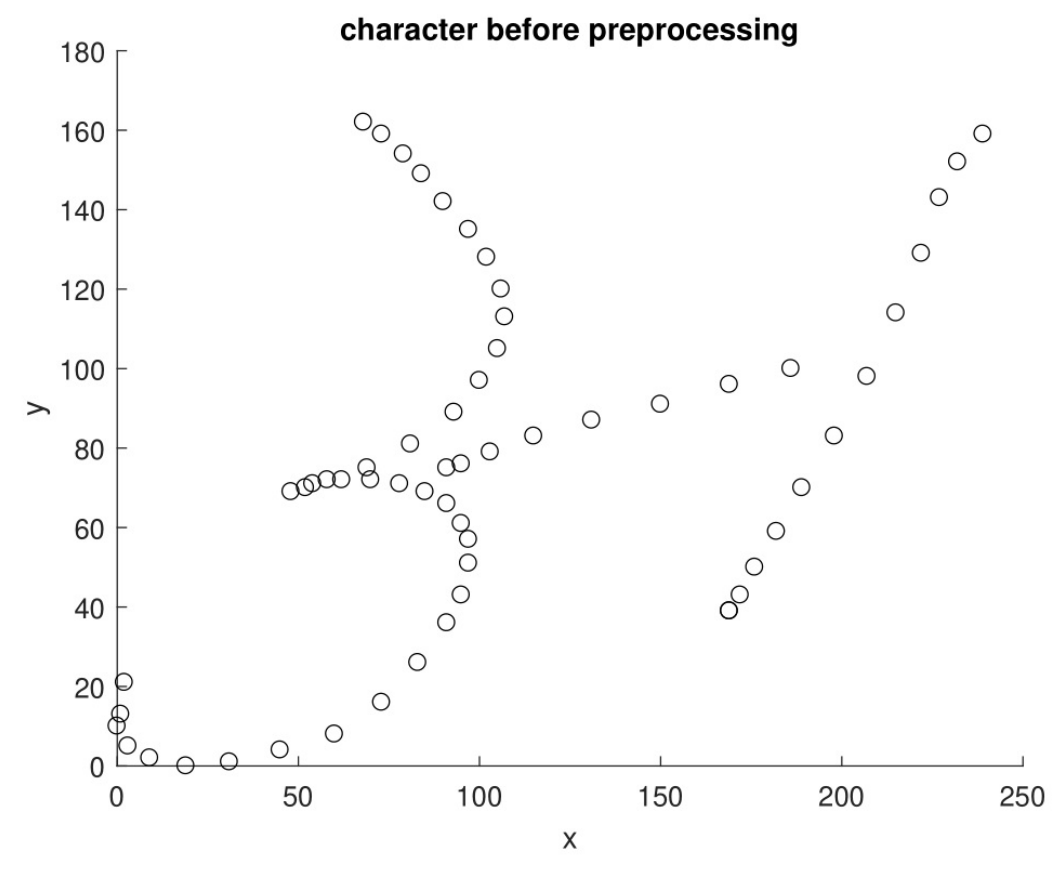}&
\includegraphics[width=2.5 in]{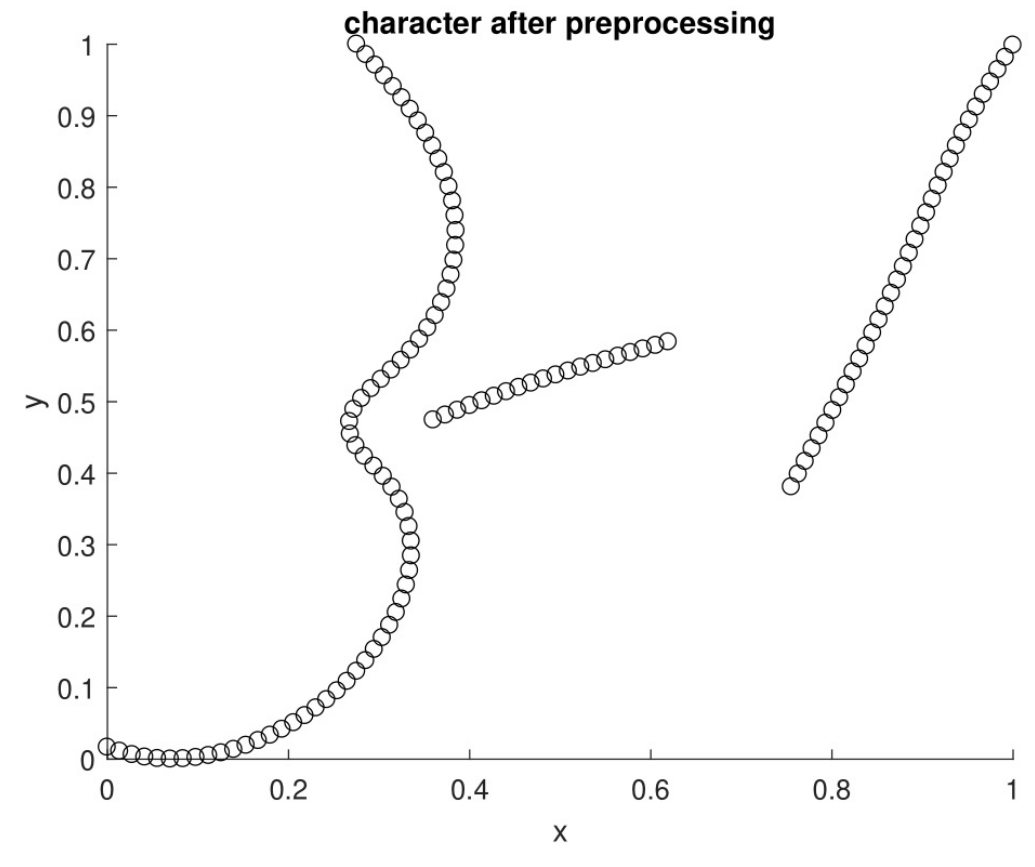}\\
(a)&(b)\\
\includegraphics[width=2.5 in]{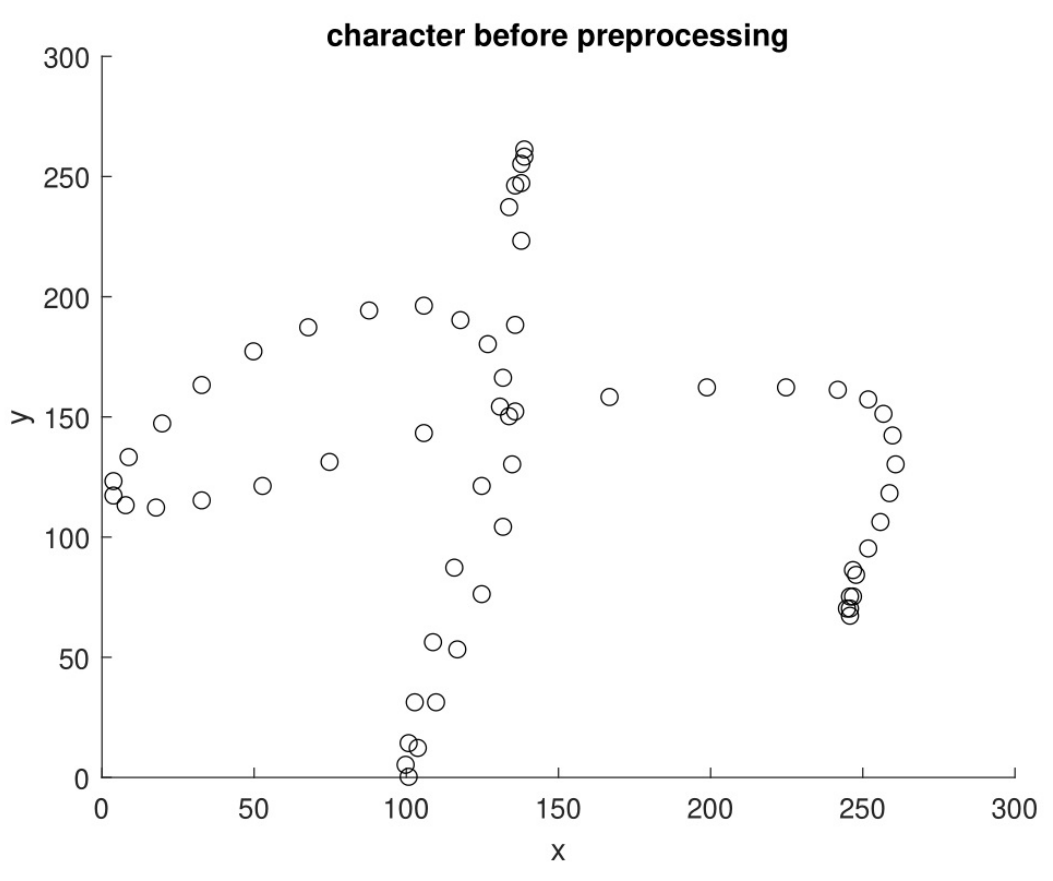}&
\includegraphics[width=2.5 in]{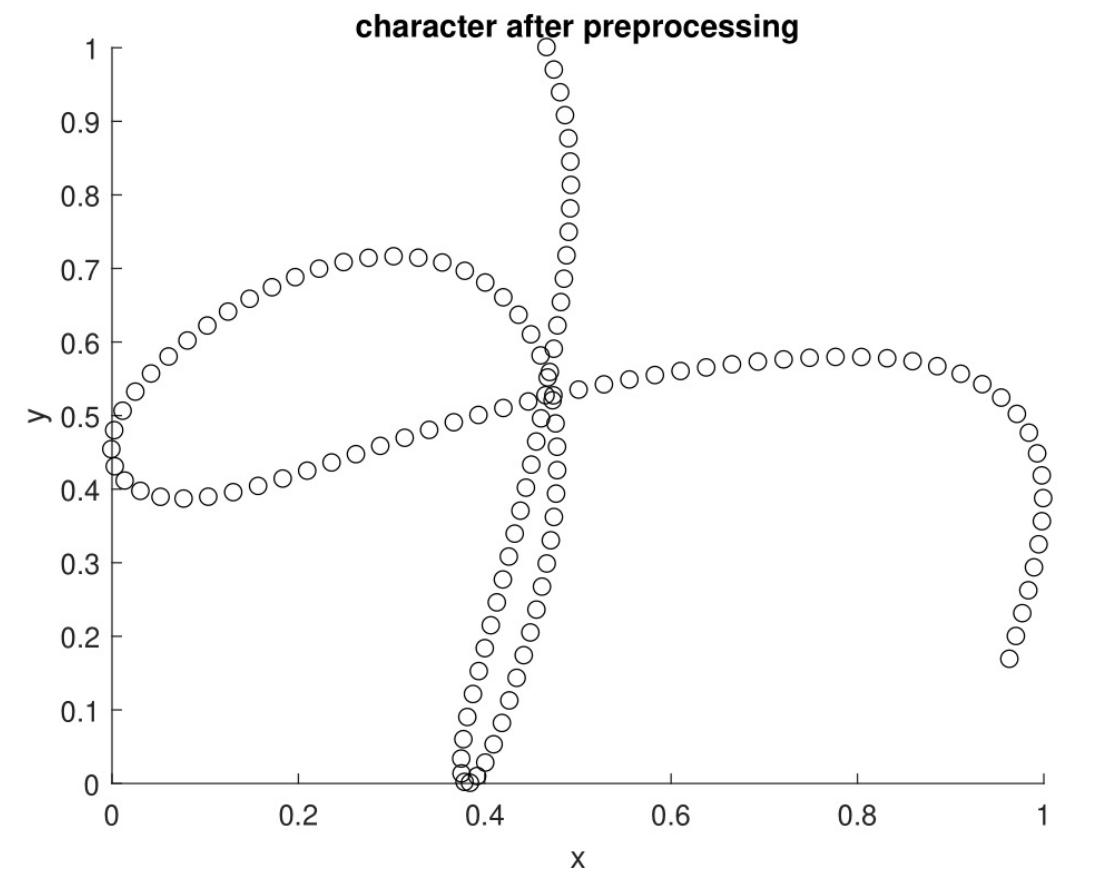}\\
(c)&(d)\\
\end{array}$
\caption{Characters before and after preprocessing. (a) and (c) Two characters before preprocessing. (b) and (d) The same characters, after application of preprocessing steps like removal of repeated points, variations in location and size, variation in distance between consecutive points, and roughness of trace of the characters.}
\end{center}
\end{figure}

%\section{Sub-unit}  
\section{Structure of online handwritten character}
\subsection{\bfseries{Online stroke generation}}
An online handwritten character is generated as a sequence of strokes produced by movement of pen on a touch sensitive screen. Stroke is a unit of online handwriting. Structure of online handwritten character can be understood in terms of structure of strokes. Stroke is a sequence of points produced between pen-down state and pen-up state and is represented as
\[S_i=[{p_1^{S_i}}^T;\dots ;{p_n^{S_i}}^T;\dots ;{p_{N_{S_i}}^{S_i}}^T],\,\,\, S_i\in\ [0\,\, 1]^{N_{S_i}\times2},\,\,\, p_n^{S_i}=[x_n\,\, y_n]^T,\,\,\, p_n^{S_i}\in\ [0\,\, 1]^{2\times1}.\] $N_{S_i}$ is the number of points in the stroke $S_i$. Point $p_n^{S_i}$ is the $n^{th}$ point in the stroke $S_i$, $x_n$ and $y_n$ are, respectively, the $x$-co-ordinate and $y$-co-ordinate of the $n^{th}$ point. Direction of the stroke $S_i$ at the point $p_n^{S_i}$ is given by \[v_n^{S_i}=\frac{(p_{n+1}^{S_i}-p_{n}^{S_i})}{ (||p_{n+1}^{S_i}-p_{n}^{S_i}||_2)},\quad v_n^{S_i}\in\mathcal{R}^{2\times 1},\quad ||v_n^{S_i}||_2=1,\quad1\le n\le N_{S_i}-1.\] Change in direction of  the stroke at the point $p_n^{S_i}$ is given by \[\theta_n^{S_i}=\cos^{-1}\big({v_{n-1}^{S_i}}^T\, v_{n}^{S_i} ),\quad 0\le\theta_n^{S_i}\le 180^o,\quad 2\le n\le N_{S_i}-1.\] If the distance between the first and the last point in the stroke $S_i$ is equal to the distance, $||p_{N_{S_i}}^{S_i}-p_1^{S_i}||_2=\Delta$, then direction of the stroke at the point $p_{N_{S_i}}^{S_i}$ towards the first point $p_1^{S_i}$ is $v_{N_{S_i}}^{S_i}=(p_1^{S_i}-p_{N_{S_i}}^{S_i})\, (||p_1^{S_i}-p_{N_{S_i}}^{S_i}||_2)^{-1}$. Changes in directions of the stroke at the points $p_{N_{S_i}}^{S_i}$ and $p_1^{S_i}$ are, respectively, $\theta_{N_{S_i}}^{S_i}=\cos^{-1}\big({v_{N_{S_i}-1}^{S_i}}^T\, v_{N_{S_i}}^{S_i} \big)$ and $\theta_1^{S_i}=\cos^{-1}\big({v_{N_{S_i}}^{S_i}}^T\, v_1^{S_i} \big)$. \\ 
\indent Stroke time span is the time elapsed between pen-down and pen-up states during generation of a stroke. Different strokes have different time spans. When time span of a stroke is so small that only at most two points are produced in the stroke then such a stroke is called a point stroke and is represented as $S^{pt}$. When the time span of a stroke is large and more than two points are produced in a stroke then such a stroke is called a line stroke. A smooth line stroke can be produced by pen movement in one of the three possible directions indicated by the function \[f^d(v_1,v_2)=v_1(1)\, v_2(2)-v_1(2)\, v_2(1),\quad \forall v_1,v_2\in\mathcal{R}^{2\times 1},\quad ||v_1||_2=||v_2||_2=1. \]
When a line stroke $S_i$ is produced by pen movement in the clockwise direction then for each point $p_n^{S_i}$,\\
\begin{equation}f^d(v_{n-1}^{S_i},v_{n}^{S_i})<0,\quad  2\le n\le N_{S_i}-1.\end{equation}
Here all the points in the stroke are said to constitute a homogeneous region because they all satisfy property $(1)$. Such a homogeneous region is represented as $S^{cw}$.        
When a line stroke $S_i$ is produced by pen movement in counter-clockwise direction then for each point $p_n^{S_i}$,\\
\begin{equation}f^d(v_{n-1}^{S_i},v_{n}^{S_i})>0,\quad 2\le n\le N_{S_i}-1.\end{equation}
 Here all the points in the stroke are said to constitute a homogeneous region because they all satisfy property $(2)$. Such a homogeneous region is represented as $S^{ccw}$. Here both the line strokes $S^{cw}$ and $S^{ccw}$ are called curve stroke, $S^c\in \{ S^{cw},S^{ccw} \}$.        
When a line stroke $S_i$ is produced by pen movement in a straight direction then for each point $p_n^{S_i}$,\\
\begin{equation}f^d(v_{n-1}^{S_i},v_{n}^{S_i})=0,\quad 2\le n\le N_{S_i}-1.\end{equation}
Here all the points in the stroke are said to constitute a homogeneous region because they all satisfy property $(3)$. Such homogeneous region is represented as $S^{st}$. Here the line stroke $S^{st}$ is called the straight stroke.         
When a line stroke $S_i$ is a curve stroke which is a loop stroke,  then each point $p_n^{S_i},\, 2\le\ n\le N_{S_i}-1$, satisfies\\
\begin{equation}||p_{N_{S_i}}^{S_i}-p_1^{S_i}||_2=\Delta,\,\,\, T_{1,N_{S_i}}^{dc}=360^o,\,\,\, T_{1,N_{S_i}}^{dc}=\sum_{n=1}^{N_{S_i}}\theta_n^{S_i},\end{equation}
where $T_{1,N_{S_i}}^{dc}$ is the total direction change of the stroke from point $p_1^{S_i}$ to $p_{N_{S_i}}^{S_i}$. Here all the points in the stroke are said to constitute a homogeneous region because they all satisfy property $(4)$. Such homogeneous region is represented as $S^{lp}$. Here the line stroke $S^{lp}$ is called a loop stroke.
\subsection{\bfseries{Sub-unit}}
A sub-unit is a point stroke $S^{pt}$ if a stroke has at most two points. If a stroke has more than two points then a sub-unit is defined as the stroke produced as the largest possible homogeneous region satisfying one of the four properties stated in section 3.1. A stroke can have one or more than one sub-units. Sub-unit corresponding to point stroke is shown in Figure 9. 
%\begin{comment}
%\setcounter{figure}{1.1}
\begin{figure}[ht!] 
\begin{center}
$\begin{array}{cc}
\includegraphics[width=.5\textwidth]{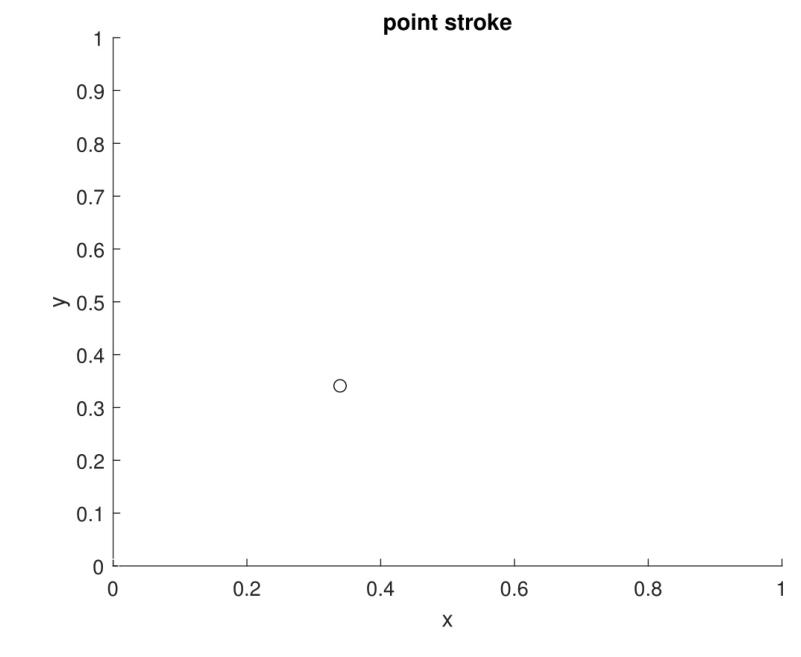}&
\includegraphics[width=.5\textwidth]{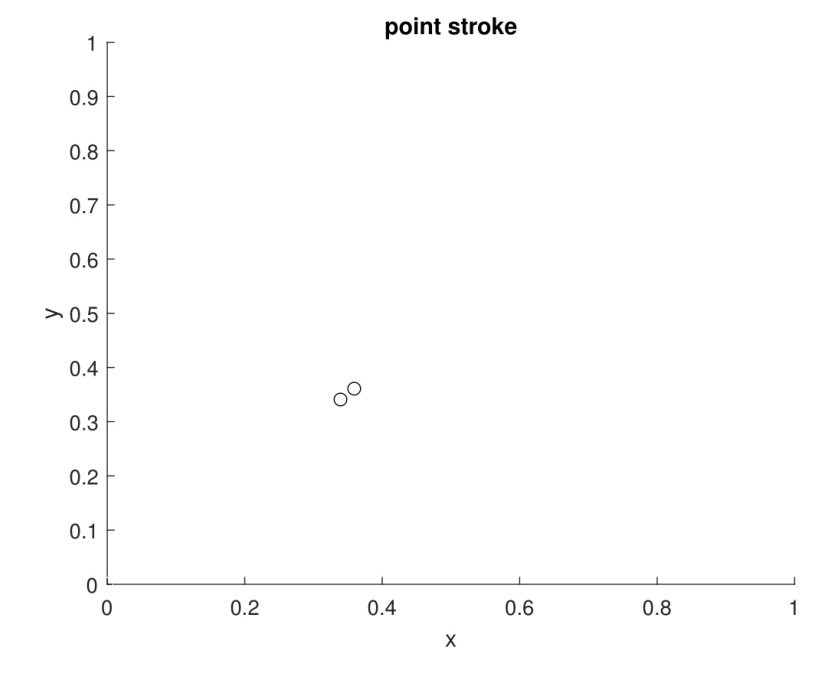}\\
\mbox{(a)} &\mbox{(b)}\\
\end{array}$
\caption{Sub-unit as a point stroke. (a) Sub-unit as a one-point stroke. (b) Sub-unit as a two-point stroke.}
\end{center}
\end{figure}
Sub-unit corresponding to a curve stroke is shown in Figure 10(a). Sub-unit in Figure 10(a) can be produced as $S^{ccw}$ shown in Figure 10(b) or can be produced as $S^{cw}$ as shown in Figure 10(c).         
\begin{figure}[ht!]
\begin{center}
$\begin{array}{ccc}
%\includegraphics[width=.35\textwidth]{figurec}&
%\includegraphics[width=.35\textwidth]{figurec1}&
%\includegraphics[width=.35\textwidth]{figurec2}\\
%\mbox{(a)} &\mbox{(b)}&\mbox{(c)}\\
%\end{array}$
%\caption{(a) Sub-unit as a curve stroke. (b) Sub-unit as a curve stroke in counter-clockwise %direction. (c) Sub-unit as a curve stroke in clockwise direction.}
\includegraphics[width=.31\textwidth]{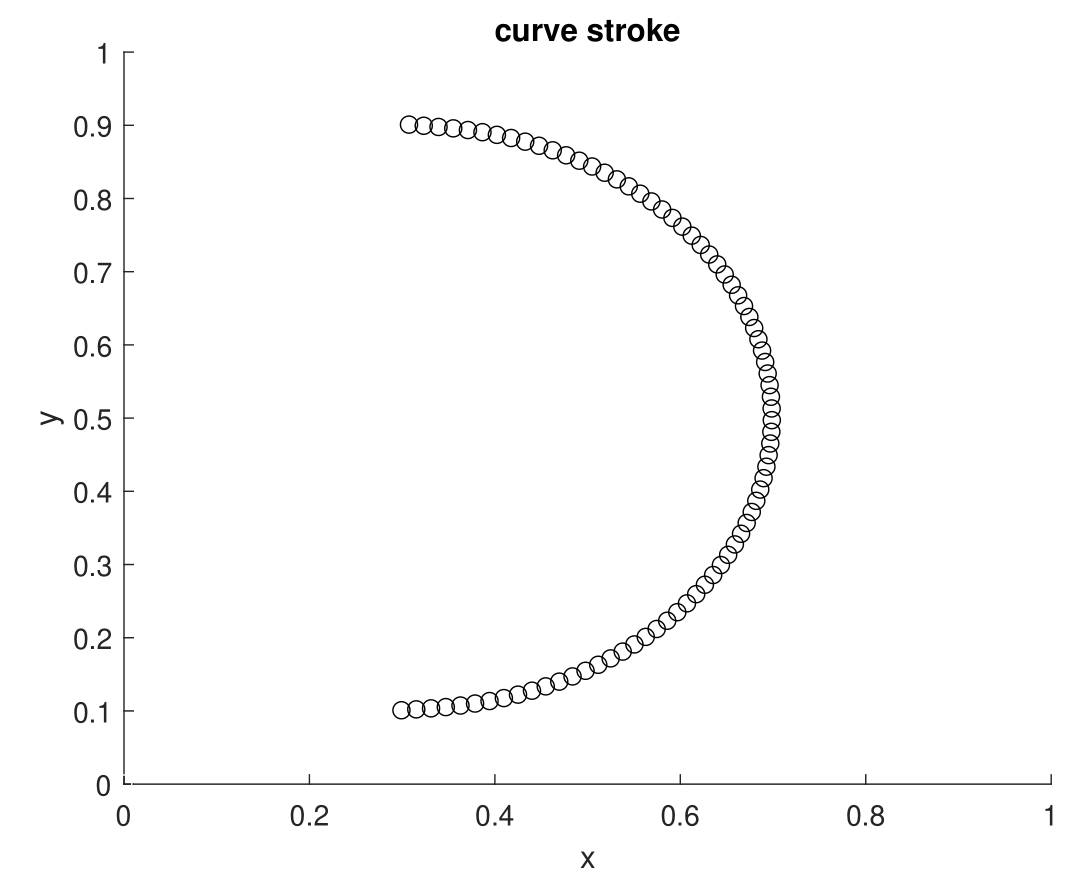}&
\includegraphics[width=.31\textwidth]{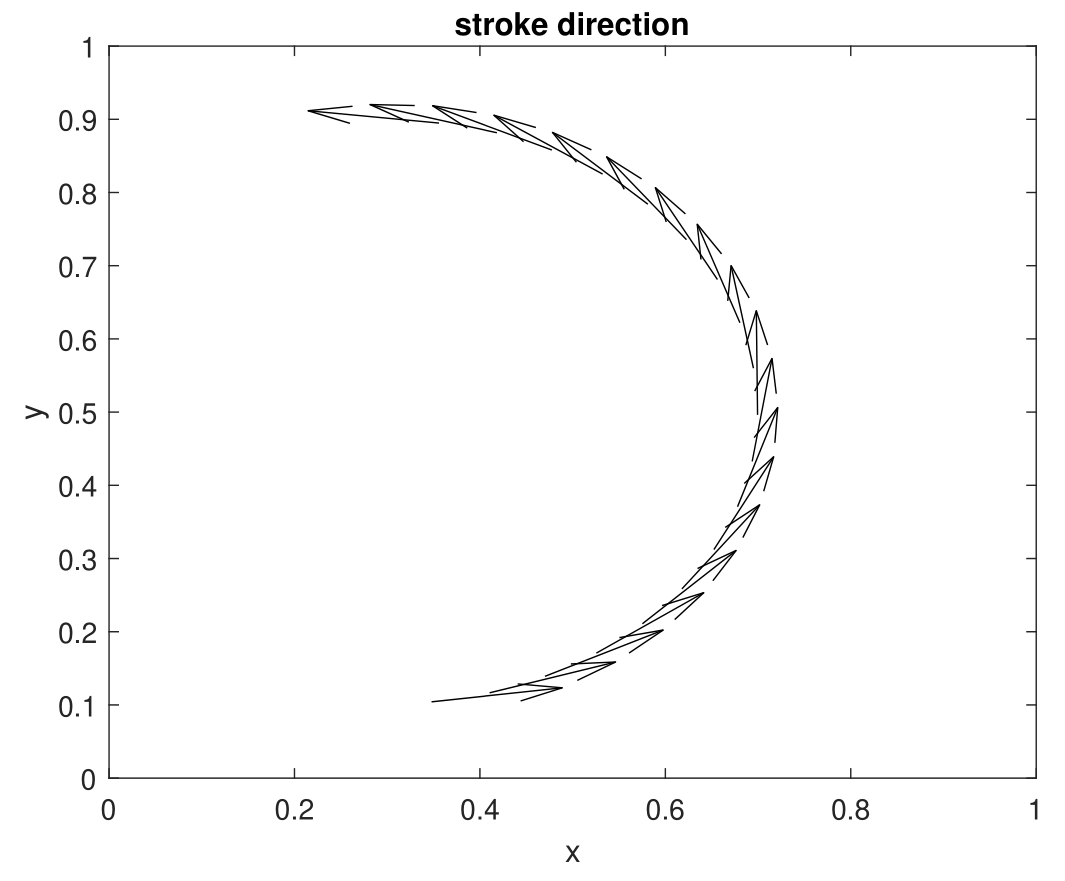}&
\includegraphics[width=.31\textwidth]{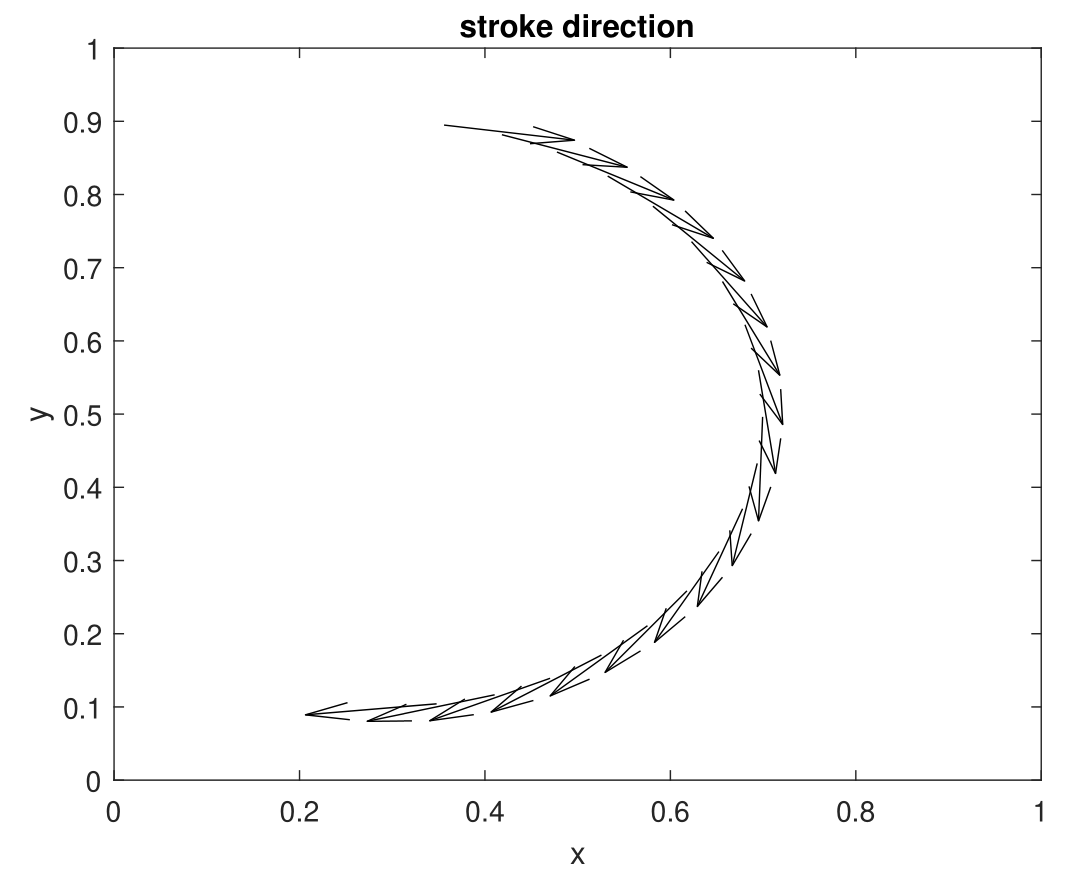}\\
\mbox{(a)} &\mbox{(b)}&\mbox{(c)}\\
\end{array}$
\caption{(a) Sub-unit as a curve stroke. (b) The sub-unit in (a) produced by pen movement in counter-clockwise direction and represented as $S^{ccw}$. (c) The sub-unit in (a) produced by pen movement in clockwise direction and represented as $S^{cw}$.}
\end{center}
\end{figure}
Sub-unit corresponding to a straight stroke is shown in Figure 11(a). Sub-unit in Figure 11(a) can be produced as $S^{st}$ by pen movement in downward direction as shown in Figure 11(b). Sub-unit in Figure 11(a) can be produced as $S^{st}$ by pen movement in upward direction as shown in Figure 11(c).        
\begin{figure}[ht!]
\begin{center}
$\begin{array}{ccc}
%\includegraphics[width=.35\textwidth]{figurest}&
%\includegraphics[width=.35\textwidth]{figurestd}&
%\includegraphics[width=.35\textwidth]{figurestu}\\
%\mbox{(a)} &\mbox{(b)}&\mbox{(c)}\\
\includegraphics[width=.31\textwidth]{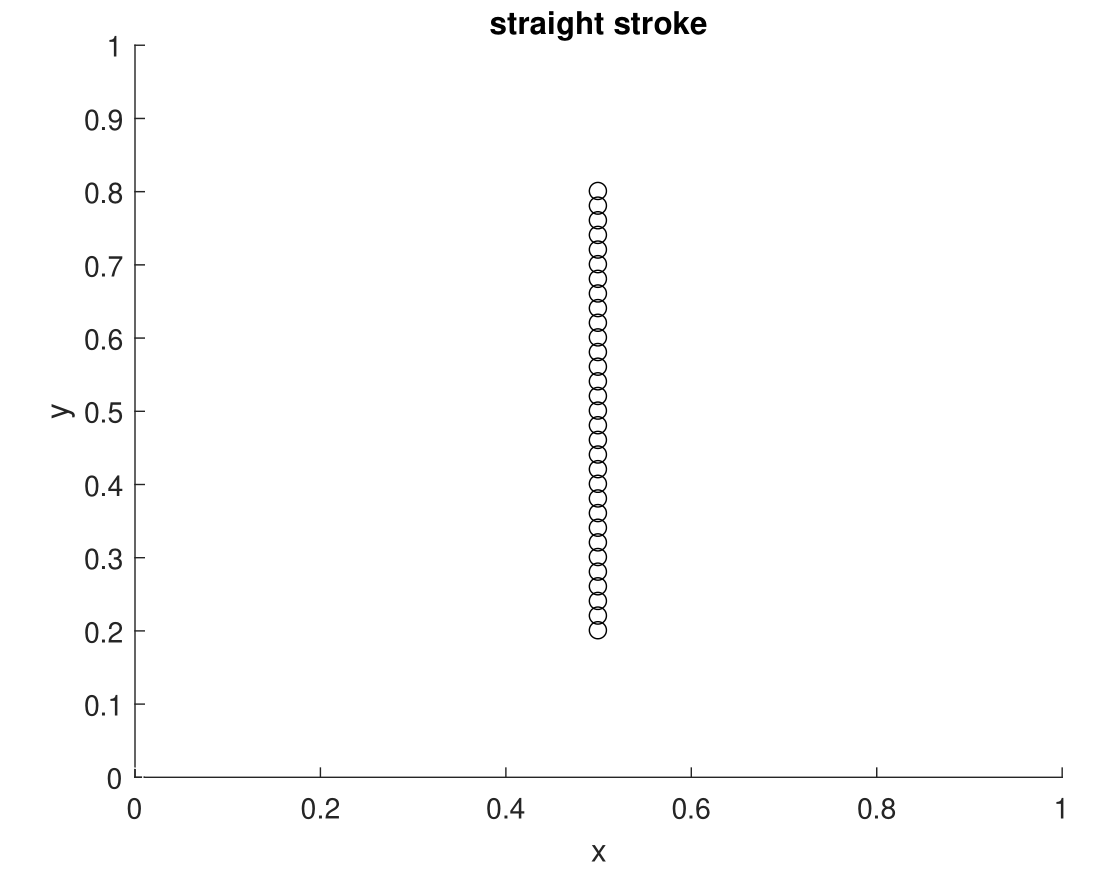}&
\includegraphics[width=.31\textwidth]{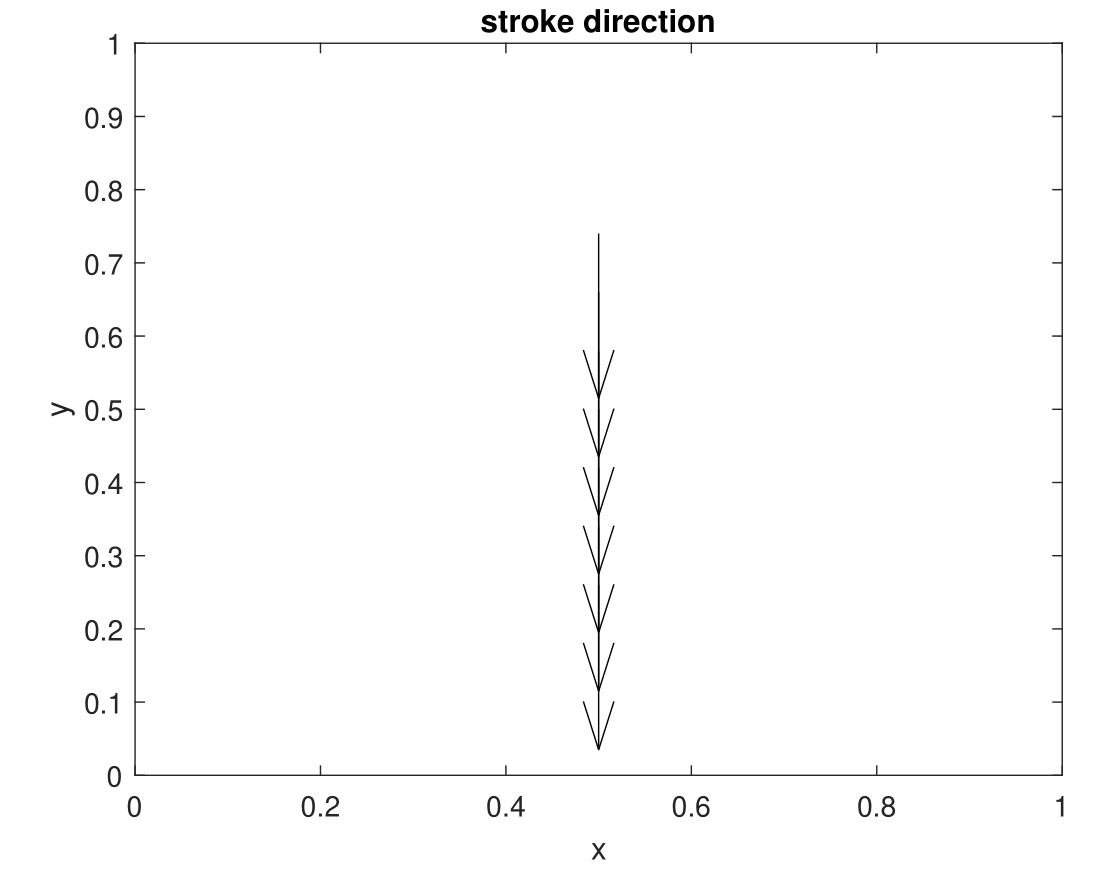}&
\includegraphics[width=.31\textwidth]{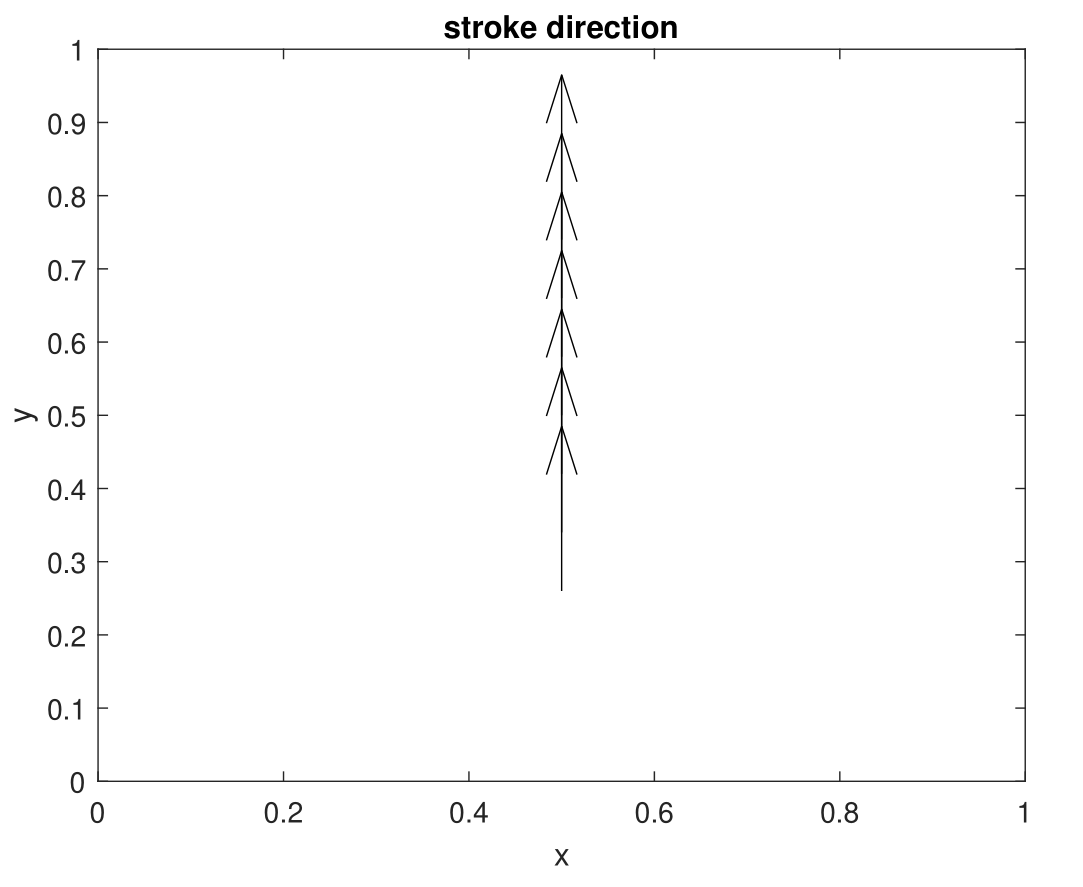}\\
\mbox{(a)} &\mbox{(b)}&\mbox{(c)}\\
\end{array}$
%\caption{(a) Sub-unit as a straight stroke. (b) Sub-unit as a straight stroke in downward %direction. (c) Sub-unit as a straight stroke in upward direction.}
\caption{(a) Sub-unit as a straight stroke. (b) The sub-unit in (a) produced by pen movement in downward direction and represented as $S^{st}$. (c) The sub-unit in (a) produced by pen movement in upward direction and represented as $S^{st}$.}
\end{center}
\end{figure}
Sub-unit corresponding to a loop stroke is shown in Figure 12. Sub-unit in Figure 12(a) can be produced as $S^{ccw}$ as shown in Figure 12(b). Sub-unit in Figure 12(a) can be produced as $S^{cw}$ as shown in Figure 12(c).        
\begin{figure}[ht!]
\begin{center}
$\begin{array}{ccc}
%\includegraphics[width=.35\textwidth]{figurelp}&
%\includegraphics[width=.35\textwidth]{figurelp1}&
%\includegraphics[width=.35\textwidth]{figurelp2}\\
%\mbox{(a)} &\mbox{(b)}&\mbox{(c)}\\
\includegraphics[width=.31\textwidth]{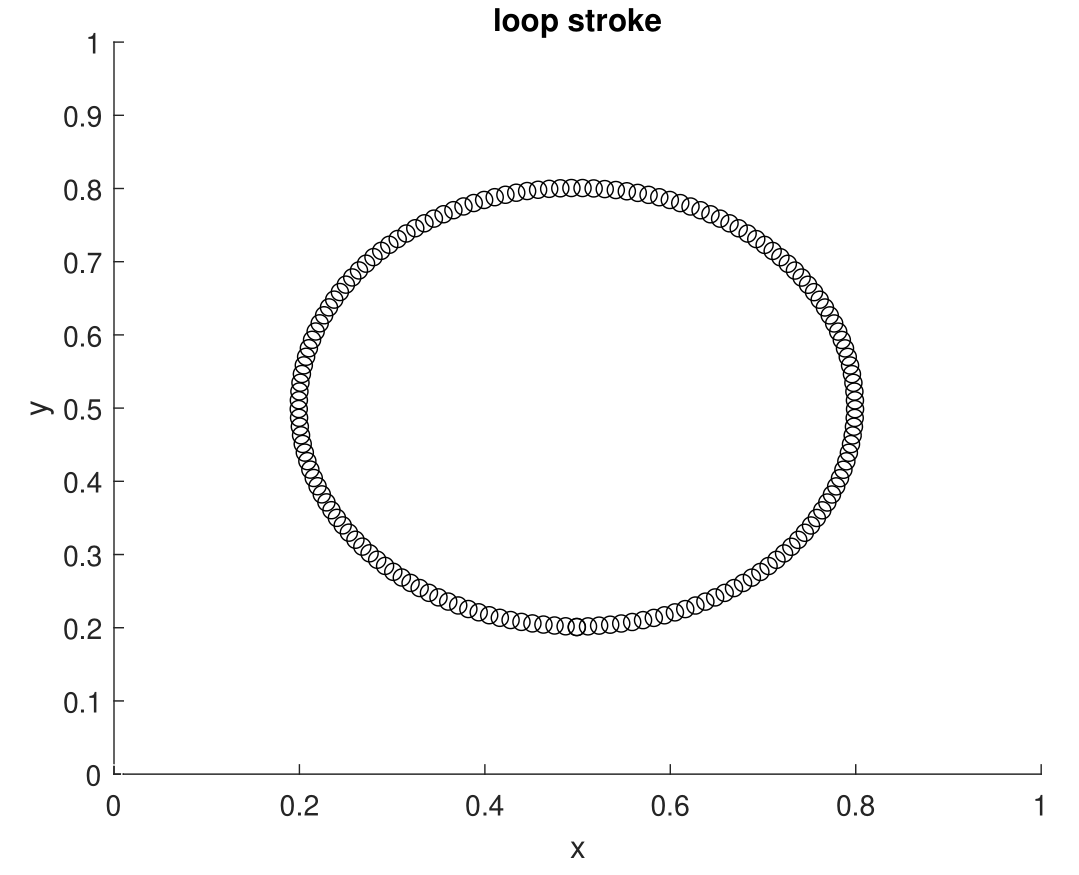}&
\includegraphics[width=.31\textwidth]{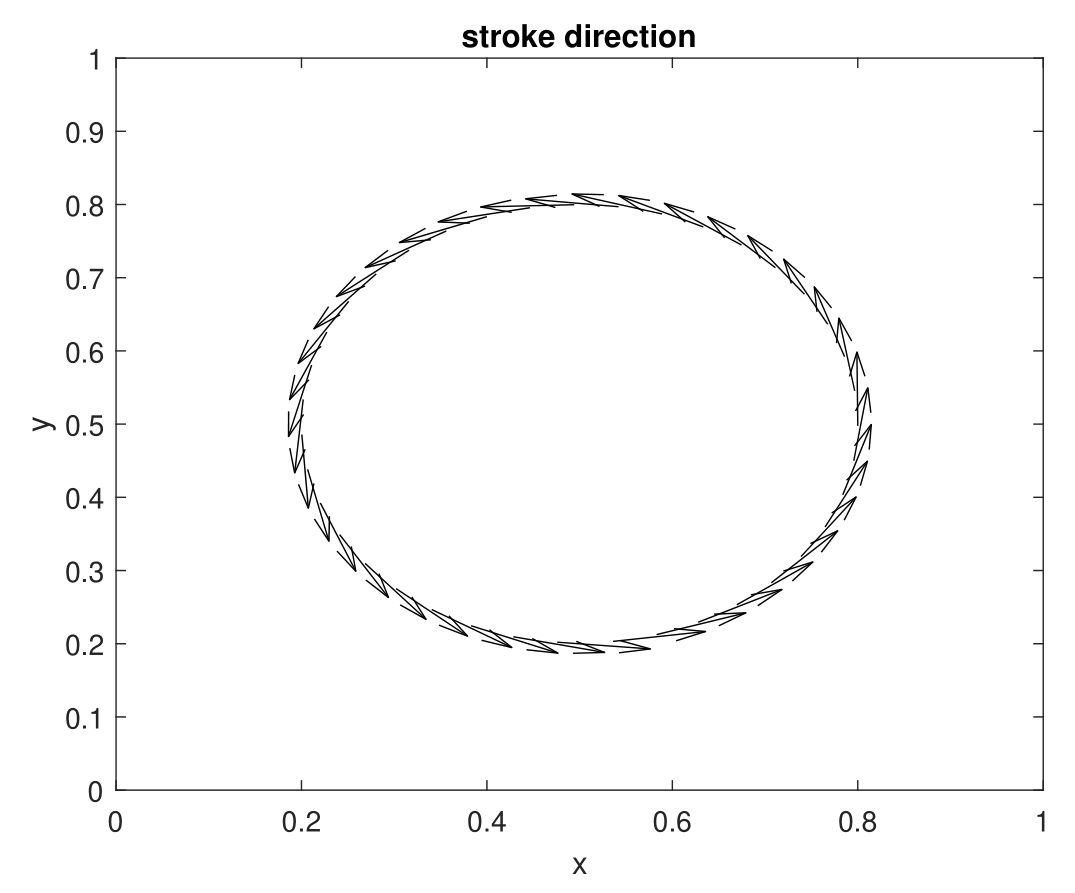}&
\includegraphics[width=.31\textwidth]{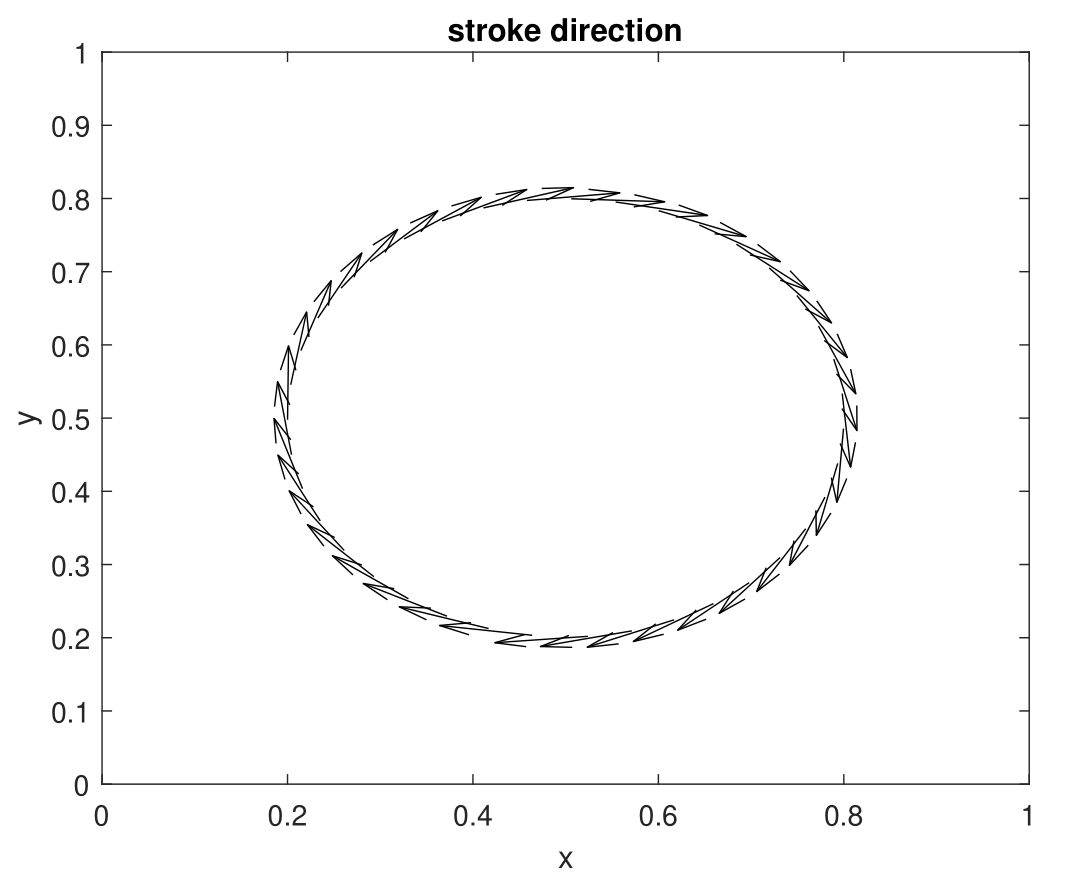}\\
\end{array}$
%\caption{(a)Sub-unit as a loop stroke. (b) Sub-unit as a loop stroke in counter-clockwise %direction.  (c) Sub-unit as a loop stroke in clockwise direction.}
\caption{(a) Sub-unit as a loop stroke. (b) The sub-unit in (a) produced by pen movement in counter-clockwise direction, satisfying (4), and represented as $S^{lp}$.  (c) The sub-unit in (a) produced by pen movement in clockwise direction, satisfying (4), and represented as $S^{lp}$.}
\end{center}
\end{figure}
It can be observed from the figures that the sub-units corresponding to curve and loop strokes can be produced by pen movement in clockwise or counter-clockwise directions. Sub-unit corresponding to a straight stroke can be produced by pen movement starting from one of the two end points. Sub-units corresponding to different line strokes can therefore be produced in two ways.

\subsection{A character in terms of sub-units}
\indent An online handwritten character is a sequence of strokes. A stroke can have one or more sub-units, and therefore, can be considered as a sequence of sub-units, and is represented as $S_i=(S_1^{u(i)},\dots\ ,S_j^{u(i)},\dots ,S_{N_{S_i^u}}^{u(i)})$. Here, $N_{S_i^u}$ is the number of sub-units in the $i^{th}$ stroke $S_i$. Therefore, a handwritten character is a sequence of one or more sub-units and can be represented as $C=(S_1^{u(1)},\dots\dots ,S_{N_{S^u_{N_S}}}^{u(N_S)})$. An example of an ideal Hindi online character is shown in Fig. 13(a). This character can be produced in many different ways; however, if it is produced as a sequence of single sub-unit strokes, then it has four sub-units. This character can be produced in a particular direction and order as shown in Fig. 13(b).
\begin{figure}[ht!]
\begin{center}
$\begin{array}{cc}
\includegraphics[width=.35\textwidth]{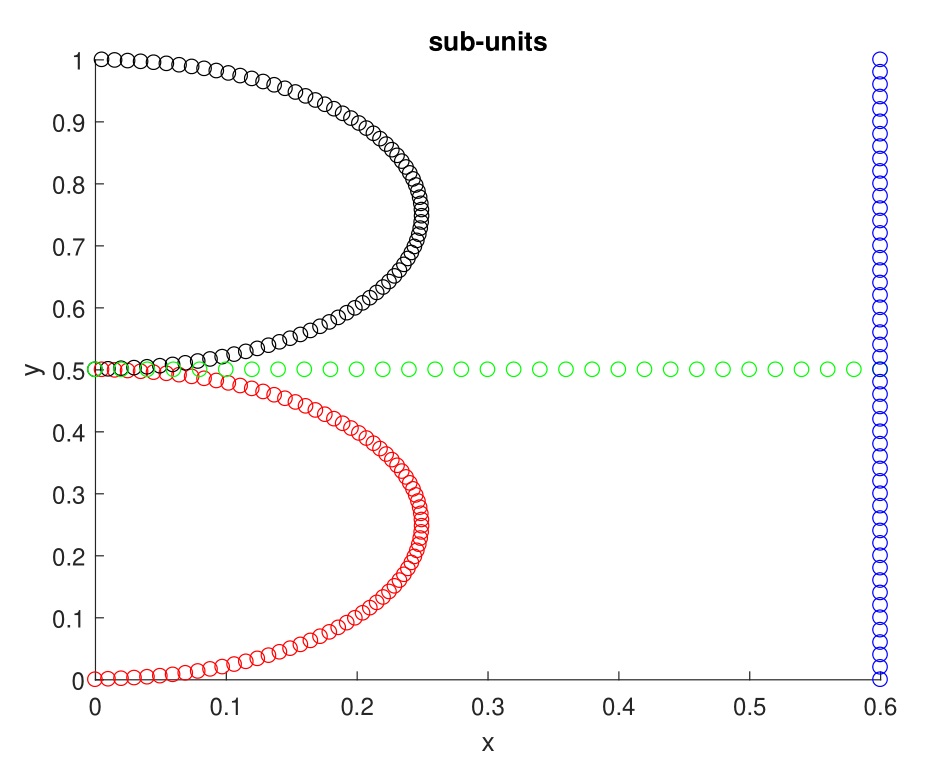}&
\includegraphics[width=.35\textwidth]{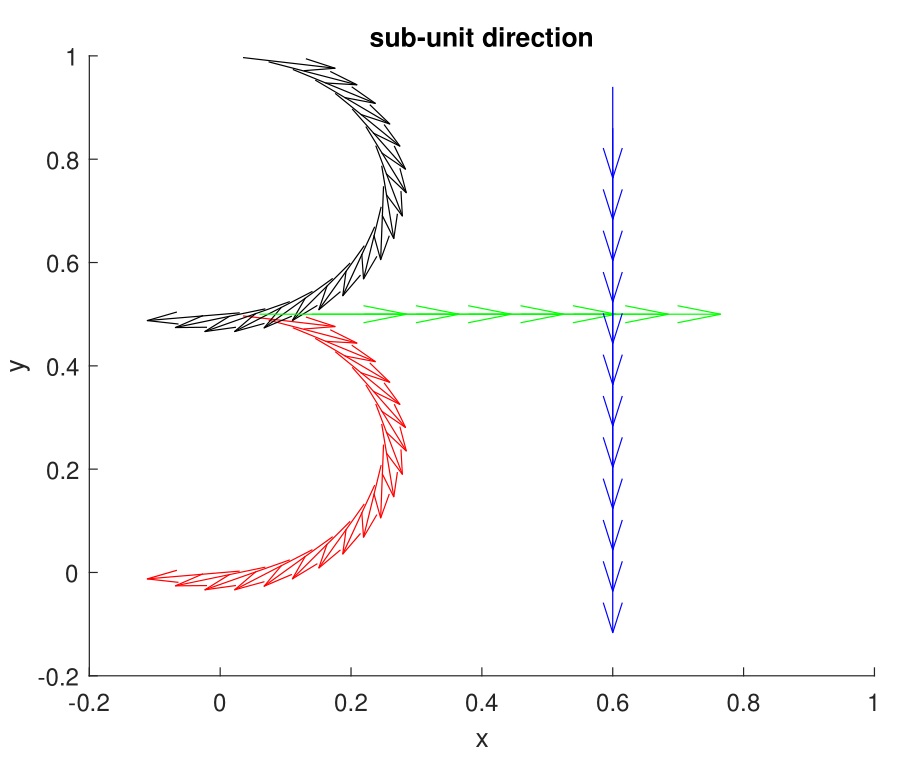}\\
\mbox{(a)} & \mbox{(b)}\\
\end{array}$
\caption{Hindi ideal online character. (a) The character is produced using four single sub-unit strokes. The order in which the strokes are generated, from first to last, is given by the color order black, red, green, and blue. The first sub-unit shown in black has been produced as $S^{cw}$. The second sub-unit shown in red is also produced as $S^{cw}$. The third sub-unit shown in green is produced as $S^{st}$ by pen movement towards the right in the horizontal direction. The fourth sub-unit shown in blue is produced as $S^{st}$ by pen movement in the downward direction. (b) The directions in which the strokes are produced are indicated by the arrows.}
\end{center}
\end{figure}
\subsection{Complexity of character representation}
Since each sub-unit in Figure 13(a) can be produced in two ways and there are four sub-units in the character, there are $2^4$ different stroke directions and $4!$ different stroke orders with total of $2^4\,4!$ different ways in which the character can be produced. Figure 14 shows some of the different sub-unit directions and orders in which the character can be produced.
 \begin{figure}[ht!]
\begin{center}
$\begin{array}{ccc}
\includegraphics[width=.35\textwidth]{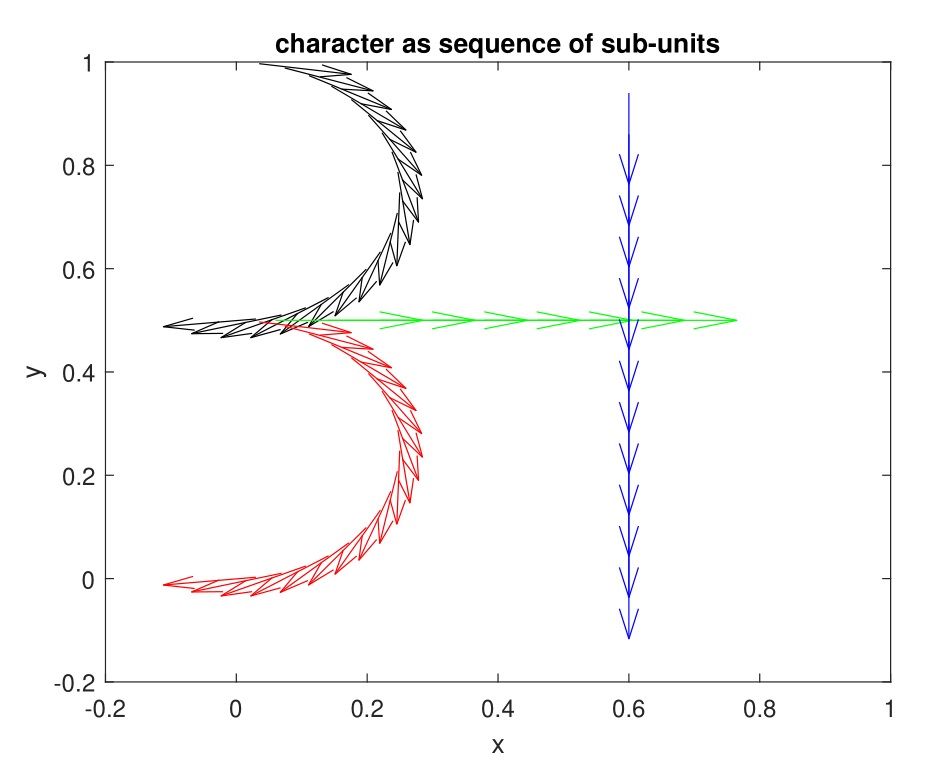} &
\includegraphics[width=.35\textwidth]{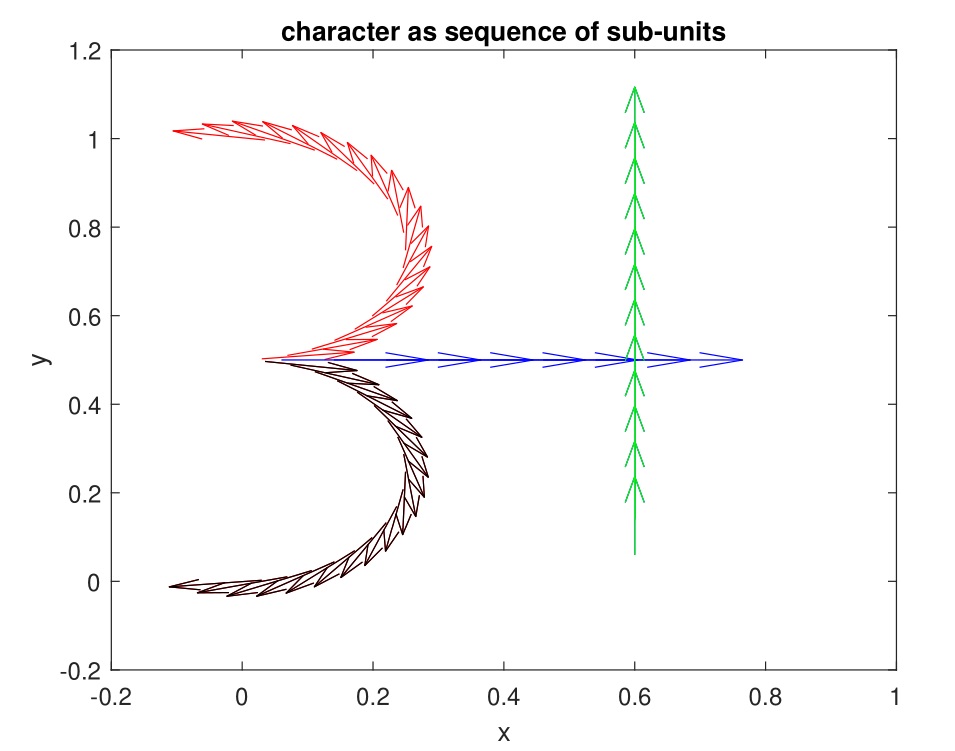}\\
\mbox{(a)} & \mbox{(b)}  \\
%\end{array}$
%\end{center}
%\end{figure}
% \begin{figure}[h]
%\begin{center}
%$\begin{array}{ccc}
\includegraphics[width=.35\textwidth]{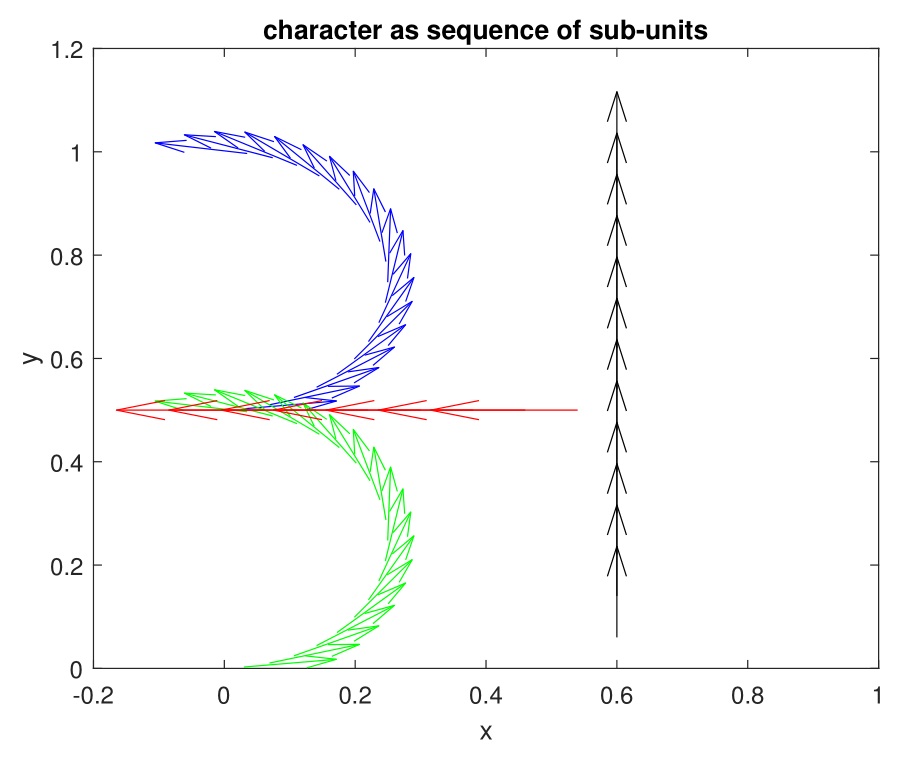}&
\includegraphics[width=.35\textwidth]{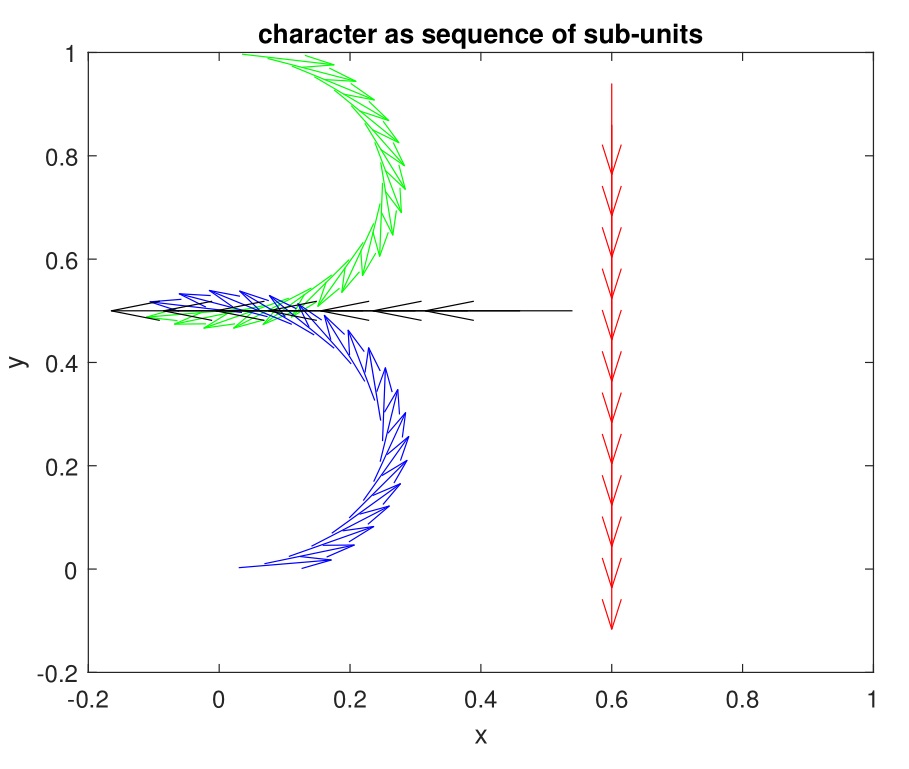}\\
\mbox{(c)} & \mbox{(d)}  \\
\end{array}$
\caption{Different ways of producing the character in Fig. 13(a). (a)-(d) Some of the different ways of producing the character in Figure 13(a). The characters are produced using four single sub-unit strokes. The order in which the strokes are generated, from first to last, is given by the color order black, red, green, and blue. The directions in which the strokes are produced are indicated by the arrows.}
\end{center}
\end{figure}
 
For a character with $N_C$ strokes and $N_C^u=\sum_{i=1}^{N_C}N_{S_i^u}$ sub-units, the character can be produced in at most $2^{N_C^u}\,\,N_C^u!$ different ways. The character as a sequence of points has different representations based on different ways in which it can be produced. The structure of the character is independent of the ways in which the character can be generated. The structure of the character is the structure of sub-units and the spatial relationships among the sub-units in the character.\\
\indent Let $x$- and $y$-co-ordinates of points in a character be considered as features for character representation for character recognition. The feature vector representation of the character can be obtained as a concatenation of the sequences of $x$- and $y$-co-ordinates of points in the character. The feature representation of the character will vary with the way the character has been produced. Therefore, a single character will have different point representations in the feature vector space. If the feature design is done in such a way that the feature representation is independent of the character generation mechanism, then the character will have a unique point representation in the feature vector space. If a statistical model based on sub-units, using such designed features of a character, is developed to capture the variations in handwritten characters, then the model recognition performance will be more robust to variations in the character generation mechanism.      
\section{Sub-unit extraction from an ideal character}
An ideal character has one or more ideal strokes. An ideal stroke has one or more sub-units constituting the structure of the character. These sub-units do not overlap with one another. A character can be produced by single sub-unit strokes, multiple sub-unit strokes, or a combination of both. Figure 15 shows a character produced by the single sub-unit strokes. The sub-unit structure in the stroke $S_i$ is given by $S_j^{u(i)}=S^{s(l)},\,\,j=1,\,\, l\in \{1,2,3,4\},\,\, S^{s(1)}=S^{c},\,\, S^{s(2)}=S^{st},\,\, S^{s(3)}=S^{lp},\,\, S^{s(4)}=S^{pt}$. The stroke $S_i$ is  
\begin{figure}[ht!]
\begin{center}
$\begin{array}{cc}
\includegraphics[width=.35\textwidth]{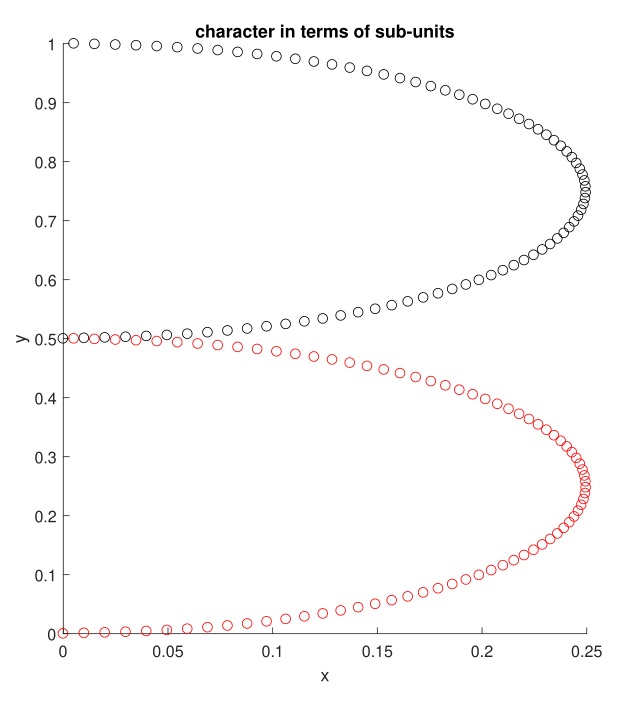} &
\includegraphics[width=.35\textwidth]{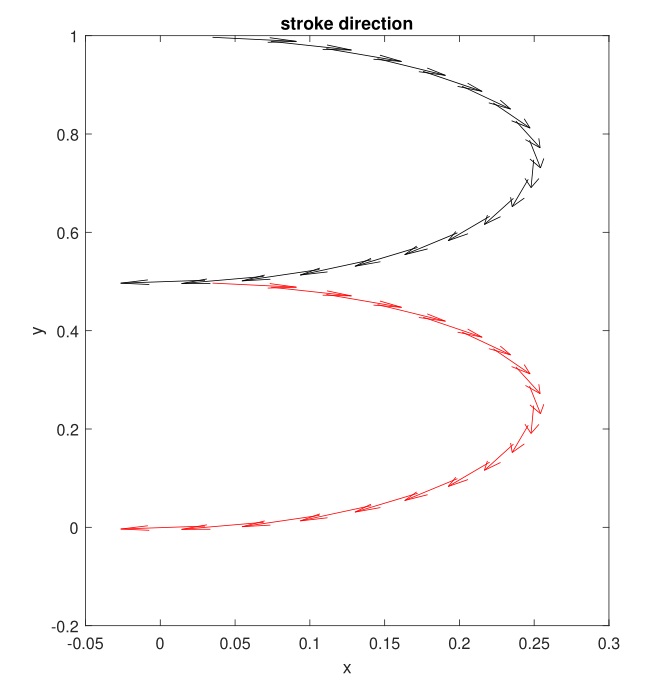}\\
\mbox{(a)} & \mbox{(b)}  \\
\end{array}$
\caption{Character produced using single sub-unit strokes. (a) The character is produced using two single sub-unit strokes. The first stroke  in the character is black and the second stroke is red. (b) The character is produced in the directions given by the arrows.}
\end{center}
\end{figure}
a single sub-unit stroke if for each point $p_n^{S_i}$,
\begin{equation} sign\big(f^d(v_{n-1}^{S_i},v_n^{S_i})\big)=sign\big(f^d(v_{n}^{S_i},v_{n+1}^{S_i})\big),\quad 2\le n \le N_{S_i}-2,
\,\,\,sign(x) = \begin{cases}
    1, &  x>0\\
    0, &  x=0\\
   -1, &  x<0\\
\end{cases}.\end{equation}   
The structure of such a stroke is,
\[S_j^{u(i)}=\begin{cases}
S^{s(l)},\quad j=1,\quad N_{S_i}>2,\quad l\in \{1,2,3\}\\
S^{s(l)},\quad j=1,\quad N_{S_i}\le 2,\quad l=4\\ 
\end{cases}.\]
Figure 16(a) shows a character produced by multiple sub-unit stroke. If a stroke is a multiple sub-unit stroke, then segmentation points in the stroke have to be determined in order to extract the sub-units from the stroke. The first and last points of every stroke are segmentation points. Let $\pi_m^i$ be the segmentation point index, then $\pi_1^i=1$ and $\pi_{N_{\pi^i}}^i=N_{S_i}$. Here $N_{\pi^i}$ is the number of segmentation points in a given stroke. A point in the stroke $S_i$ is a segmentation point for all stroke segments except for a loop stroke segment if, 
\begin{equation}\exists n,\,\, 2<n<N_{S_i}-2,\,\, \pi_m^i=n+1,\,\,  1<m\le N_{\pi^i},\,\,\mbox{s.t.}\,\,sign\big(f^d(v_{n-1}^{S_i},v_n^{S_i})\big)\ne sign\big(f^d(v_{n}^{S_i},v_{n+1}^{S_i})\big),\end{equation}
The points $p_{\pi_m^i}^{S_i}$ and $p_{\pi_{m+1}^i}^{S_i}$ in the stroke $S_i$ are the segmentation points for a loop stroke segment if, 
\[||p_{\pi_m^i}^{S_i}-p_{\pi_{m+1}^i-1}^{S_i}||_2=\Delta\,\,\,\mbox{for}\,\,\,\,1< \pi_m^i<\pi_{m+1}^i< N_{S_i}\,\,\mbox{and}\,\,\] 
\begin{equation}||p_{\pi_m^i}^{S_i}-p_{\pi_{m+1}^i}^{S_i}||_2=\Delta\,\,\,\mbox{for}\,\,\,\,\pi_m^i<\pi_{m+1}^i,\,\,\pi_{m+1}^i= N_{S_i}.\end{equation} 
{\flushleft Once $\pi_m^i,\,\,1\le m\le N_{\pi_i}$, have been determined, the sub-units in the stroke $S_i$ are extracted as,} 
\[S_j^{u(i)}=\begin{cases}S_i(\pi_m^i:\pi_{m+1}^i-1,*), & j=m,\quad 1\le m\le N_{\pi^i}-2\\ \\
S_i(\pi_m^i:\pi_{m+1}^i,*), & j=m,\quad m=N_{\pi^i}-1\\ \\ \end{cases}.\]
$S_i(n_1:n_2,*)$ is a sequence of points $p_n^{S_i}$ from $n=n_1$ to $n=n_2$. The structure of the extracted sub-unit is
\[S_j^{u(i)}=S^{s(l)},\quad 1\le j\le N_{\pi^i}-1,\quad l\in \{1,2,3\}.\]
\begin{figure}[ht!]
\begin{center}
$\begin{array}{cc}
\includegraphics[width=.35\textwidth]{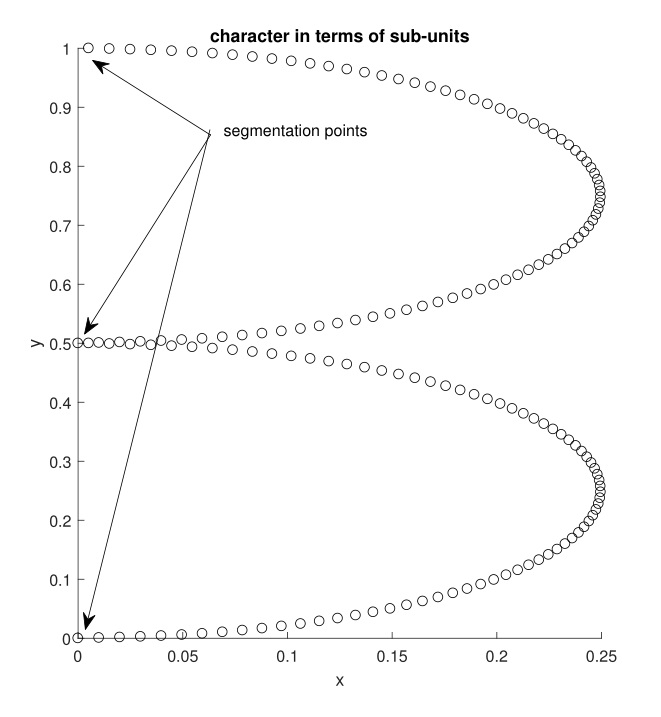} &
\includegraphics[width=.35\textwidth]{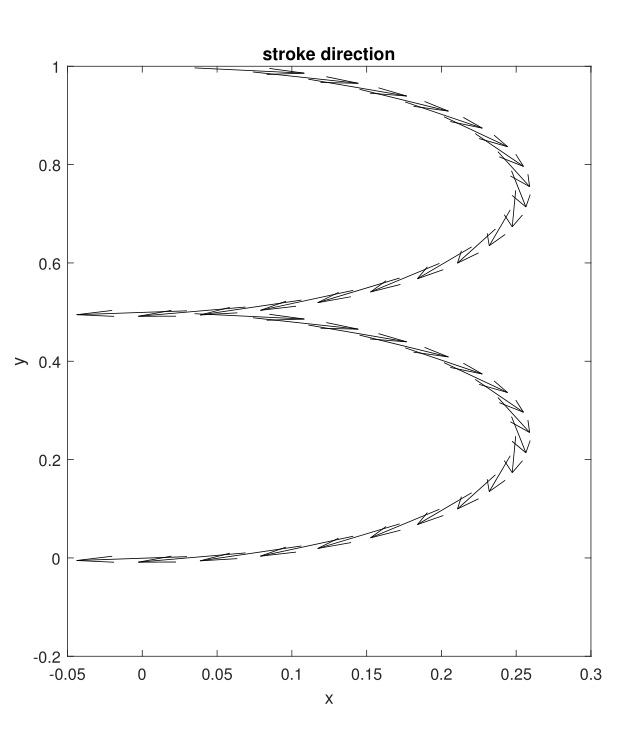}\\
\mbox{(a)} & \mbox{(b)}  \\
\end{array}$
\caption{Character produced using a multiple sub-unit stroke. (a) The character is produced using the stroke consisting of two sub-units. Each sub-unit is delimited by two segmentation points. (b) The character is produced in the directions given by the arrows.}
\end{center}
\end{figure}
\section{Sub-unit extraction from a handwritten character}
A sub-unit region in a handwritten character is the result of modification of the corresponding sub-unit region in the corresponding ideal character. This modification is caused by variations introduced in the character during the character generation. Therefore, the direction properties satisfied by a sub-unit region in an ideal character are not satisfied by the corresponding sub-unit region in the corresponding handwritten character. So, these properties cannot be directly used to extract sub-units from samples of handwritten characters. However, the structural properties of sub-units and the relationship among sub-units in a character can be used to build heuristics. These heuristics can then be used to extract sub-units from the handwritten characters.
\subsection{Direction property and direction change of an ideal character}
\indent Direction property of a stroke $S_i$ at a point $p_n^{S_i}$ in a character $C$ is obtained from the function, $f^d(v_{n-1}^{S_i},v_n^{S_i})=v_{n-1}^{S_i}(1)\, v_n^{S_i}(2)-v_{n-1}^{S_i}(2)\, v_n^{S_i}(1),\,\, 2\le n\le N_{S_i}-1$. Direction change of the stroke at the point $p_n^{S_i}$ is obtained as $\theta_n=\cos^{-1}\big({v_{n-1}^{S_i}}^T\, v_{n}^{S_i} \big),\,\, 2\le n\le N_{S_i}-1$.\\
\indent Figure 17(a) shows the Hindi ideal online character produced using three strokes. The first stroke, in black color, has two sub-unit regions, $S^{(s_1,s_2)}$ and $S^{(s_2,s_3)}$, and are produced as $S^{cw}$, as shown in Fig. 18. The transition region between these two sub-unit regions is a segmentation point, which is indicated by the presence of the point with index $s_2$, that satisfies (6). The region at $s_2$ is also a region of large direction change ($rldc$), $\theta_{rldc}\ge 85^{\circ}$, as shown in Fig. 19. The second stroke in red color and the third stroke in green color are both single sub-unit strokes and satisfy (5). They are the sub-unit regions $S^{(s_3,s_4)}$ and $S^{(s_4,s_5)}$ produced as $S^{st}$.
\begin{figure}[ht!]
\begin{center}
$\begin{array}{cc}
\includegraphics[width=.45\textwidth]{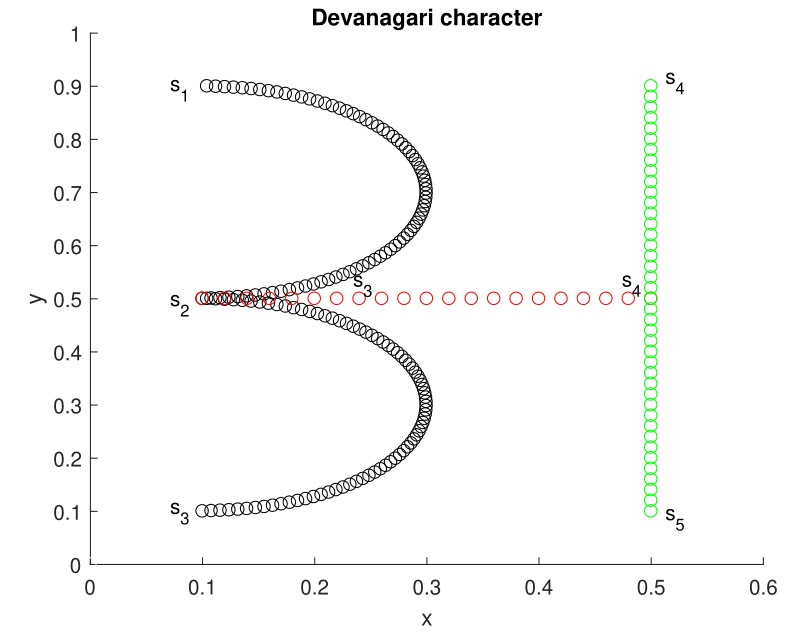}&
\includegraphics[width=.45\textwidth]{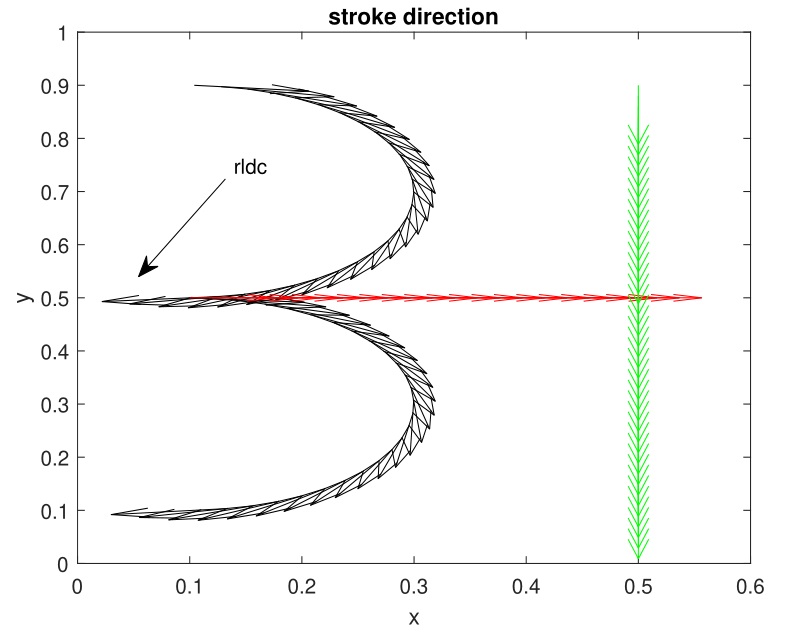}\\
\mbox{(a)} & \mbox{(b)}\\
\end{array}$
\caption{Hindi ideal online character. (a) The character is produced using three strokes. The first stroke in black color is a multiple sub-unit stroke and has two sub-units. Both the second stroke in red color and the third stroke in green color are single sub-unit strokes. (b) The character is produced in the directions given by the arrows.}
\end{center}
\end{figure}
   
\begin{figure}[ht!]
\begin{center}
\includegraphics[width=1\textwidth]{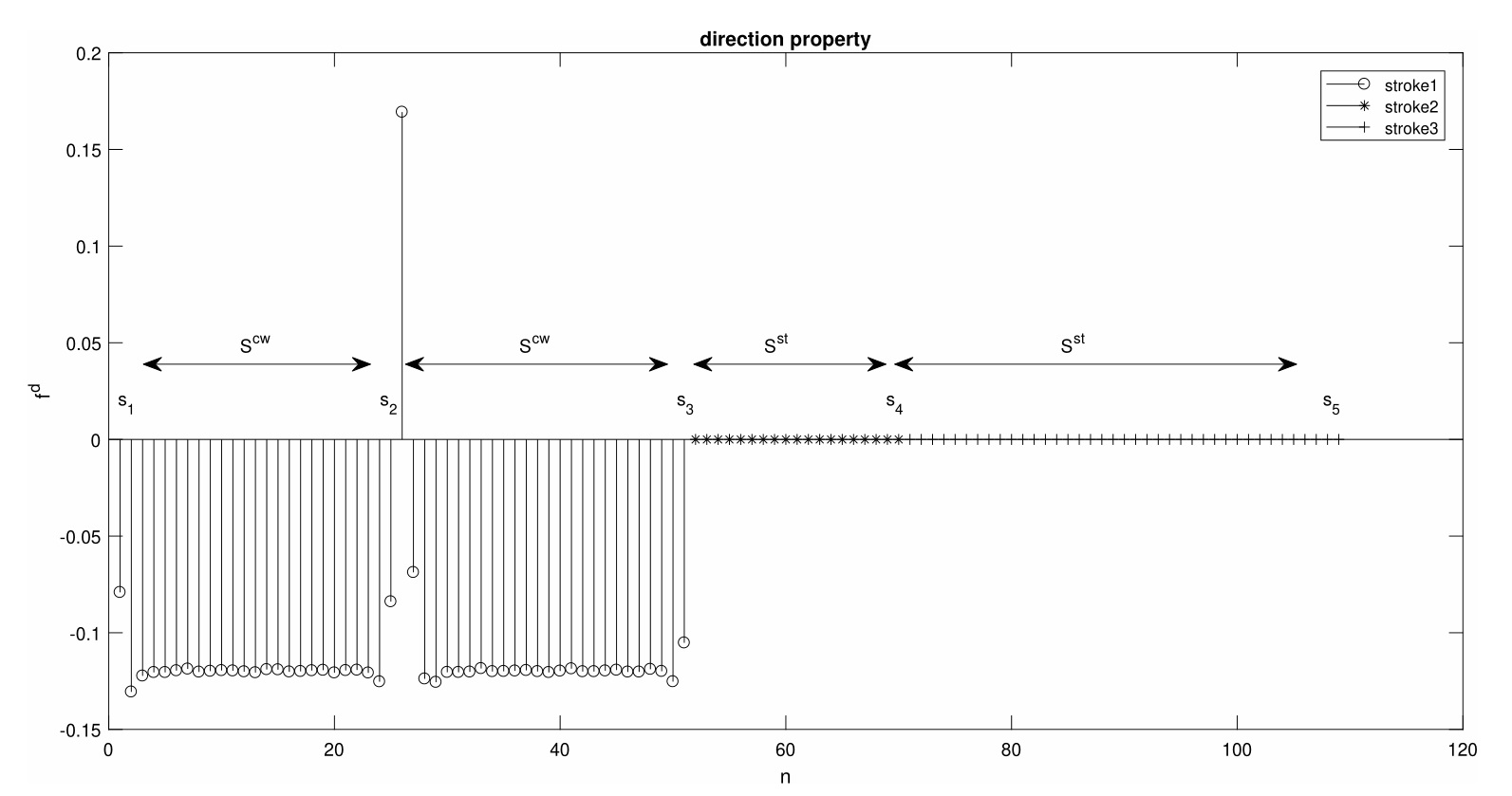}
\caption{Direction property of the character in Fig. 17(a). X-axis: n, y-axis: $f^d$. The direction property of the first stroke in black color, labeled stroke 1, is shown in the region delimited by the points with indices $s_1$ and $s_3$. By definition, the end points of a stroke are segmentation points. So, the points with indices $s_1$ and $s_3$ are the segmentation points. The point within the stroke with index $s_2$ is a segmentation point which is determined using (6). The first sub-unit region $S^{(s_1,s_2)}$ in the first stroke satisfies (1) and is produced as $S^{cw}$. The second sub-unit region $S^{(s_2,s_3)}$ in the first stroke also satisfies (1) and is also produced as $S^{cw}$. The direction property of the second stroke in red color, labeled stroke 2, is shown in the region delimited by $s_3$ and $s_4$. This stroke is a single sub-unit stroke as it satisfies (5). The corresponding sub-unit region $S^{(s_3,s_4)}$ satisfies (3) and is produced as $S^{st}$. The direction property of the third stroke in green color, labeled stroke 3, is shown in the region between $s_4$ and $s_5$. This stroke is also a single sub-unit stroke satisfying (5). The corresponding sub-unit region $S^{(s_4,s_5)}$ satisfies (3) and is produced as $S^{st}$.}
\end{center}
\end{figure}

\begin{figure}[ht!]
\begin{center}
\includegraphics[width=1\textwidth]{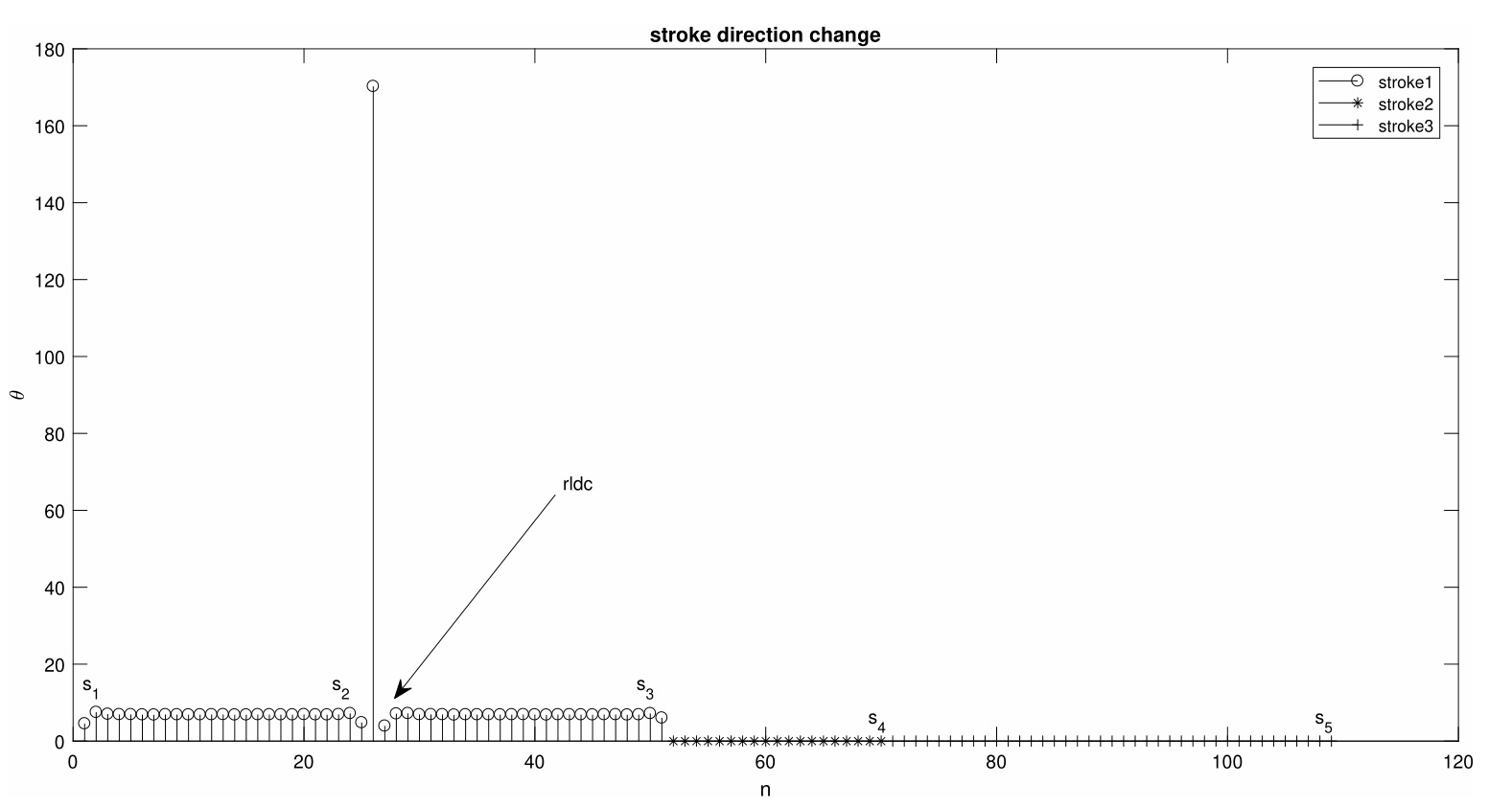} 
\caption{Direction change of the character in Fig. 17(a). X-axis: n, y-axis: $\theta$. The direction change of the first stroke in black color, labeled stroke 1, is shown in the region delimited by the points with indices $s_1$ and $s_3$. The point within the stroke with index $s_2$ is the segmentation point as determined using (6). This point is an $rldc$ and this property will be used for determination of segmentation points in handwritten characters. The direction change of the second stroke in red color, labeled stroke 2, is shown in the region delimited by $s_3$ and $s_4$. The direction change in this region is zero and is produced as $S^{st}$. The direction change of the third stroke in green color, labeled stroke 3, is shown in the region between $s_4$ and $s_5$. The direction change in this region is also zero and is produced as $S^{st}$.}
\end{center}
\end{figure}
Figure 20(a) shows another Hindi ideal online character produced by single stroke having multiple sub-units. This stroke has five sub-units. The first sub-unit region $S^{(s_1,s_2)}$ is a stroke segment produced as $S^{st}$ and the second sub-unit region $S^{(s_2,s_3)}$ is a stroke segment produced as $S^{ccw}$. The transition region between these two sub-unit regions is a segmentation point with index $s_2$ because it satisfies (6). This is also a region of large direction change as shown in Fig. 21. The third sub-unit region $S^{(s_3,s_4)}$ is a stroke segment produced as $S^{cw}$. The transition region between the second sub-unit region and third sub-unit region is a segmentation point with index $s_3$ that satisfies (6) and is shown in Fig. 21. This transition region $s_3$ is not a region of large direction change as shown in Fig. 22. The fourth sub-unit region $S^{(s_4,s_5)}$ is a stroke segment produced as $S^{lp}$ and can be extracted using (7). The fifth sub-unit region $S^{(s_5,s_6)}$ is a stroke segment produced as $S^{st}$ and follows the loop sub-unit region. The transition region between the fourth and the fifth sub-unit regions is determined using (7).
\begin{figure}[ht!]
\begin{center}
$\begin{array}{cc}
\includegraphics[width=.45\textwidth]{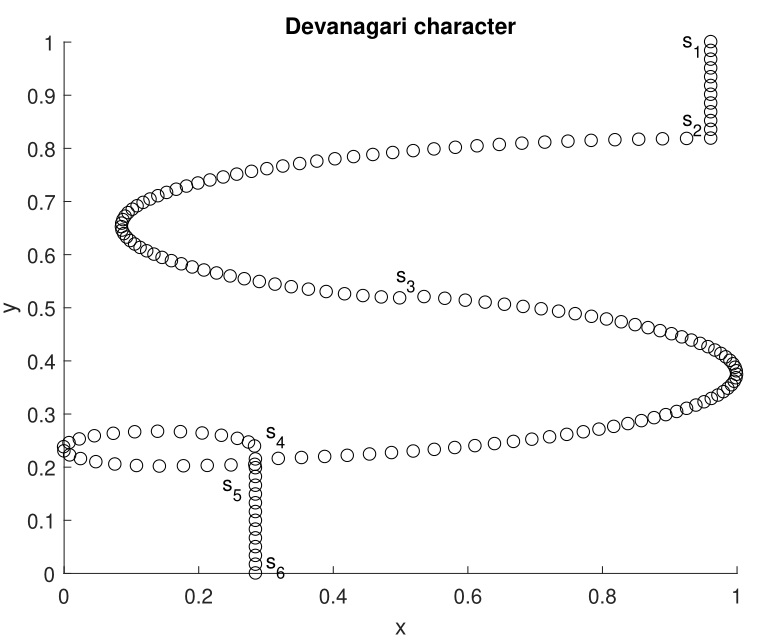} &
\includegraphics[width=.47\textwidth]{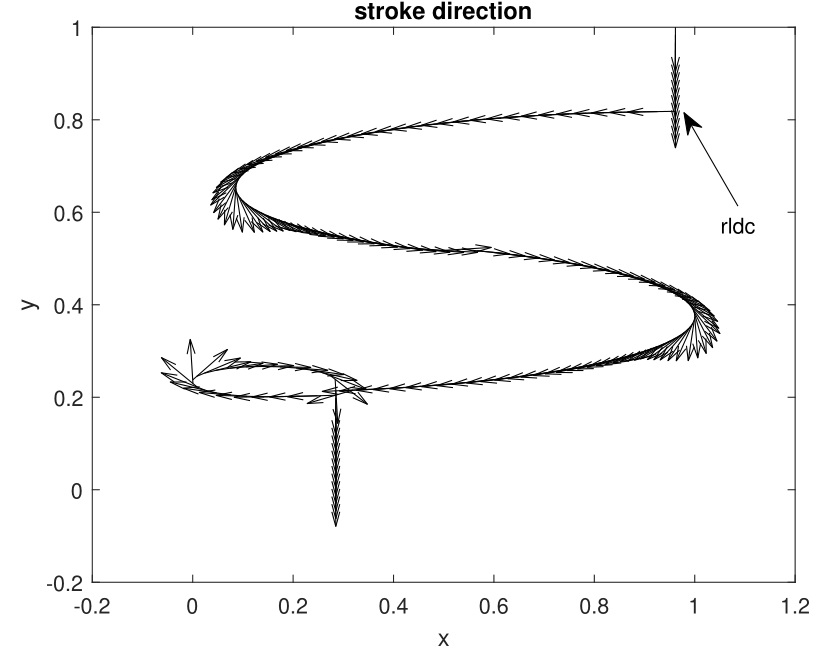}\\
\mbox{(a)} & \mbox{(b)}  \\
\end{array}$
\caption{Hindi ideal online character. (a) The character is produced using one stroke. The stroke is a multiple sub-unit stroke and has five sub-units. (b) The character is produced in the directions given by the arrows.}
\end{center}
\end{figure}

\begin{figure}[ht!]
\begin{center}
\includegraphics[width=1\textwidth]{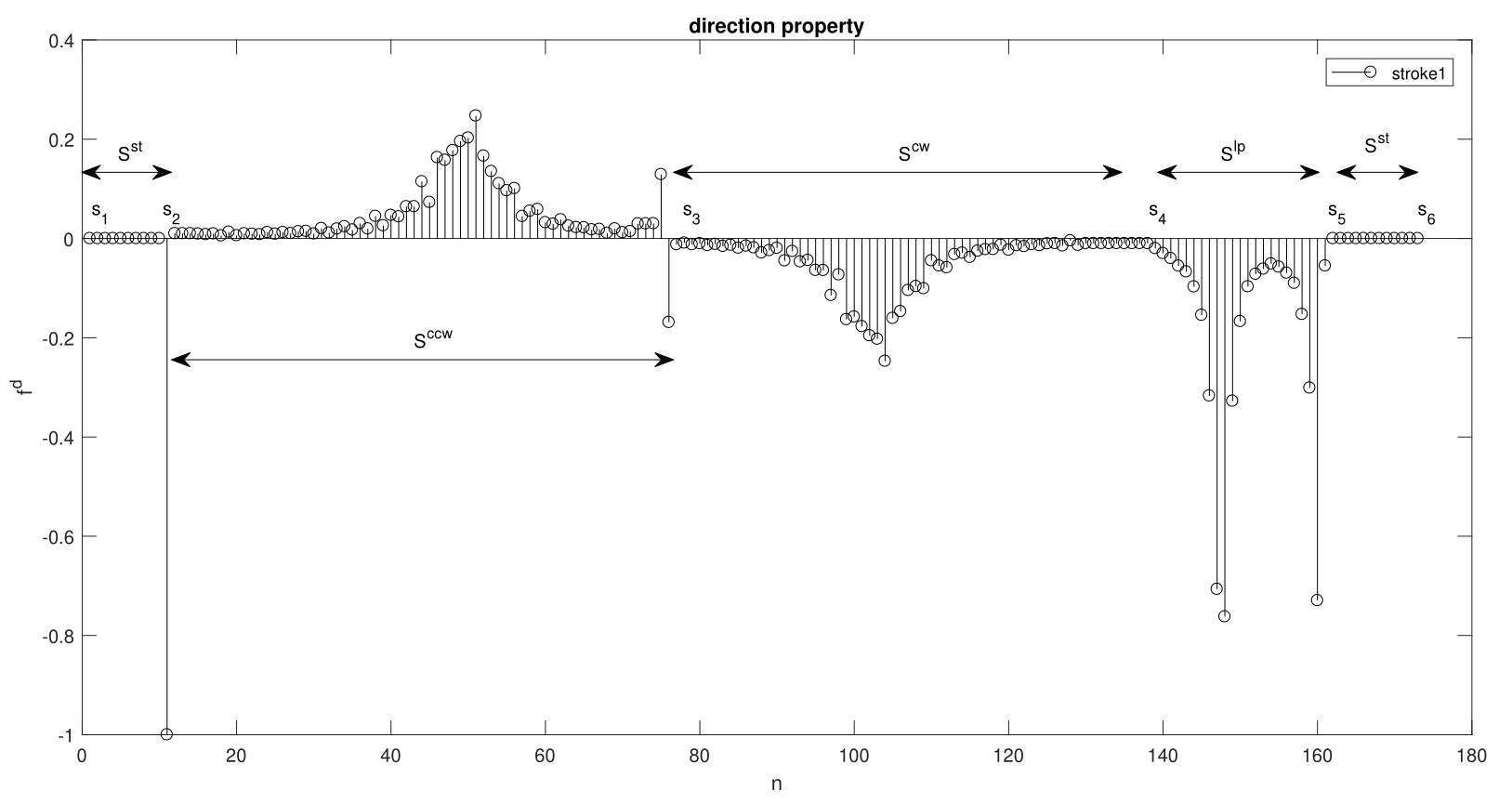} 
\caption{Direction property of the character in Fig. 20(a). X-axis: n, y-axis: $f^d$. The direction property of the stroke, labeled stroke 1, is shown in the region delimited by the points with indices $s_1$ and $s_6$. The end points with indices $s_1$ and $s_6$ are the segmentation points. The points within the stroke with indices $s_2$ and $s_3$ are the segmentation points, which are determined using (6). The first sub-unit region $S^{(s_1,s_2)}$ satisfies (3) and is produced as $S^{st}$. The second sub-unit region $S^{(s_2,s_3)}$ satisfies (2) and is produced as $S^{ccw}$. The points with indices $s_4$ and $s_5$ are the segmentation points for a loop stroke segment as they satisfy (7). The third sub-unit region $S^{(s_3,s_4)}$ satisfies (1) and is produced as $S^{cw}$. The fourth sub-unit region $S^{(s_4,s_5)}$ satisfies (4) and is produced as $S^{lp}$. The fifth sub-unit region $S^{(s_5,s_6)}$ satisfies (3) and is produced as $S^{st}$.}
\end{center}
\end{figure}

\begin{figure}[ht!]
\begin{center}
\includegraphics[width=1\textwidth]{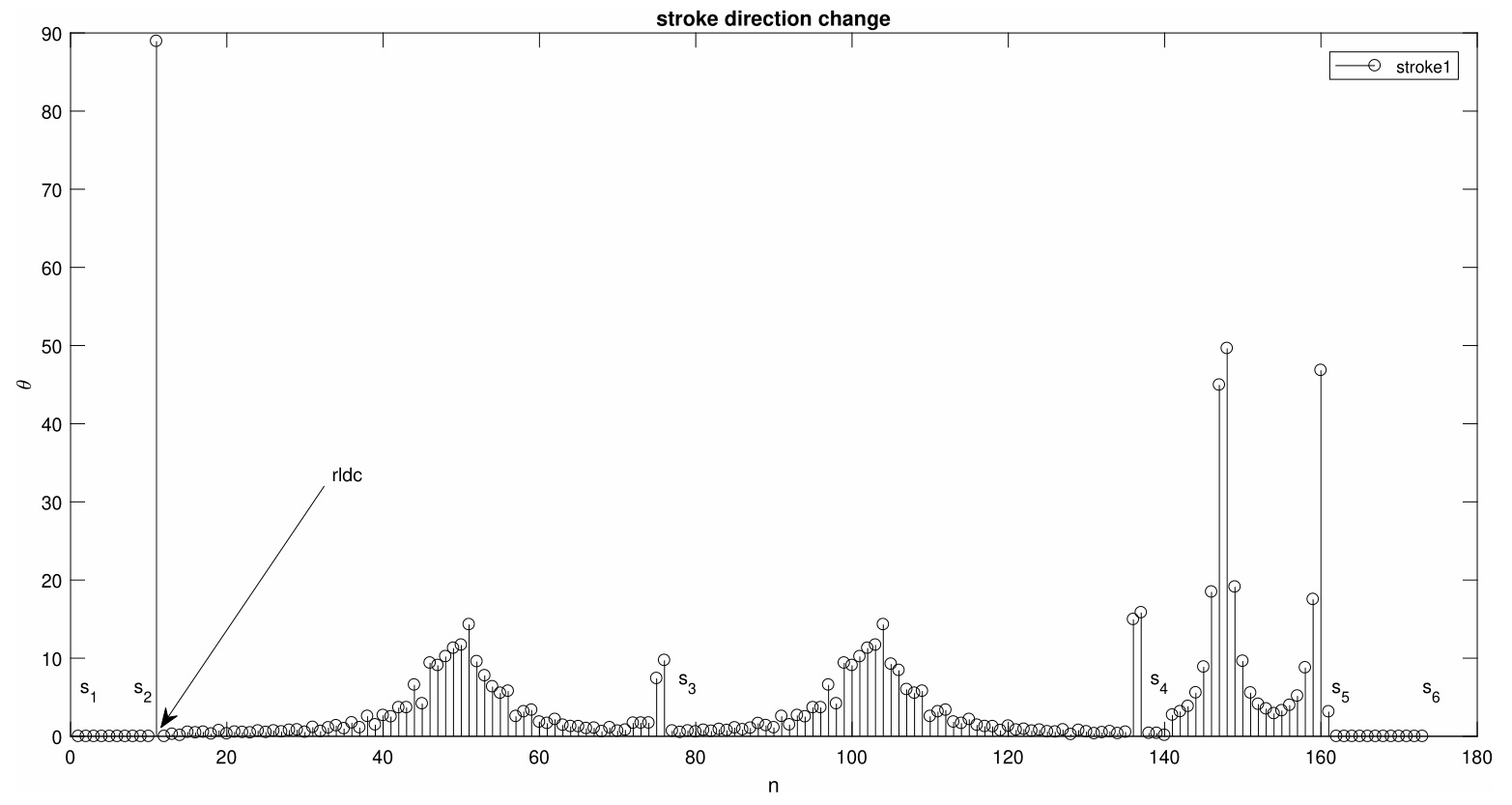} 
\caption{Direction change of the character in Figure 20(a). X-axis: n, y-axis: $\theta$. The direction change of the stroke, labeled stroke 1, is shown in the region delimited by the points with indices $s_1$ and $s_6$. The point within the stroke with index $s_2$ is the segmentation point as determined using (6). This point is an $rldc$ and will be used for determination of segmentation points in handwritten characters. The direction change values of the sub-unit regions $S^{(s_1,s_2)}$ and $S^{(s_5,s_6)}$ produced as $S^{st}$ are zero. The direction change values of the sub-unit regions $S^{(s_2,s_3)}$, $S^{(s_3,s_4)}$, and $S^{(s_4,s_5)}$ produced as $S^c$ are greater than zero.}
\end{center}
\end{figure}

\subsection{Direction property and direction change of a handwritten character}
\begin{figure}[ht!]
\begin{center}
$\begin{array}{cc}
\includegraphics[width=.45\textwidth]{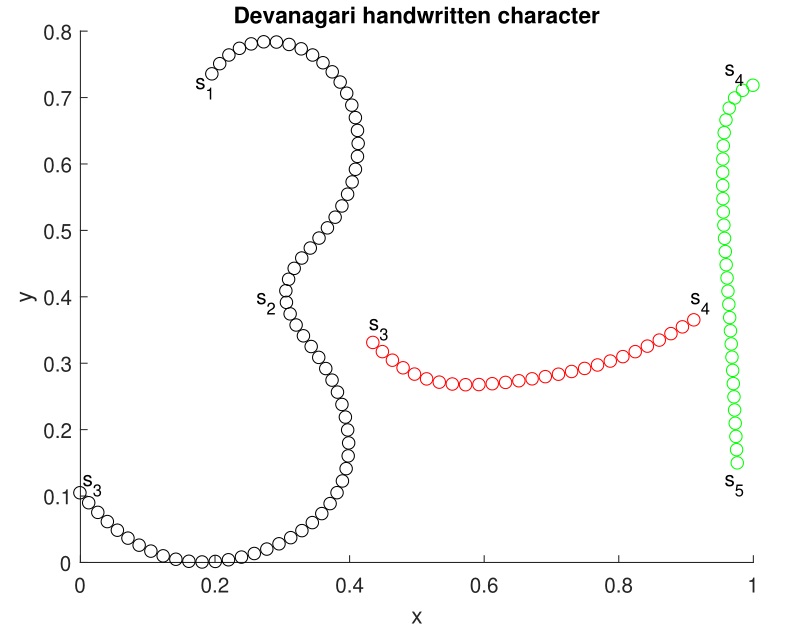}&
\includegraphics[width=.45\textwidth]{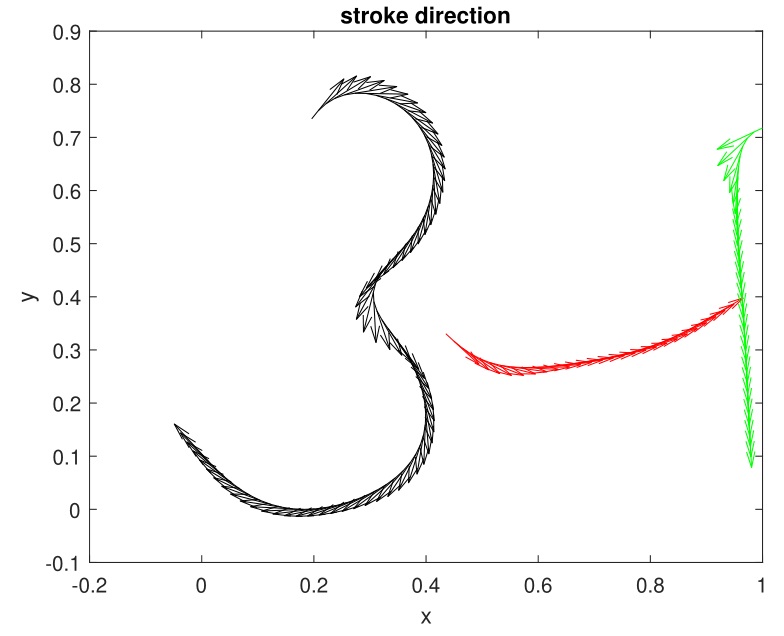}\\
\mbox{(a)} & \mbox{(b)}\\
\end{array}$
\caption{Character in Fig. 17(a) produced by handwriting. (a) The character is produced using three strokes. The three strokes in the color order black, red, and green are similar but not the same as the strokes in Fig. 17(a), in the same order. (b) The character is produced in the directions given by the arrows.}
\end{center}
\end{figure}
\indent Direction property and direction change of a handwritten character is different from that of the corresponding ideal character because of the writing style of an individual writing the character. Figure 23(a) shows the character in Fig. 17(a) produced by handwriting. This character is produced using three strokes. The first stoke is a multiple sub-unit stroke. It has two sub-unit regions $S^{(s_1,s_2)}$ and $S^{(s_2,s_3)}$ corresponding to the curve stroke segments, most of which are produced as $S^{cw}$. The transition region between these two sub-units at $s_2$ is indicated by the presence of the curve stroke segment produced as $S^{ccw}$, as shown in Fig. 24. This transition region does not correspond to the structure of the character shown in Fig. 17(a), therefore, it is defined as a pseudo sub-unit represented as ${S^\prime}_j^{u(i)},\,\, i=1,\,\, j=1$. Here, ${S^\prime}_j^{u(i)}$ is the transition region between the sub-units $S_j^{u(i)}$ and $S_{j+1}^{u(i)}$. The pseudo sub-unit region at $s_2$ is not a region of large direction change, as shown in Fig. 25. The second stroke is a single sub-unit stroke with sub-unit region $S^{(s_3,s_4)}$ produced as $S^{ccw}$, as shown in Fig. 24. The third stroke is also a single sub-unit stroke with sub-unit region $S^{(s_4,s_5)}$. The first half of $S^{(s_4,s_5)}$ is produced as $S^{ccw}$ and most of the second half of $S^{(s_4,s_5)}$ is produced as $S^{cw}$, as shown in Fig. 24. The second and third strokes should ideally be straight stroke segments, but are produced as curve stroke segments.
\begin{figure}[ht!]
\begin{center}
$\begin{array}{cc}
\includegraphics[width=1\textwidth]{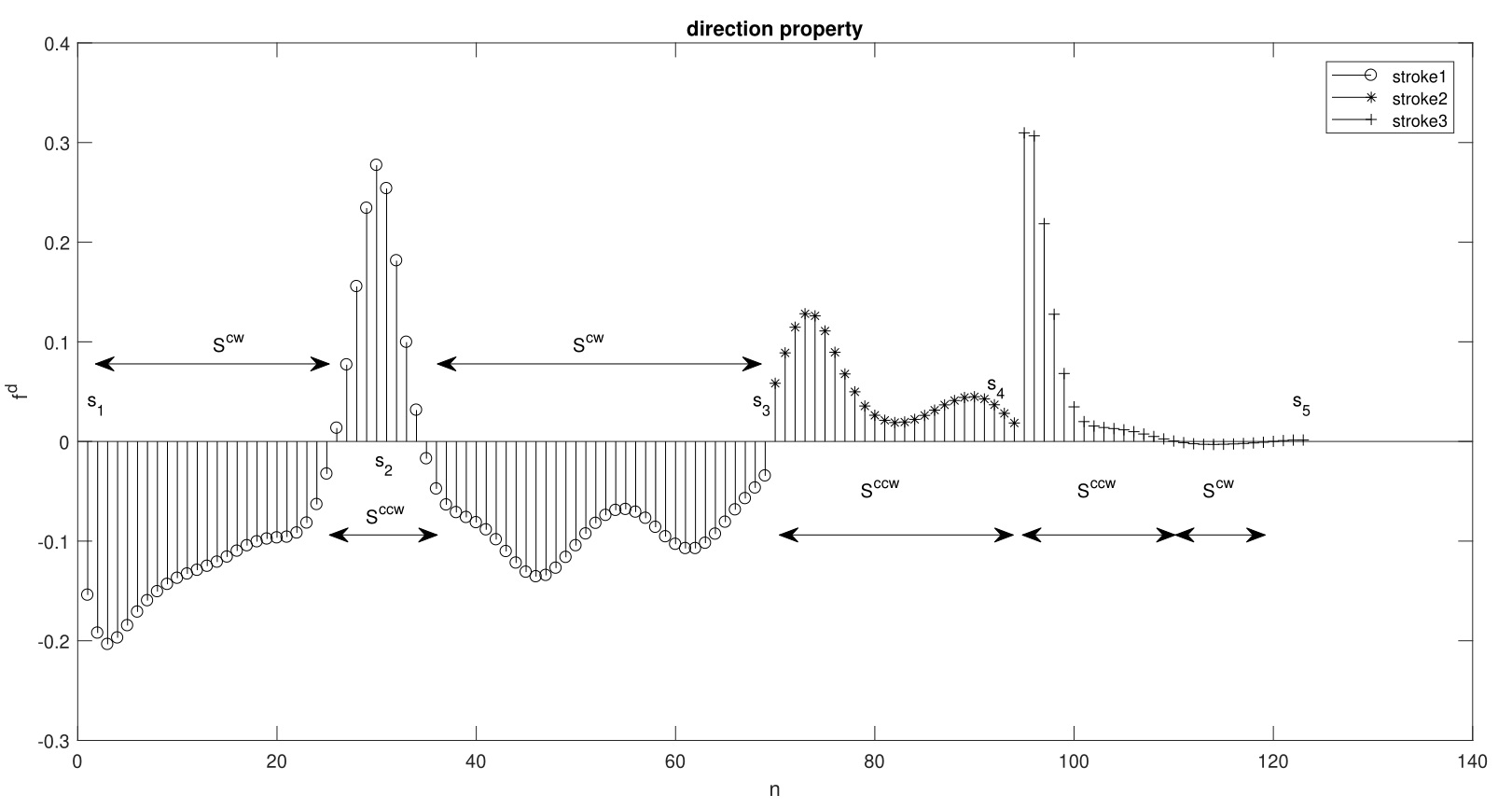}
\end{array}$
\caption{Direction property of the handwritten character in Fig.23(a). X-axis: n, y-axis: $f^d$. The direction property of the first stroke in black color, labeled stroke 1, is shown in the region delimited by the points with indices $s_1$ and $s_3$. The end points of the first stroke with these indices are the segmentation points. The first and second sub-unit regions $S^{(s_1,s_2)}$ and $S^{(s_2,s_3)}$ are not well defined because the point with index $s_2$ does not satisfy (6). The point is part of the stroke segment produced as $S^{ccw}$ which does not contribute to the structure of the corresponding ideal character in Fig. 17(a). Such stroke segment is called a pseudo sub-unit region and is represented as ${S^\prime}_j^{u(i)},\,\, i=1,\,\, j=1$. The direction property of the second stroke in red color, labeled stroke 2, is shown in the region delimited by $s_3$ and $s_4$. This stroke is produced as $S^{ccw}$ rather than as $S^{st}$ produced in the corresponding ideal character in Fig. 17(a). The direction property of the third stroke in green color, labeled stroke 3, is shown in the region between $s_4$ and $s_5$. The first half of this stroke is produced as $S^{ccw}$ and most of the second half as $S^{cw}$, unlike the corresponding ideal stroke in Fig. 17(a), which is produced as $S^{st}$.}
\end{center}
\end{figure}

\begin{figure}[ht!]
\begin{center}
$\begin{array}{cc}
\includegraphics[width=1\textwidth]{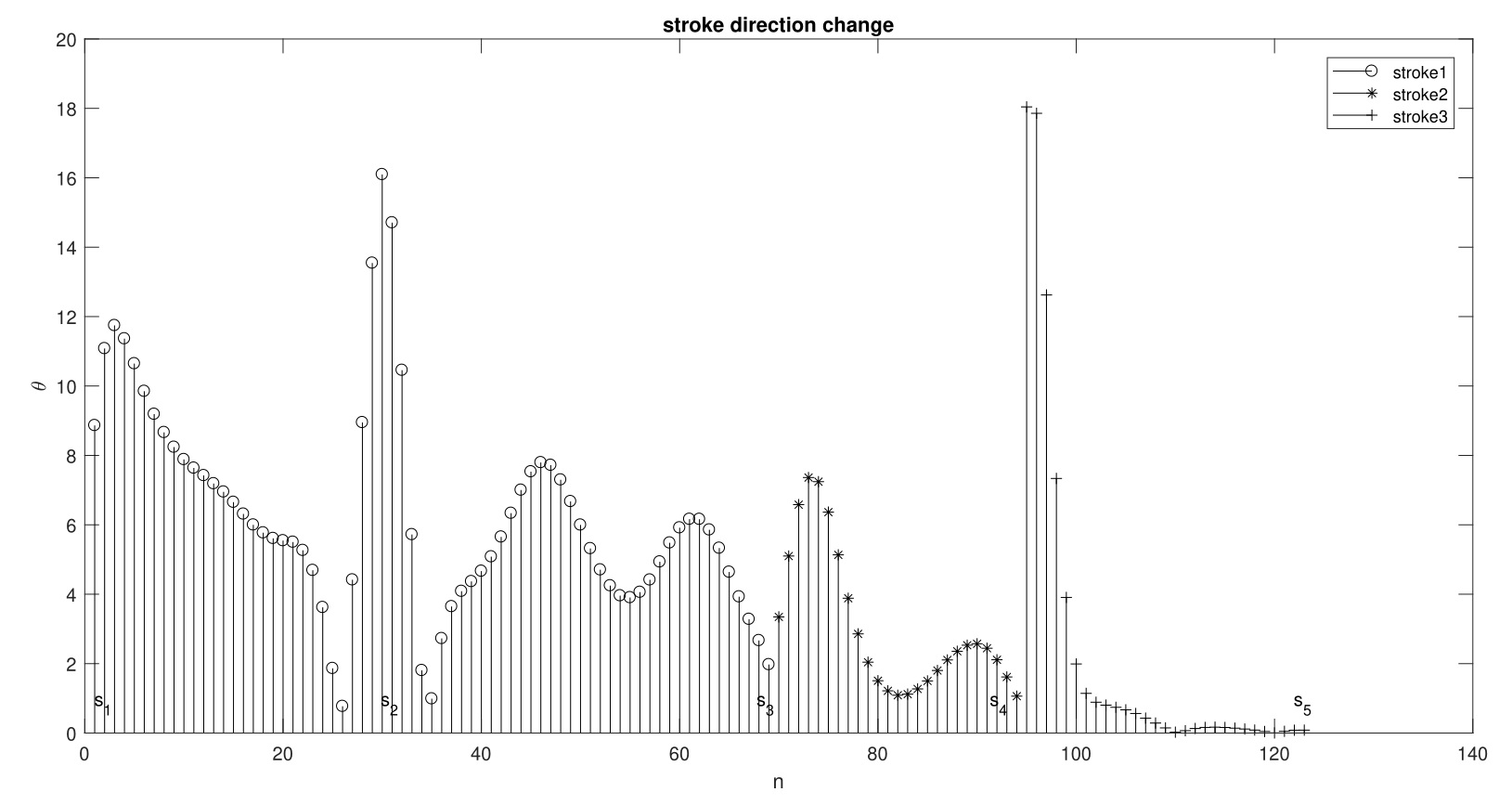} 
\end{array}$
\caption{Direction change of the handwritten character in Figure 23(a). X-axis: n, y-axis: $\theta$.The direction change of the first stroke in black color, labeled stroke 1, is shown in the region delimited by the points with indices $s_1$ and $s_3$. The point with index $s_2$ does not satisfy (6) and is also not an $rldc$. The direction change of the second stroke in red color, labeled stroke 2, is shown in the region delimited by $s_3$ and $s_4$. The direction change in this region is not zero. The direction change of the third stroke in green color, labeled stroke 3, is shown in the region between $s_4$ and $s_5$. The direction change in this region is also not zero.}
\end{center}
\end{figure}

\begin{figure}[ht!]
\begin{center}
$\begin{array}{cc}
\includegraphics[width=.45\textwidth]{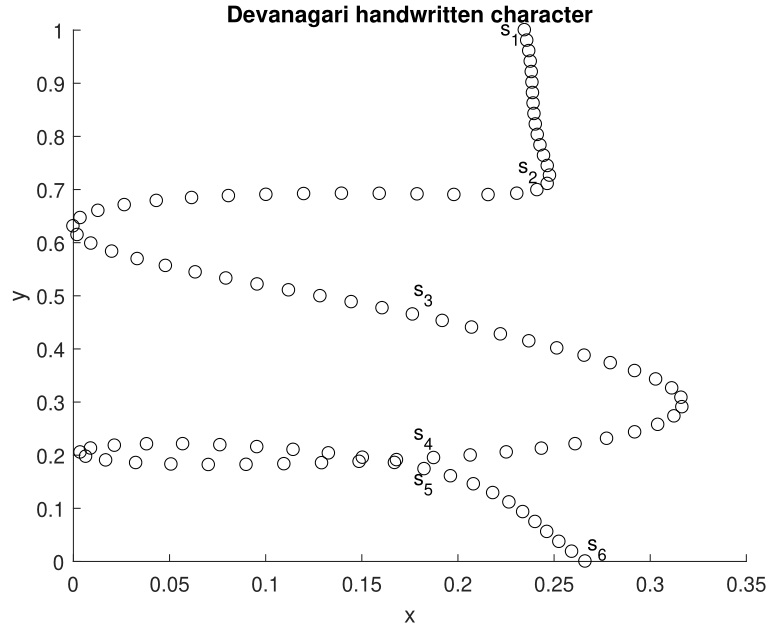}&
\includegraphics[width=.45\textwidth]{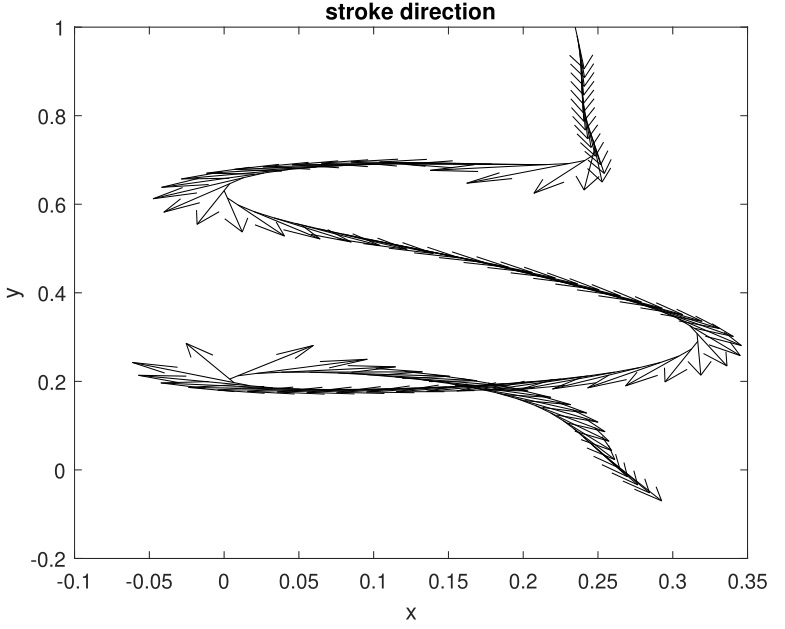}\\
\mbox{(a)} & \mbox{(b)}  \\
\end{array}$
\caption{Character in Fig. 20(a) produced by handwriting. (a) The character is produced using one stroke. The stroke is similar but not the same as the corresponding stroke in Fig. 20(a). (b) The character is produced in the directions given by the arrows.}
\end{center}
\end{figure}

\indent Figure 26(a) shows the character in Fig. 20(a) produced by handwriting. This character is produced using one stroke. This stoke is a multiple sub-unit stroke. It has five sub-units. The sub-unit region $S^{(s_1,s_2)}$ should ideally be a straight stroke segment. The first half of $S^{(s_1,s_2)}$ is produced as $S^{cw}$ and the second half, as $S^{ccw}$, as shown in Fig. 27. The transition region between the sub-unit regions $S^{(s_1, s_2)}$ and $S^{(s_2, s_3)}$ at $s_2$ is a stroke segment produced as $S^{cw}$, which is a pseudo sub-unit region, and is represented as ${S^\prime}_j^{u(i)},\,\, i=1,\,\, j=1$. The second sub-unit region $S^{(s_2,s_3)}$ is mostly produced as $S^{ccw}$. The transition region between the second sub-unit region $S^{(s_2,s_3)}$ and the third sub-unit region $S^{(s_3,s_4)}$ is at the point with index $s_3$ that satisfies (6), as shown in Fig. 27. The fourth sub-unit region $S^{(s_4,s_5)}$ is a loop stroke segment. If the direction from the last point to the first point of the loop stroke segment is considered as in (4), then the total change in direction is $552^\circ(> 360^\circ)$, and if not considered, is $227^\circ(< 360^\circ)$. The distance between the end points of this segment, with indices $s_4$ and $s_5$, is not equal to $\Delta$ as required by (4). The fifth sub-unit region $S^{(s_5,s_6)}$ should ideally be a straight stroke segment. Most of $S^{(s_5,s_6)}$ is produced as $S^{cw}$ as shown in Fig. 27.\\
\begin{figure}[ht]
\begin{center}
$\begin{array}{cc}
\includegraphics[width=1\textwidth]{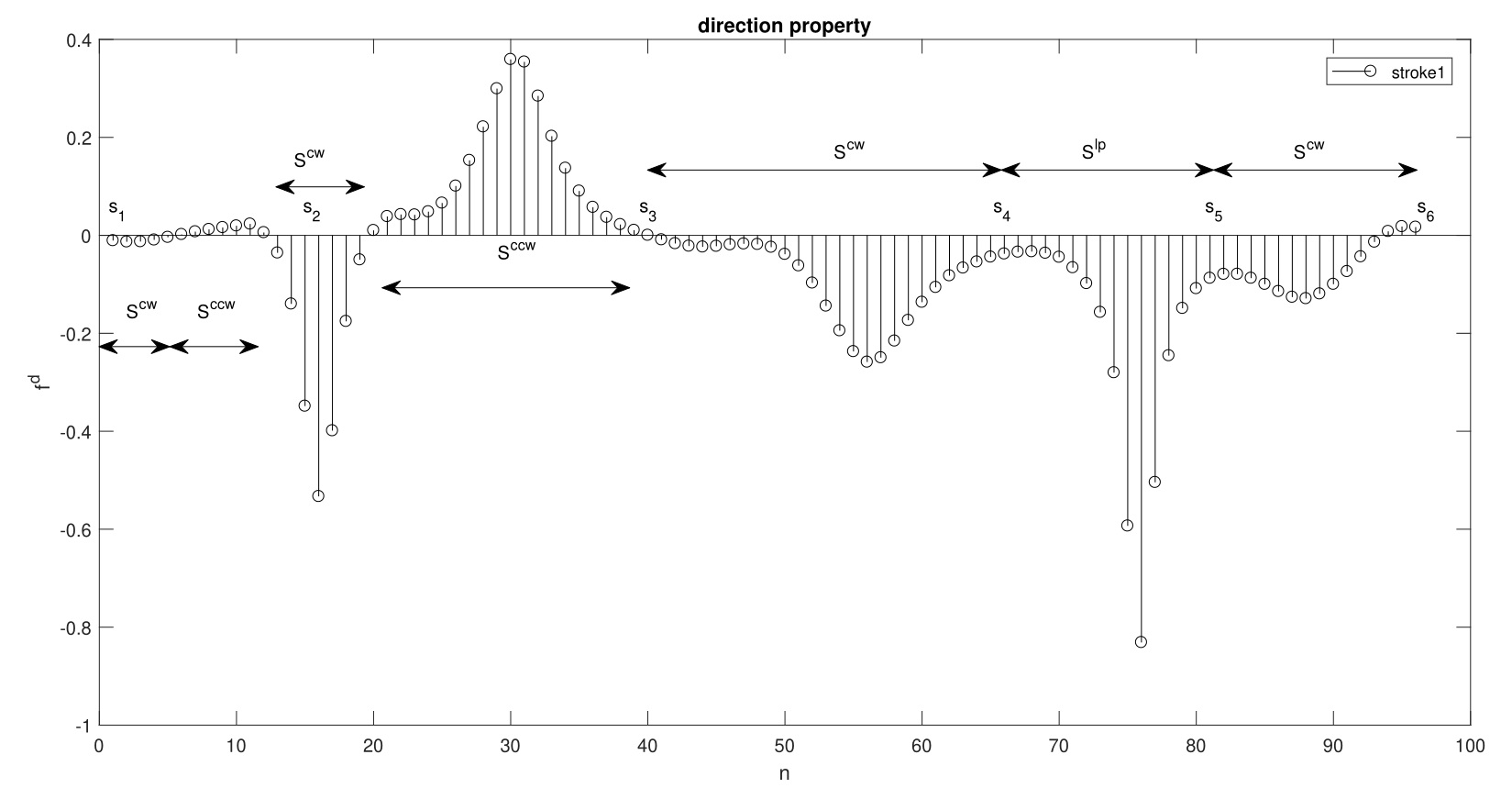}
\end{array}$
\caption{Direction property of the handwritten character in Fig. 26(a). X-axis: n, y-axis: $f^d$. The direction property of the stroke, labeled stroke 1, is shown in the region delimited by the points with indices $s_1$ and $s_6$. The end points with indices $s_1$ and $s_6$ are the segmentation points. The region delimited by $s_1$ and $s_2$, which should ideally have been produced as $S^{st}$, as in Fig. 20, is partly produced as $S^{cw}$, and partly, as $S^{ccw}$. The point with the index $s_2$ which is the segmentation point in Fig. 21 does not satisfy (6) and is part of a stroke segment $S^{cw}$, which is a pseudo sub-unit region. The point with index $s_3$ is a segmentation point as it satisfies (6). The end points of the loop stroke segment with indices $s_4$ and $s_5$ do not satisfy (7). The region delimited by $s_5$ and $s_6$, which is produced as $S^{st}$ in Fig. 21, is produced here as $S^{cw}$.}
\end{center}
\end{figure}

\begin{figure}[ht!]
\begin{center}
$\begin{array}{cc}
\includegraphics[width=1\textwidth]{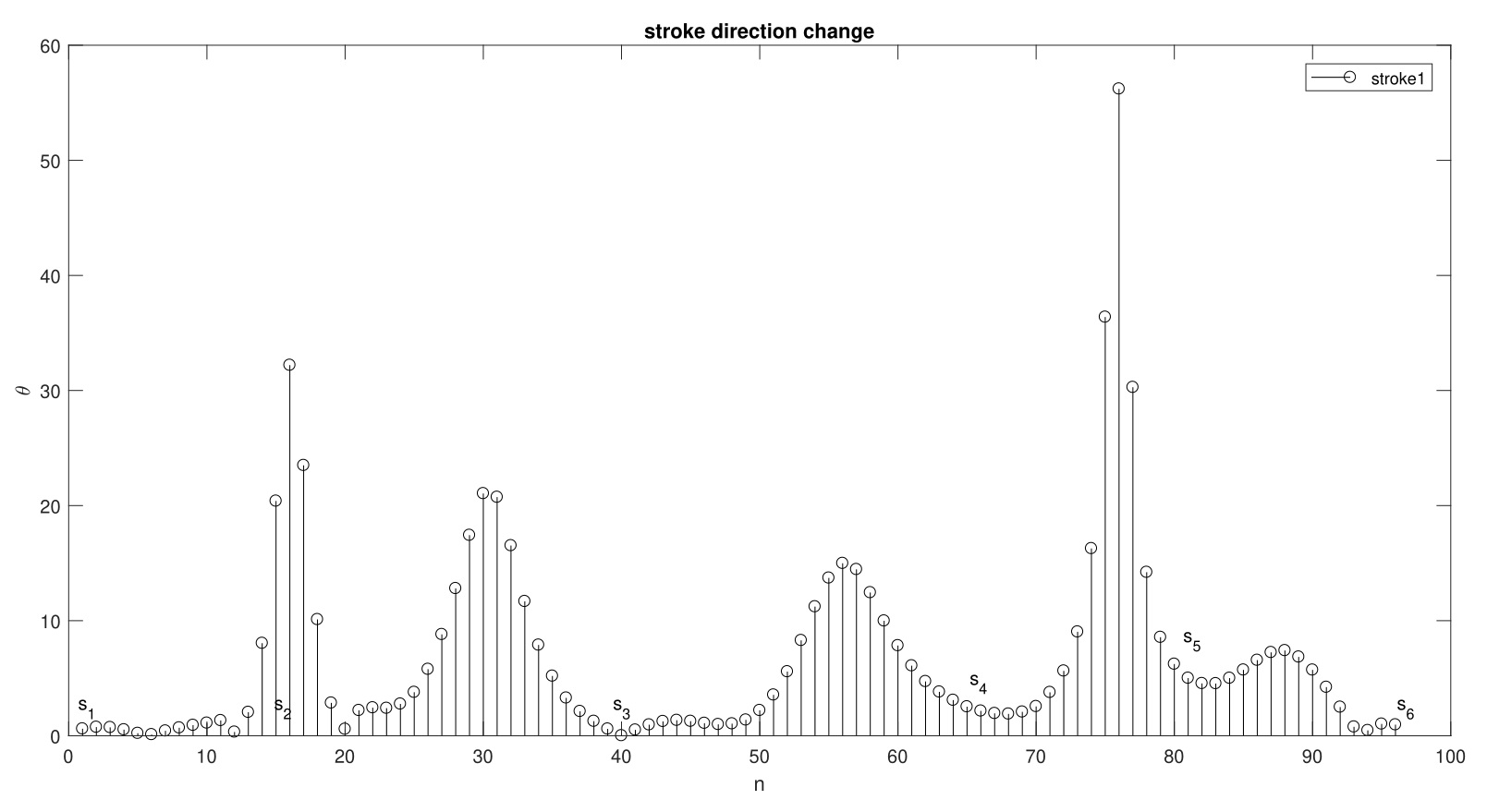} 
\end{array}$
\caption{Direction change of the handwritten character in Figure 26(a). X-axis: n, y-axis: $\theta$. The direction change of the stroke, labeled stroke 1, is shown in the region delimited by the points with indices $s_1$ and $s_6$. The point with index $s_2$ is not the segmentation point and is also not an $rldc$. The values of direction change are greater than zero at all the points in the stroke.}
\end{center}
\end{figure}

\subsection{Heuristics for sub-unit extraction}
\indent It is observed from section 5.1 and 5.2 that properties satisfied by an ideal stroke segment are not satisfied by the corresponding handwritten stroke segment. However, the handwritten stroke segment has enough structural correspondence to the ideal stroke segment and has properties that can be used to construct heuristics for extraction of sub-units from a handwritten character.\\
\indent A handwritten stroke segment which should ideally be produced as a straight stroke segment is often produced as a curve stroke segment. The values of direction property of the handwritten straight stroke segment are not zero at all the points, but are usually less in magnitude than the values of direction property of a handwritten stroke segment produced as $S^c$.
\begin{itemize}
\item Property (3) cannot be used to extract a handwritten stroke segment corresponding to an ideal straight stroke segment.
\item Direction relation between points, as defined in Section 5.4.1, rather than the direction property at points will have to be used. Direction relation between points along with threshold values $n_{cw}$ and $n_{ccw}$ determined from the character dataset are used to differentiate a handwritten curve stroke segment from a handwritten straight stroke segment.
\item Two points $p_n^{S_i}$ and $p_q^{S_i}$ are said have index proximity if, $|n-q|\le n^{ip}$. The index proximity values $n_{cw}^{ip}$ and $n_{ccw}^{ip}$ for clockwise and counter-clockwise curve stroke segments, respectively, are used in determining direction relation. These values are determined from the character data set.                
\end{itemize}
There is an $rldc$ or a pseudo sub-unit region between two consecutive sub-unit regions produced as $(S^{cw},\, S^{cw})$ or  $(S^{ccw},\, S^{ccw})$. Similarly, there is an $rldc$ or a pseudo sub-unit region between two consecutive sub-unit regions produced as $(S^c,\, S^{st})$ or $(S^{st},\, S^c)$. Usually, the length of a sub-unit is larger than the length of a pseudo sub-unit or an $rldc$. The value of $\theta_{rldc}$ for handwritten character is larger than that assumed for an ideal character.
\begin{itemize}
\item The threshold value $n_{su}^l$ for the length of a sub-unit corresponding to a curve stroke segment and the threshold value $n_{rldc}$ for $\theta_{rldc}$ are determined from the character data set.
\item If a stroke segment produced as $S^c$ has length greater than or equal to $n_{su}^l$, then it is a sub-unit region, otherwise it is a pseudo sub-unit region.
\item The locations of the middle points of a pseudo sub-unit region and  an $rldc$ are segmentation point candidates. 
\end{itemize}  
When there is no $rldc$ between two consecutive sub-unit regions produced as $(S^{cw},\, S^{ccw})$ or $(S^{ccw},\, S^{cw})$, then the region of change in direction relation between the consecutive sub-units gives the location of the segmentation point. 
\begin{itemize}
\item The segmentation point for two consecutive sub-unit regions $(S^{cw},\, S^{ccw})$ or $(S^{ccw},\, S^{cw})$, that do not have an $rldc$ between them, is determined as the midpoint between the last point of the first sub-unit and the first point of the second sub-unit.
\end{itemize}
The handwritten loop stroke segment has total direction change greater than $360^\circ$, when the direction from the last point towards the first point is considered, and is smaller than $360^\circ$, when it is not considered. Upper and lower thresholds $n_{tdc}^l$ and $n_{tdc}^u$, respectively, are used along with the total modified direction change for the determination of a loop stroke segment. Also, distance threshold $\Delta'$ is used for determining the proximity of the end points of a loop stroke segment.
\begin{itemize}
\item The modified direction change $\theta_n^{mdc(i)}$ along with $n_{mdc}$ is used in place of the direction change $\theta_n^{S_i}$. The total modified direction change $T_{n,q}^{mdc(i)}$ along with thresholds $n_{tdc}^l$ and $n_{tdc}^u$ is used in place of the total direction change $T_{n,q}^{dc}$. The values of $n_{tdc}^l$, $n_{tdc}^u$, and $n_{mdc}$ are determined from the character dataset.

\item The segmentation point for a loop stroke segment is determined based on $T_{n,q}^{mdc(i)}$ and the distance between the end points of the loop stroke segment. Verification of the loop stroke segment is done by finding the point of intersection of the segment direction at the end points of the loop stroke segment. The parameter $n_{lv}$ is used for determining the segments at the end points of the loop stroke segment. The values of $\Delta'$ and $n_{lv}$ are determined from the character dataset.
   
\end{itemize}
%\newpage
\subsection{Method for sub-unit extraction}
\indent The direction property and direction change of an ideal and the corresponding handwritten character are not same. However, the similarities in these properties allow for construction of heuristics, as explained in Section 5.3, to extract the clockwise, counter-clockwise, loop, and $rldc$ stroke segments. The extracted stroke segments and their spatial relationships are used to extract sub-units from the handwritten character. The terms used in the sub-unit extraction process are given below.       
\subsubsection{\bfseries{Definition of terms used in the sub-unit extraction process}}
Direction from the point $p_n^{S_i}$ to $p_q^{S_i}$ is\\
\[v_{n,q}^{S_i}=(p_q^{S_i}-p_n^{S_i})(||p_q^{S_i}-p_n^{S_i}||_2)^{-1},\quad n\ne q,\quad 1\leq n,q\leq N_{S_i}.\] \\
Direction relation between points $p_n^{S_i}$ and $p_q^{S_i}$ is\\
\[g_{n,q}^{d(i)} =f^d(v_{n,q+1}^{S_i},v_{q,q+1}^{S_i}),\,\,\,\mbox{for}\,\,\, n< q,\,\,\,\mbox{and}\,\,\, |g_{q,n}^{d(i)}|sign(g_{q,n}^{d(i)})\,\,\,\mbox{for}\,\,\,n>q,\,\,\, 1\leq n\leq N_{S_i}-2.\] \\
Modified direction change at $p_n^{S_i}$ is\\
\[\theta_n^{mdc(i)}=cos^{-1}({v_{n-n_{mdc},n}^{S_i}}^T\, v_{n,n+n_{mdc}}^{S_i}),\quad n_{mdc}+1\le n\le N_{S_i}-n_{mdc}.\] \\
Total modified direction change from point $p_n^{S_i}$ to $p_q^{S_i}$ is \\
\[T_{n,q}^{mdc(i)}=\sum_{l=n}^{q}\theta_l^{mdc(i)}.\] \\
Distance between the points $p_n^{S_i}$ and $p_q^{S_i}$ is \\
\[\Delta_{n,q}^i=||p_n^{S_i}-p_q^{S_i}||_2.\] 
\subsubsection{\bfseries{Clockwise curve stroke segment}}
\indent A clockwise curve stroke segment is determined based on the direction relation between the points in $S_i$. Let $g_{n,q_1}^{cw(i)(r_1)}$ be the $r_1^{th},\,\,\,r_1\ge 1$, sequence of consecutive points in $S_i$. The direction relations of the points $p_{q_1}^{S_i},\,\,\, n_1\le q_1\le n_2$, with the point $p_n^{S_i}$ are considered, $n_1$ and $n_2$ are arbitrary indices. The points in the sequence $g_{n,q_1}^{cw(i)(r_1)}$ are said to have clockwise direction relations with the point $p_n^{S_i}$ if\\[10pt]
$g_{n,q_1}^{d(i)}\le n_{cw},\quad n<q_1,\quad g_{n,q_1}^{d(i)}=|g_{q_1,n}^{d(i)}|\quad sign(g_{q_1,n}^{d(i)}),\quad n>q_1,$
\begin{equation} 1\le n\le N_{S_i}-2,\quad (n_1-n_{cw}^{ip})\le n\le(n_2+n_{cw}^{ip}).\end{equation}
$M'^{cw(i)}$ is the $N_{S_i}\times N_{S_i}$ matrix with the $n^{th}$ row corresponding to the direction relation of the point $p_n^{S_i}$ with the other points in $S_i$. 
\[M'^{cw(i)}(n,q_1)=1\,\, \mbox{if}\,\, p_{q_1}^{S_i} \,\,\mbox{is in}\,\, g_{n,q_1}^{cw(i)(r_1)},\,\, r_1\ge 1,\,\, 1\le n,q_1\le N_{S_i}.\]
\[M_n^{cw(i)}=\sum_{l=1}^nM'^{cw(i)}(n,l),\,\, 1\le n\le N_{S_i}.\]
Let $\pi_{j'_1}^{cw(i)}$ be the ${j'_1}^{th}$ sequence of indices of consecutive points in $S_i$, such that if $q_1$ is in $\pi_{j'_1}^{cw(i)}$, then 
$M_{q_1}^{cw(i)}>0,\,\, 1\le j'_1\le N'_{cw(i)}$, as shown in Fig. 29.
Here, $N'_{cw(i)}$ and $N_{\pi^{cw(i)}_{j'_1}}$ are the number of clockwise curve stroke segments and their lengths, respectively, in the stroke $S_i$ and $p_{q_1}^{S_i}$ is said to be a point in the clockwise curve stroke segment $S_{j'_1}^{cw(i)}$. The clockwise curve stroke segment is obtained as, 
\[S_{j'_1}^{cw(i)}=S_i(\pi_{j'_1}^{cw(i)},*), \quad 1\le j'_1 \le N'_{cw(i)}.\]  
\begin{figure}[ht!]
\begin{center}
$\begin{array}{cc}
\includegraphics[width=.45\textwidth]{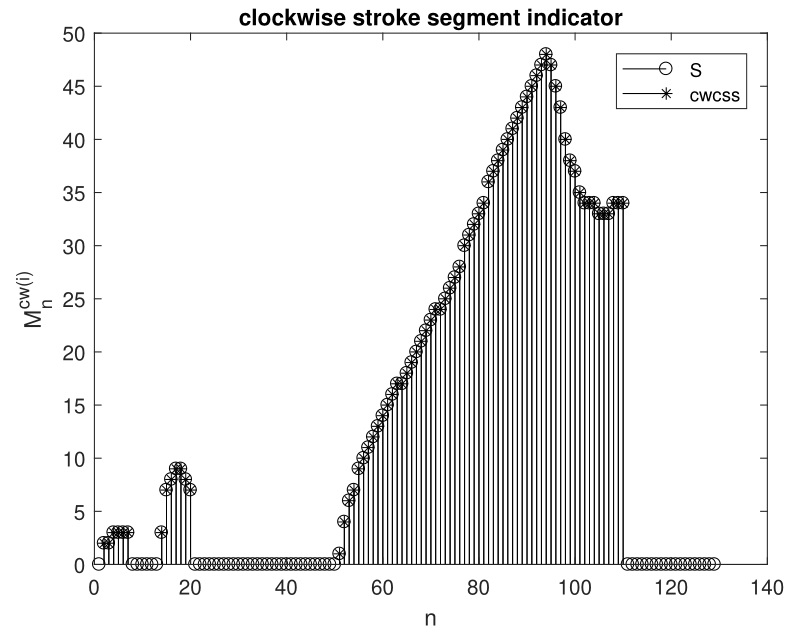}&
\includegraphics[width=.45\textwidth]{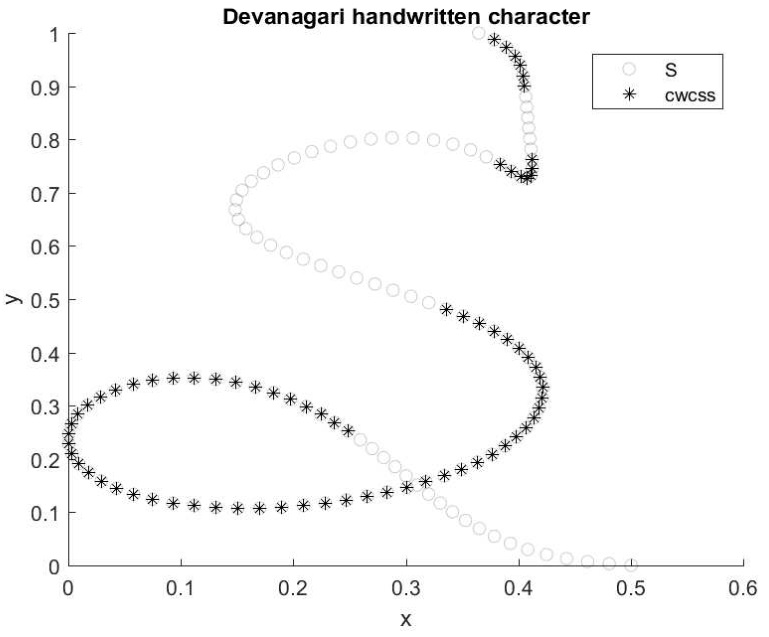}\\ 
(a) & (b)\\
\end{array}$
\caption{Extraction of clockwise curve stroke segment. (a) X-axis: n, y-axis: $M_n^{cw(i)}$. The vector $M_n^{cw(i)}$ is obtained from the direction relations of the points in the stroke using (8). The elements of the vector with values greater than zero correspond to the points belonging to the clockwise curve stroke segment. (b) The clockwise curve stroke segments correspond to the sequence of elements in $M_n^{cw(i)}$ with values greater than zero, and are shown as the segments drawn with `*' .}
\end{center}
\end{figure}
{\flushleft The algorithm for finding clockwise curve stroke segment (FCWSS) in $S_i$ is given below.}\\[15pt]
{\underline {Algorithm FCWSS}}\\
begin program\\
given: stroke $S_i$\\
use (8) to find $g_{n,q_1}^{cw(i)(r_1)},\,\, r_1\ge 1,\,\, 1\le n\le N_{S_i}-2$\\
if $r_1\ge 1$\\
find $M'^{cw(i)}$ from $g_{n,q_1}^{cw(i)(r_1)},\,\, 1\le n,q_1\le N_{S_i},\,\, r_1\ge 1$\\
$M_n^{cw(i)}=\sum_{l=1}^nM'^{cw(i)}(n,l),\,\, 1\le n\le N_{S_i}$\\
find $\pi_{j'_1}^{cw(i)}$ from $M_n^{cw(i)}$\\
$S_{j'_1}^{cw(i)}=S_i(\pi_{j'_1}^{cw(i)},*), \quad 1\le j'_1 \le N'_{cw(i)}$\\ 
end if\\
end program
\subsubsection{\bfseries{Counter-clockwise curve stroke segment}}
\indent The counter-clockwise curve stroke segment is also determined based on the direction relation between the points in $S_i$. The consecutive points in the sequence $g_{n,q_2}^{ccw(i)(r_2)},\,\, r_2\ge 1,\,\, n_1\le q_2\le n_2$, are said to have counter-clockwise direction relations with the point $p_n^{S_i}$ if\\[10pt]
$g_{n,q_2}^{d(i)}\ge n_{ccw},\quad n<q_2,\quad g_{n,q_2}^{d(i)}=|g_{q_2,n}^{d(i)}|\,\,sign(g_{q_2,n}^{d(i)}),\quad n>q_2,$
\begin{equation} 1\le n\le N_{S_i}-2,\quad (n_1-n_{ccw}^{ip})\le n\le (n_2+ n_{ccw}^{ip}).\end{equation}
$M'^{ccw(i)}$ is the $N_{S_i}\times N_{S_i}$ the matrix with the $n^{th}$ row corresponding to the direction relation of the point $p_n^{S_i}$ with the other points in $S_i$.
\[M'^{ccw(i)}(n,q_2)=1 \,\,\mbox{if}\,\, p_{q_2}^{S_i} \,\,\mbox{is in}\,\, g_{n,q_2}^{ccw(i)(r_2)},\,\, r_2\ge 1,\,\, 1\le n,q_2\le N_{S_i}.\]
\[M_n^{ccw(i)}=\sum_{l=1}^nM'^{ccw(i)}(n,l),\,\, 1\le n\le N_{S_i}.\]
Let $\pi_{j'_2}^{ccw(i)}$ be the ${j'_2}^{th}$ sequence of indices of consecutive points in $S_i$ such that, if $q_2$ is in $\pi_{j'_2}^{ccw(i)}$ then, 
$M_{q_2}^{ccw(i)}>0,\,\, 1\le j'_2\le N'_{ccw(i)}$ as shown in Fig. 30.
Here, $N'_{ccw(i)}$ and $N_{\pi^{ccw(i)}_{j'_2}}$ are the number of counter-clockwise curve stroke segments and their lengths, respectively, in the stroke $S_i$ and $p_{q_2}^{S_i}$ is said to be a point in the counter-clockwise curve stroke segment $S_{j'_2}^{ccw(i)}$. The counter-clockwise curve stroke segment is obtained as 
\[S_{j'_2}^{ccw(i)}=S_i(\pi_{j'_2}^{ccw(i)},*), \quad 1\le j'_2 \le N'_{ccw(i)}.\] 
\begin{figure}[ht!]
\begin{center}
$\begin{array}{cc}
\includegraphics[width=.45\textwidth]{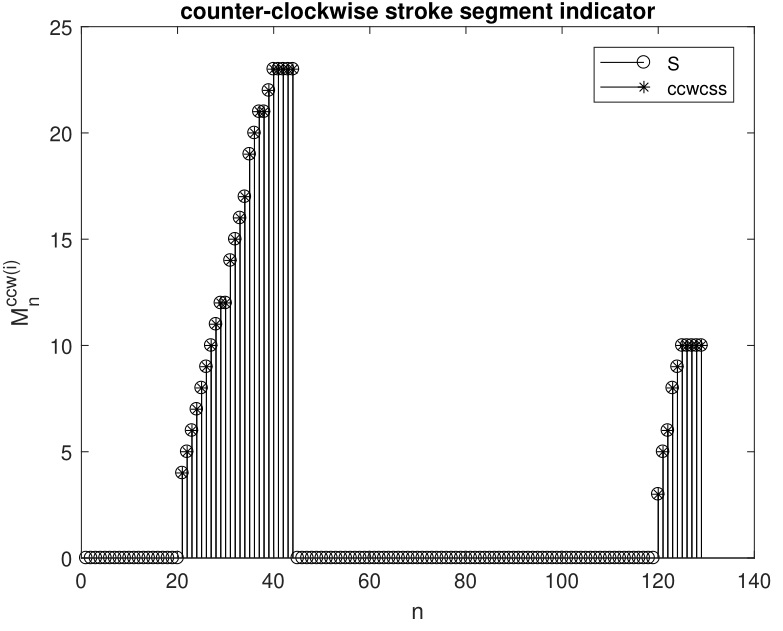}&
\includegraphics[width=.45\textwidth]{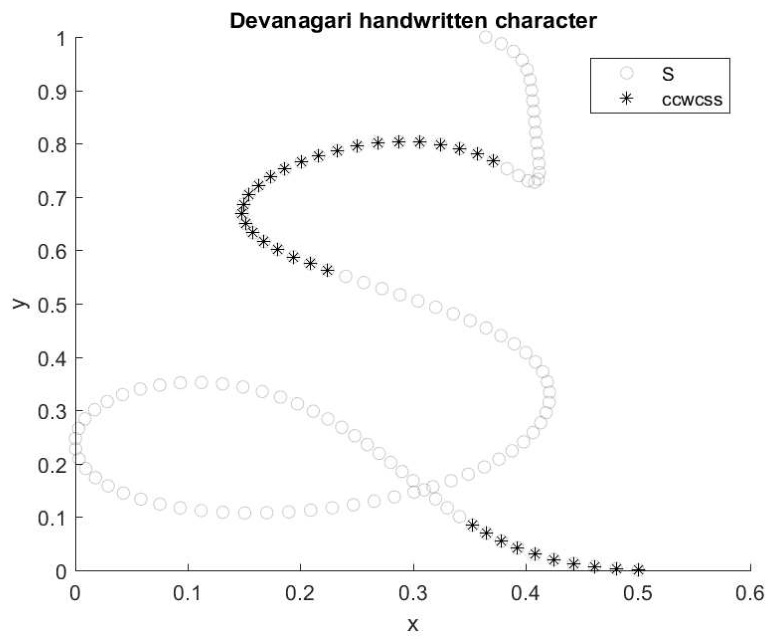}\\
(a) & (b)\\
\end{array}$
\caption{Extraction of counter-clockwise curve stroke segment. (a) X-axis: n, y-axis: $M_n^{ccw(i)}$.  The vector $M_n^{ccw(i)}$ is obtained from the direction relations of the points in the stroke using (9). The elements of the vector with values greater than zero correspond to the points belonging to the counter-clockwise curve stroke segments. (b) The counter-clockwise curve stroke segments correspond to the sequence of elements in $M_n^{ccw(i)}$ with values greater than zero, and are shown as the segments drawn with `*' .}
\end{center}
\end{figure}
{\flushleft The algorithm for finding the counter-clockwise curve stroke segment (FCCWSS) in $S_i$ is given below.} %\newpage%\\[15pt]
{\flushleft\underline{Algorithm FCCWSS}}\\
begin program\\
given: stroke $S_i$\\
use (9) to find  $g_{n,q_2}^{ccw(i)(r_2)},\,\, r_2\ge 1,\,\, 1\le n\le N_{S_i}-2$\\
if $r_2\ge 1$\\
find $M'^{ccw(i)}$ from $g_{n,q_2}^{ccw(i)(r_2)},\,\, 1\le n,q_2\le N_{S_i},\,\, r_2\ge 1$\\
$M_n^{ccw(i)}=\sum_{l=1}^nM'^{ccw(i)}(n,l),\,\, 1\le n\le N_{S_i}$\\
find $\pi_{j'_2}^{ccw(i)}$ from $M_n^{ccw(i)}$\\
$S_{j'_2}^{ccw(i)}=S_i(\pi_{j'_2}^{ccw(i)},*), \quad 1\le j'_2 \le N'_{ccw(i)}$\\ 
end if\\
end program
\subsubsection{\bfseries{Loop stroke segment}}
\indent A loop stroke segment is determined based on the distance and the total modified direction change between the end points of the stroke segment in a stroke $S_i$. A point $p_{n_3}^{S_i}$ is said to be a candidate point for the end point of the loop stroke segment, if there is a corresponding point $p_{q_3}^{S_i}$ such that,
\begin{equation}\Delta_{n_3,q_3}^i\le \Delta',\quad 360^\circ-n_{tdc}^l\le T_{n_3,q_3}^{mdc(i)}\le 360^\circ+n_{tdc}^u,\quad n_3<q_3.\end{equation}
Let $g^{lp(i)(r_3)},\,\,r_3\ge 1$, be a sequence of candidate points for the end point of the $r_3^{th}$ candidate loop stroke segment in $S_i$.
The point $p_n^{S_i}$ is an end point and the point $p_q^{S_i}$ is the corresponding other end point of the $r_3^{th}$ candidate loop stroke segment if, 
\begin{equation}(n,q)=\underset{(n_3,q_3)}{arg\,min}\,\, \Delta_{n_3,q_3}^i,\quad p_{n_3}^{S_i}\,\,\,\mbox{is in}\,\,\, g^{lp(i)(r_3)}.\end{equation}
Let $\pi_{j'_3}^{clp(i)}$ be the sequence of indices from $n$ to $q$. Then, the candidate loop stroke segment is 
\[S_{j'_3}^{clp(i)}=S_i(\pi_{j'_3}^{clp(i)},*),\,\, 1\le j'_3\le N'_{lp(i)}.\] 
$N'_{lp(i)}$ is the number of candidate loop stroke segments. \\ \indent Directions near the end points of a candidate loop stroke segment are needed to determine whether or not the candidate is a loop stroke segment. Let $p_n^{S_i}$ and $p_q^{S_i}$ be the first and the last points in $S_{j'_3}^{clp(i)}$, respectively. The directions at the points $p_{n+n_{lv}}^{S_i}$ and $p_{q-n_{lv}}^{S_i}$ are, respectively, $v_{n+n_{lv},n}^{S_i}$ and $v_{q-n_{lv},q}^{S_i}$. The slope of the segment $s_{n+n_{lv},n}$ between the points $p_n^{S_i}$ and $p_{n+n_{lv}}^{S_i}$ is, 
$s^{lp}_1=\big( p_{n}^{S_i}(2)-p_{n+n_{lv}}^{S_i}(2) \big) \big( p_{n}^{S_i}(1)-p_{n+n_{lv}}^{S_i}(1)+\epsilon \big)^{-1}$. The slope of the segment $s_{q-n_{lv},q}$ between the points $p_{q-n_{lv}}^{S_i}$ and $p_q^{S_i}$ is, 
$s^{lp}_2=\big(  p_{q}^{S_i}(2)-p_{q-n_{lv}}^{S_i}(2) \big) \big( p_{q}^{S_i}(1)-p_{q-n_{lv}}^{S_i}(1)+\epsilon \big)^{-1}$, and $\epsilon>0$ is a very small number. If the directions $v_{n+n_{lv},n}^{S_i}$ and $v_{q-n_{lv},q}^{S_i}$ along the segments $s_{n+n_{lv},n}$ and $s_{q-n_{lv},q}$, respectively, meet at the point 
\begin{equation}s_{nq}=\left(\begin{array}{c}s_x\\s_y\end{array}\right)=\left(\begin{array}{c}\frac{\big(p_q^{S_i}(2)-s^{lp}_2\,\, p_q^{S_i}(1)\big)-\big(p_n^{S_i}(2)-s^{lp}_1\,\, p_n^{S_i}(1)\big)}{s^{lp}_1-s^{lp}_2}\\s^{lp}_1\,\,\frac{\big(p_q^{S_i}(2)-s^{lp}_2\,\, p_q^{S_i}(1)\big)-\big(p_n^{S_i}(2)-s^{lp}_1\,\, p_n^{S_i}(1)\big)}{s^{lp}_1-s^{lp}_2}+\big(p_n^{S_i}(2)-s^{lp}_1\,\, p_n^{S_i}(1)  \big)\end{array}\right),\end{equation}
then the candidate loop stroke segment $S_{j'_3}^{clp(i)}$ is a loop stroke segment if it satisfies  
\begin{equation}{v_{n+n_{lv},n}^{S_i}}^T\,\,(s_{nq}-p_n^{S_i})>0\,\,\,\mbox{or}\,\,\,{v_{q-n_{lv},q}^{S_i}}^T\,\,(s_{nq}-p_q^{S_i})>0.\end{equation}
Let $\pi_{j_3}^{lp(i)}$ be the sequence of indices of consecutive points constituting the loop stroke segments. Here, $N_{lp(i)}$ and $N_{\pi^{lp(i)}_{j_3}}$ are the number of loop stroke segments and their lengths, respectively, in the stroke $S_i$.
$M'^{lp(i)}$ is the $N_{S_i}\times N_{S_i}$ matrix indicating the points in $S_i$ belonging to the loop stroke segment. 
\[M'^{lp(i)}(n,q)=1, \,\,\, \pi_{j_3}^{lp(i)}(1)\le n,q\le \pi_{j_3}^{lp(i)}(N_{\pi_{j_3}^{lp(i)}}),\,\,\, 1\le j_3 \le N_{lp(i)}. \]
\[M_n^{lp(i)}=\sum_{l=1}^nM'^{lp(i)}(n,l),\,\,\, 1\le n\le N_{S_i}.\]
If $q_3$ is in $\pi_{j_3}^{lp(i)}$, then 
$M_{q_3}^{lp(i)}>0,\,\, 1\le j_3\le N_{lp(i)}$, as shown in Fig. 31 and the corresponding loop stroke segment is 
\[S_{j_3}^{lp(i)}=S_i(\pi_{j_3}^{lp(i)},*),\,\,\,1\le j_3\le N_{lp(i)}.\]
\begin{figure}[ht!]
\begin{center}
$\begin{array}{cc}
\includegraphics[width=.45\textwidth]{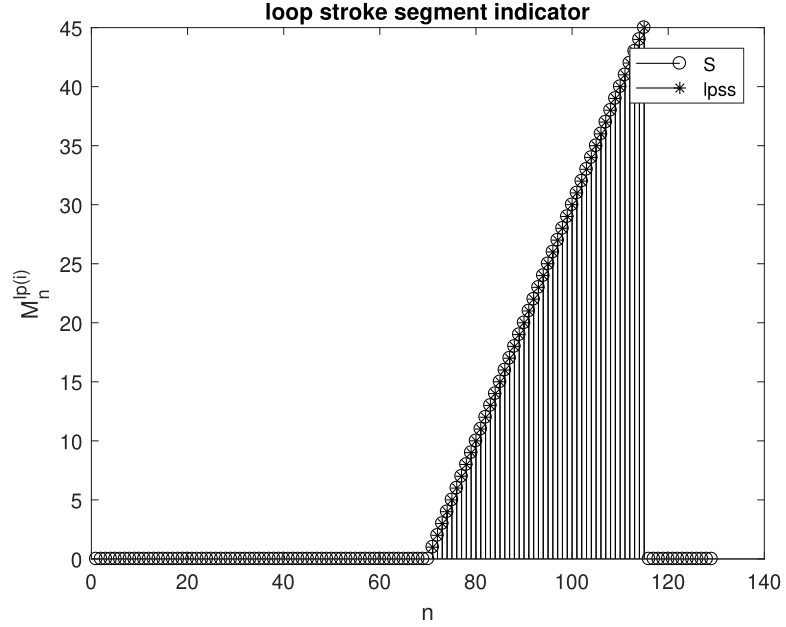}&
\includegraphics[width=.45\textwidth]{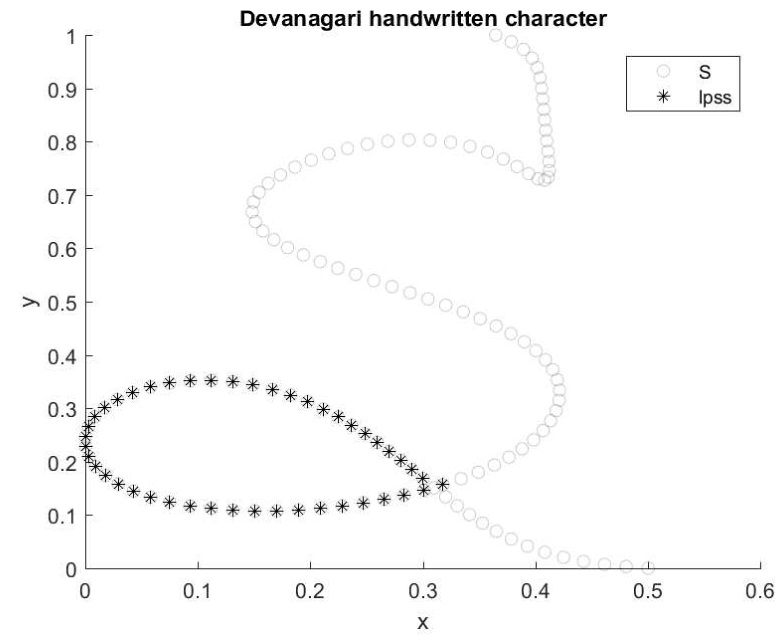} \\
(a) & (b)\\
\end{array}$
\caption{Extraction of the loop stroke segment. (a) X-axis: n, y-axis: $M_n^{lp(i)}$. The vector $M_n^{lp(i)}$ is obtained from the distance and total modified direction change between the end points of the loop stroke segment. The elements of the vector with values greater than zero correspond to the points belonging to the loop stroke segment. (b) The loop stroke segment corresponds to the sequence of elements in $M_n^{lp(i)}$ with values greater than zero, and is shown as the segment drawn with `*'.}
\end{center}
\end{figure}
{\flushleft The algorithm for finding a loop stroke segment (FLSS) in $S_i$ is given below.}\\[15pt]
{\underline{Algorithm FLSS}}\\
begin program\\
given: stroke $S_i$\\
use (10) and (11) to find $\pi_{j'_3}^{clp(i)}$\\
if$_1$ $N'_{lp(i)}\ge 1$ \\
$S_{j'_3}^{clp(i)}=S_i(\pi_{j'_3}^{clp(i)},*),\,\, 1\le j'_3\le N'_{lp(i)}$\\
use (12) and (13) to find $\pi_{j_3}^{lp(i)}$ from $S_{j'_3}^{clp(i)}$\\
if$_2$ $N_{lp(i)}\ge 1$\\
$S_{j_3}^{lp(i)}=S_i(\pi_{j_3}^{lp(i)},*),\,\,\,1\le j_3\le N_{lp(i)}$\\
end if$_2$\\
end if$_1$\\
end program
\subsubsection{\bfseries{Region of large direction change}}
A region of large direction change in a stroke $S_i$ is determined by finding the locations in $S_i$, where the modified direction change is greater than $n_{rldc}$. Let $\pi_{j'_4}^{rldc(i)},\,\,\, 1\le j'_4\le N'_{rldc(i)}$, be the sequences of indices of consecutive points in $S_i$ such that, if $q$ is in $\pi_{j'_4}^{rldc(i)}$ then,
\begin{equation}\theta_q^{mdc(i)}\ge n_{rldc}.\end{equation}
Here, $N'_{rldc(i)}$ and $N_{\pi^{rldc(i)}_{j'_4}}$ are the number of such sequences and their lengths, respectively, in the stroke $S_i$ . The region of large direction change is
\[S_{j'_4}^{rldc(i)}=S_i(\pi_{j'_4}^{rldc(i)},*),\,\,\, 1\le j'_4\le N'_{rldc(i)}.\]
\begin{figure}[ht!]
\begin{center}
$\begin{array}{cc}
\includegraphics[width=.45\textwidth]{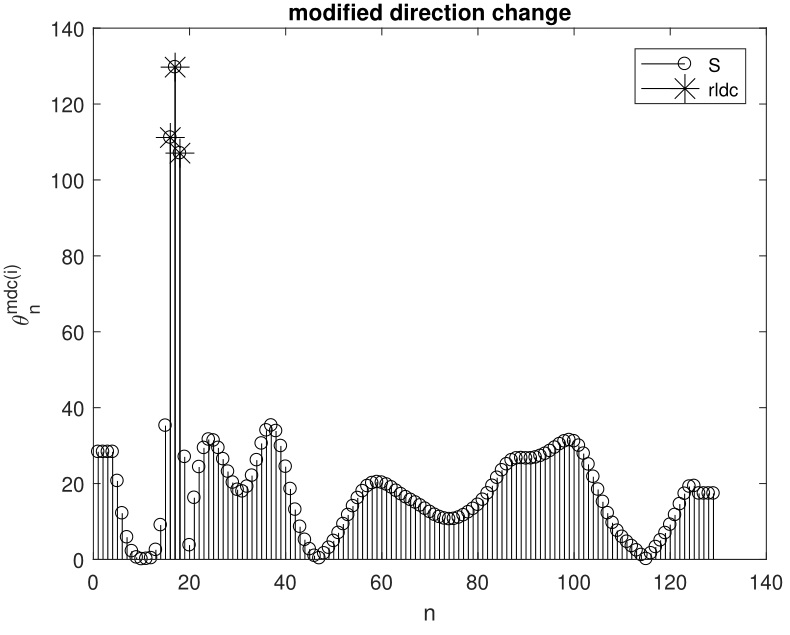}&
\includegraphics[width=.45\textwidth]{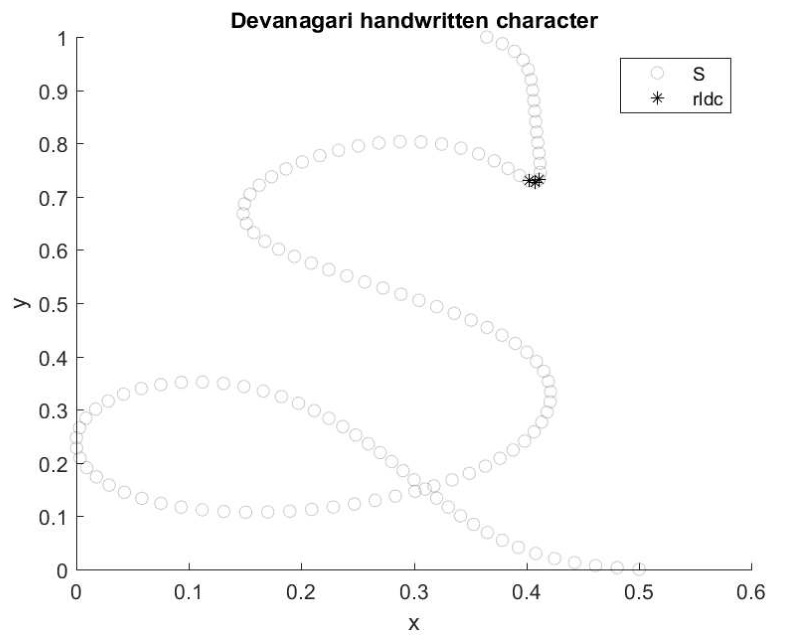}\\
(a) & (b)\\
\end{array}$
\caption{Extraction of the region of large direction change. (a) X-axis: n, y-axis: $\theta_n^{mdc(i)}$.  The region of large direction change $rldc$ is the region in the stroke, where the modified direction change $\theta_n^{mdc(i)}$ is greater than the threshold $n_{rldc}=105^{\circ}$, and is obtained using (14). (b) The region of large direction change is shown as the segment drawn with `*'.}
\end{center}
\end{figure}
{\flushleft The algorithm for finding the region of large direction change (FRLDC) in $S_i$ is given below.}\\[15pt]
{\flushleft\underline{Algorithm FRLDC}}\\
begin program\\
given: stroke $S_i$\\
use (14) to find $\pi_{j'_4}^{rldc(i)}$\\
if $N'_{rldc(i)}\ge 1$\\
$S_{j'_4}^{rldc(i)}=S_i(\pi_{j'_4}^{rldc(i)},*),\,\,\, 1\le j'_4\le N'_{rldc(i)}$.\\
end if\\
end program

\subsubsection{\bfseries{Removal and merger of curve stroke segments}}
\indent The clockwise curve stroke segments, counter-clockwise curve stroke segments, loop stroke segments, and regions of large direction change determined above may have overlapping regions. If $S_{j'_1}^{cw(i)}$ and $S_{j'_2}^{ccw(i)}$ have lengths greater than or equal to $n_{su}^l$, then they are considered as sub-units, otherwise as pseudo sub-units. These curve stroke segments are merged or removed based on their relations with the neighboring stroke segments as given below.
%\[\mbox{\bfseries{Merger of $S^{cw(i)}$}}\]
\begin{flalign}
\mbox{\bfseries{Merger of $S^{cw(i)}$}}
\end{flalign}
Let a curve stroke segment $S_{j'_1}^{cw(i)}$ be a sub-unit region. If there is no $rldc$ or $S_{j'_2}^{ccw(i)}$ between $S_{j'_1}^{cw(i)}$ and its similar immediate neighbor $S_{j'_1-1}^{cw(i)}$  or $S_{j'_1+1}^{cw(i)}$, then $S_{j'_1}^{cw(i)}$ is merged with its immediate neighbor.
\begin{equation}\mbox{\bfseries{Merger of $S^{ccw(i)}$}}\end{equation}
Let a curve stroke segment $S_{j'_2}^{ccw(i)}$ be a sub-unit region. If there is no $rldc$ or $S_{j'_1}^{cw(i)}$ between $S_{j'_2}^{ccw(i)}$ and its immediate neighbor $S_{j'_2-1}^{ccw(i)}$ or $S_{j'_2+1}^{ccw(i)}$, then $S_{j'_2}^{ccw(i)}$ is merged with its immediate neighbor.
\begin{equation}\mbox{\bfseries{Removal of $S^{cw(i)}$ and $S^{ccw(i)}$}}\end{equation}
If $S_{j'_4}^{rldc(i)}$ and the pseudo sub-unit region $S_{j'_1}^{cw(i)}$ or $S_{j'_2}^{ccw(i)}$ overlap, then the pseudo sub-unit region is removed.
\begin{equation}\mbox{\bfseries{Removal of $S^{cw(i)}$, $S^{ccw(i)}$ and $S^{rldc(i)}$}}\end{equation}
If the pseudo sub-unit region $S_{j'_1}^{cw(i)}$ or $S_{j'_2}^{ccw(i)}$ or the $rldc$ $S_{j'_4}^{rldc(i)}$ is contained in a loop stroke segment, then the pseudo sub-unit region or the $rldc$ is removed.\\[15 pt]
Let $\pi_{j_1}^{cw(i)},\,\,\, 1\le j_1\le N_{cw(i)}$, be the sequences of indices of consecutive points in $S_i$ obtained after merger and removal of clockwise curve stroke segments. The clockwise curve stroke segments are \[S_{j_1}^{cw(i)}=S_i(\pi_{j_1}^{cw(i)},*),\,\,\ 1\le j_1\le N_{cw(i)}.\]   
Let $\pi_{j_2}^{ccw(i)},\,\,\, 1\le j_2\le N_{ccw(i)}$, be the sequences of indices of consecutive points in $S_i$ obtained after merger and removal of counter-clockwise curve stroke segments. The counter-clockwise curve stroke segments are\[S_{j_2}^{ccw(i)}=S_i(\pi_{j_2}^{ccw(i)},*),\,\,\ 1\le j_2\le N_{ccw(i)}.\] 
Let $\pi_{j_4}^{rldc(i)},\,\,\, 1\le j_4\le N_{rldc(i)}$, be the sequences of indices of consecutive points in $S_i$ obtained after removal of regions of large direction change. The regions of large direction change are\[S_{j_4}^{rldc(i)}=S_i(\pi_{j_4}^{rldc(i)},*),\,\,\ 1\le j_4\le N_{rldc(i)}.\]   
{\flushleft The algorithm for removal and merger of curve stroke segments (RMCSS) in $S_i$ is given below.}\\[15pt]
{\underline{Algorithm RMCSS}}\\
begin program\\
given: stroke $S_i$\\
use (15) to merge $S_{j'_1}^{cw(i)}$ with $S_{j'_1-1}^{cw(i)}$, $S_{j'_1+1}^{cw(i)}$\\
use (16) to merge $S_{j'_2}^{ccw(i)}$ with $S_{j'_2-1}^{ccw(i)}$, $S_{j'_2+1}^{ccw(i)}$\\
use (17) to remove pseudo sub-unit regions $S_{j'_1}^{cw(i)}$ and $S_{j'_2}^{ccw(i)}$\\
use (18) to remove $S_{j'_1}^{cw(i)}$, $S_{j'_2}^{ccw(i)}$, and $S_{j'_4}^{rldc(i)}$\\
find $\pi_{j_1}^{cw(i)},\,\,\, 1\le j_1\le N_{cw(i)}$ after above steps\\
if $N_{cw(i)}\ge 1$\\

$S_{j_1}^{cw(i)}=S_i(\pi_{j_1}^{cw(i)},*),\,\,\ 1\le j_1\le N_{cw(i)}$ \\
end if\\
find $\pi_{j_2}^{ccw(i)},\,\,\, 1\le j_2\le N_{ccw(i)}$ after above steps\\
if $N_{ccw(i)}\ge 1$\\
$S_{j_2}^{ccw(i)}=S_i(\pi_{j_2}^{ccw(i)},*),\,\,\ 1\le j_2\le N_{ccw(i)}$\\
end if\\
find $\pi_{j_4}^{rldc(i)},\,\,\, 1\le j_4\le N_{rldc(i)}$ after above steps\\
if $N_{rldc(i)}\ge 1$\\
$S_{j_4}^{rldc(i)}=S_i(\pi_{j_4}^{rldc(i)},*),\,\,\ 1\le j_4\le N_{rldc(i)}$\\
end if\\
end program

\subsubsection{\bfseries{Sub-unit extraction}}
Let $S_j^{su(i)},\,\,\, 1\le j\le N_{S^{su(i)}}$, be the sequence of stroke segments, $S_{j_1}^{cw(i)}$, $S_{j_2}^{ccw(i)}$, $S_{j_3}^{lp(i)}$, and $S_{j_4}^{rldc(i)}$ obtained by ordering them according to their locations, from first to last, in the stroke $S_i$. Here, the number of segments in $S_i$ is given by
\[N_{S^{su(i)}}=N_{cw(i)}+N_{ccw(i)}+N_{lp(i)}+N_{rldc(i)}.\]
Two consecutive stroke segments $S_j^{su(i)}$ and $S_{j+1}^{su(i)}$ are considered for determination of a segmentation point. If $S_{j}^{su(i)}$ in the pair is a pseudo sub-unit region or a region of large direction change already considered, then the next two consecutive stroke segments $S_{j+1}^{su(i)}$ and $S_{j+2}^{su(i)}$ are considered for the determination of the segmentation point. Here, the consecutive segmentation points $\pi_m^i,\,\,\, 1<m<N_{\pi^i}$, are dependent on the sequence of pair of stroke segments, where the first stroke segment in the pair has not been used before for segmentation.
\begin{equation}\mbox{{\bfseries{$S_j^{su(i)}$ is a sub-unit region $S_{j_1}^{cw(i)}$:}}}\end{equation}
If $S_{j+1}^{su(i)}$ is a pseudo sub-unit region $S_{j_2}^{ccw(i)}$ then $\pi_{m}^i=\left\lfloor 0.5\,\big(\pi_{j_2}^{ccw(i)}(1)+\pi_{j_2}^{ccw(i)}(N_{\pi_{j_2}^{ccw(i)}})\big)\right\rfloor$.\\ 
If $S_{j+1}^{su(i)}$ is a sub-unit region $S_{j_2}^{ccw(i)}$ then $\pi_{m}^i=\left\lfloor 0.5\,\big(\pi_{j_1}^{cw(i)}(N_{\pi_{j_1}^{cw(i)}})+\pi_{j_2}^{ccw(i)}(1)\big)\right\rfloor$.\\
If $S_{j+1}^{su(i)}$ is $S_{j_3}^{lp(i)}$ then $\pi_{m}^i=\pi_{j_3}^{lp(i)}(1)$.\\
If $S_{j+1}^{su(i)}$ is $S_{j_4}^{rldc(i)}$ then $\pi_{m}^i=\left\lfloor 0.5\,\big(\pi_{j_4}^{rldc(i)}(1)+\pi_{j_4}^{rldc(i)}(N_{\pi_{j_4}^{rldc(i)}})\big)\right\rfloor$. \\
\begin{equation}\mbox{{\bfseries{$S_j^{su(i)}$ is a pseudo sub-unit region $S_{j_1}^{cw(i)}$:}}}\end{equation}
If $S_{j+1}^{su(i)}$ is a pseudo sub-unit region $S_{j_2}^{ccw(i)}$,  $N_{\pi_{j_1}^{cw(i)}}>N_{\pi_{j_2}^{ccw(i)}}$, $\pi_{m}^i=\left\lfloor 0.5\,\big(\pi_{j_2}^{ccw(i)}(1)+\pi_{j_2}^{ccw(i)}(N_{\pi_{j_2}^{ccw(i)}})\big)\right\rfloor$.\\
If $S_{j+1}^{su(i)}$ is a pseudo sub-unit region $S_{j_2}^{ccw(i)}$,  $N_{\pi_{j_1}^{cw(i)}}<=N_{\pi_{j_2}^{ccw(i
)}}$, $\pi_{m}^i=\left\lfloor 0.5\,\big(\pi_{j_1}^{cw(i)}(1)+\pi_{j_1}^{cw(i)}(N_{\pi_{j_1}^{cw(i)}})\big)\right\rfloor$.\\ 
If $S_{j+1}^{su(i)}$ is a sub-unit region $S_{j_2}^{ccw(i)}$ then $\pi_{m}^i=\left\lfloor 0.5\,\big(\pi_{j_1}^{cw(i)}(1)+\pi_{j_1}^{cw(i)}(N_{\pi_{j_1}^{cw(i)}})\big)\right\rfloor$.\\
If $S_{j+1}^{su(i)}$ is $S_{j_3}^{lp(i)}$ then $\pi_{m}^i=\left\lfloor 0.5\,\big(\pi_{j_1}^{cw(i)}(1)+\pi_{j_1}^{cw(i)}(N_{\pi_{j_1}^{cw(i)}})\big)\right\rfloor$.\\
If $S_{j+1}^{su(i)}$ is $S_{j_4}^{rldc(i)}$ then $\pi_{m}^i=\left\lfloor 0.5\,\big(\pi_{j_1}^{cw(i)}(1)+\pi_{j_1}^{cw(i)}(N_{\pi_{j_1}^{cw(i)}})\big)\right\rfloor$.\\ 
\begin{equation}\mbox{{\bfseries{$S_j^{su(i)}$ is a sub-unit region $S_{j_2}^{ccw(i)}$:}}}\end{equation}
If $S_{j+1}^{su(i)}$ is a pseudo sub-unit region $S_{j_1}^{cw(i)}$ then $\pi_{m}^i=\left\lfloor 0.5\,\big(\pi_{j_1}^{cw(i)}(1)+\pi_{j_1}^{cw(i)}(N_{\pi_{j_1}^{cw(i)}})\big)\right\rfloor$.\\ 
If $S_{j+1}^{su(i)}$ is a sub-unit region $S_{j_1}^{cw(i)}$ then $\pi_{m}^i=\left\lfloor 0.5\,\big(\pi_{j_2}^{ccw(i)}(N_{\pi_{j_2}^{ccw(i)}})+\pi_{j_1}^{cw(i)}(1)\big)\right\rfloor$.\\
If $S_{j+1}^{su(i)}$ is $S_{j_3}^{lp(i)}$ then $\pi_{m}^i=\pi_{j_3}^{lp(i)}(1)$.\\
If $S_{j+1}^{su(i)}$ is $S_{j_4}^{rldc(i)}$ then $\pi_{m}^i=\left\lfloor 0.5\,\big(\pi_{j_4}^{rldc(i)}(1)+\pi_{j_4}^{rldc(i)}(N_{\pi_{j_4}^{rldc(i)}})\big)\right\rfloor$. \\
\begin{equation}\mbox{{\bfseries{$S_j^{su(i)}$ is a pseudo sub-unit region $S_{j_2}^{ccw(i)}$:}}}\end{equation}
If $S_{j+1}^{su(i)}$ is a pseudo sub-unit region $S_{j_1}^{cw(i)}$, $N_{\pi_{j_2}^{ccw(i)}}>N_{\pi_{j_1}^{cw(i)}}$, $\pi_{m}^i=\left\lfloor 0.5\,\big(\pi_{j_1}^{cw(i)}(1)+\pi_{j_1}^{cw(i)}(N_{\pi_{j_1}^{cw(i)}})\big)\right\rfloor$.\\
If $S_{j+1}^{su(i)}$ is a pseudo sub-unit region $S_{j_1}^{cw(i)}$, $N_{\pi_{j_2}^{ccw(i)}}<=N_{\pi_{j_1}^{cw(i)}}$, $\pi_{m}^i=\left\lfloor 0.5\,\big(\pi_{j_2}^{ccw(i)}(1)+\pi_{j_2}^{ccw(i)}(N_{\pi_{j_2}^{ccw(i)}})\big)\right\rfloor$.\\ 
If $S_{j+1}^{su(i)}$ is a sub-unit region $S_{j_1}^{cw(i)}$ then $\pi_{m}^i=\left\lfloor 0.5\,\big(\pi_{j_2}^{ccw(i)}(1)+\pi_{j_2}^{ccw(i)}(N_{\pi_{j_2}^{ccw(i)}})\big)\right\rfloor$.\\
If $S_{j+1}^{su(i)}$ is $S_{j_3}^{lp(i)}$ then $\pi_{m}^i=\left\lfloor 0.5\,\big(\pi_{j_2}^{ccw(i)}(1)+\pi_{j_2}^{ccw(i)}(N_{\pi_{j_2}^{ccw(i)}})\big)\right\rfloor$.\\
If $S_{j+1}^{su(i)}$ is $S_{j_4}^{rldc(i)}$ then $\pi_{m}^i=\left\lfloor 0.5\,\big(\pi_{j_2}^{ccw(i)}(1)+\pi_{j_2}^{ccw(i)}(N_{\pi_{j_2}^{ccw(i)}})\big)\right\rfloor$.\\ 
\begin{equation}\mbox{{\bfseries{$S_j^{su(i)}$ is a loop stroke segment $S_{j_3}^{lp(i)}$:}}}\end{equation}
$\pi_{m}^i=\pi_{j_3}^{lp(i)}(N_{\pi_{j_3}^{lp(i)}}+1)$.\\ 
\begin{equation}\mbox{{\bfseries{$S_j^{su(i)}$ is a region of large direction change $S_{j_4}^{rldc(i)}$:}}}\end{equation}
$\pi_{m}^i=\left\lfloor 0.5\,\big(\pi_{j_4}^{rldc(i)}(1)+\pi_{j_4}^{rldc(i)}(N_{\pi_{j_4}^{rldc(i)}})\big)\right\rfloor$.\[\] 
Once the segmentation points have been determined, the sub-unit regions, $S_m^{u(i)},\,\,\, 1\le m\le N_{\pi^i}-1$, are extracted as, 
\[S_m^{u(i)}=\begin{cases}S_i(\pi_m^i:\pi_{m+1}^i-1,*),\quad 1\le m\le N_{\pi^i}-2\\ 
S_i(\pi_m^i:\pi_{m+1}^i,*), \quad m=N_{\pi^i}-1\\  \end{cases}.\]
{\flushleft The algorithm for sub-unit extraction (SUE) in $S_i$ is given below.}\\[15pt]
{\underline{Algorithm SUE}}\\
begin program\\
given: stroke $S_i$\\
use FCWSS to find clockwise curve stroke segments $S_{j'_1}^{cw(i)},\,\,1\le j'_1\le N'_{cw(i)}$\\
use FCCWSS to find counter-clockwise curve stroke segments $S_{j'_2}^{ccw(i)},\,\,1\le j'_2\le N'_{ccw(i)}$\\
use FLSS to find loop stroke segments $S_{j_3}^{lp(i)},\,\,1\le j_3\le N_{lp(i)}$\\
use FRLDC to find region of large direction change $S_{j'_4}^{rldc(i)},\,\,1\le j'_4\le N'_{rldc(i)}$\\
use RMCSS for removal and merger of curve stroke segments to get $S_{j_1}^{cw(i)}$, $S_{j_2}^{ccw(i)}$, and $S_{j_4}^{rldc(i)}$\\
collect $S_{j_1}^{cw(i)}$, $S_{j_2}^{ccw(i)}$, $S_{j_3}^{lp(i)}$, and $S_{j_4}^{rldc(i)}$ in $S_j^{su(i)}$ according to their occurrences in $S_i$\\
Determine segmentation points $\pi_m^i,\,\,\, 1\le m\le N_{\pi^i}$, using (19), (20), (21), (22), (23), and (24)\\
extract sub-units $S_m^{u(i)},\,\,\, 1\le m\le N_{\pi^i}-1$, from $S_i$ using $\pi_m^i$ \\
end program
\section{Experiments and results}
\indent The dataset for extraction of sub-units and assessment of importance of local representation of characters in terms of sub-units consists of samples of 96 different character classes. There are 12832 samples in the training set and 2821 samples in the testing set. The sub-units are extracted from handwritten characters from both the training and testing datasets.\\
\indent The extracted sub-units from some characters in the dataset are shown in the figure below. Sub-unit based classifier (SUB) developed in Sharma \cite{hpod} is used to model a character locally in terms of features extracted from the sub-units and globally in terms of features extracted from the character as a whole. The performance of SUB classifier is compared to that of the classifiers trained only with the global representations of characters. The performances of SUB and other classifiers on the test dataset are given in the table 1.
\subsection{Sub-unit extraction examples}
Different threshold values defined in the construction of heuristics are: (1) $\tau_{cw}=-0.1$, (2) $\tau_{ccw}=0.1$, (3) $\tau_{cw}^{ip}=6$, (4) $\tau_{ccw}^{ip}=6$, (5) $\tau_{su}^l=14$,  (6) $\tau_{rldc}=105^o$, (7) $\tau_{tdc}^l=180$, (8) $\tau_{tdc}^u=0$, (9) $\tau_{\Delta}=0.04$, (10) $\tau_{mdc}=3$, (11) $\tau_{lv}=5$. \\ \\
Figure 33 shows a Hindi handwritten character corresponding to the ideal character in Fig. 17. All the sub-units extracted from this character are sub-unit structures of the corresponding ideal character.
\begin{figure}[ht!]
\begin{center}
$\begin{array}{c}
\includegraphics[width=1\textwidth]{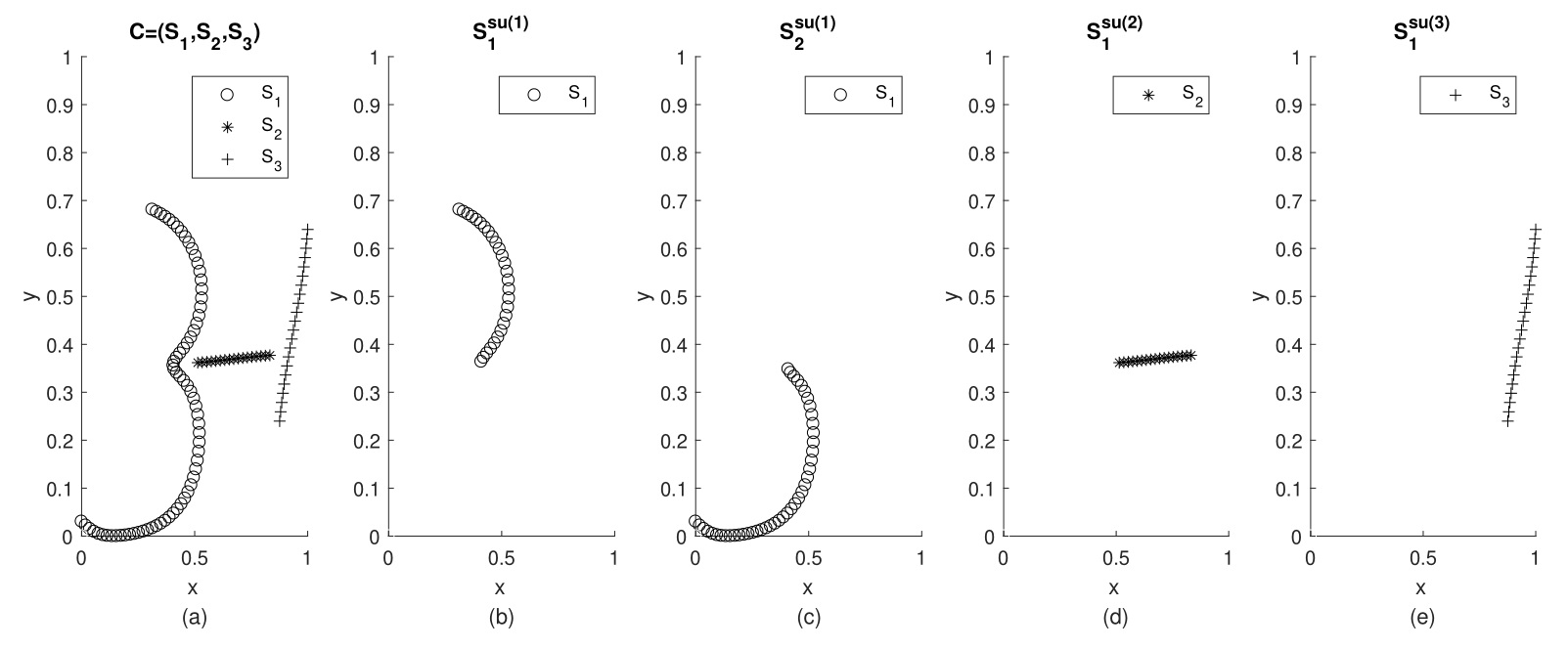}
\end{array}$
\caption{Illustration of extraction of sub-units from a handwritten character. (a) The handwritten character corresponding to the ideal online character in Fig. 17(a). This handwritten character is produced using three strokes $S_1$, $S_2$, and $S_3$. The stroke $S_1$ has two sub-units. Both the strokes $S_2$ and $S_3$ are single sub-unit strokes. (b) $S_1^{su(1)}$ is the first sub-unit in the first stroke and is similar to the first sub-unit in the first stroke of the corresponding ideal character. (c) $S_2^{su(1)}$ is the second sub-unit in the first stroke and is similar to the second sub-unit in the first stroke of the corresponding ideal character. (d) $S_1^{su(2)}$ is the sub-unit in the second stroke and is similar to the sub-unit in the second stroke of the corresponding ideal character. (e) $S_1^{su(3)}$ is the sub-unit in the third stroke and is similar to the sub-unit in the third stroke of the corresponding ideal character.}
\end{center}
\end{figure}
Figure 34 shows the same ideal character in Fig. 17 but written differently. All the sub-units extracted from this character are not sub-unit structures of the corresponding ideal character. The second sub-unit of the first stroke in this character is a pseudo sub-unit but is extracted as sub-unit because its length is greater than $\tau_{su}^l$. The second sub-unit of the second stroke in this character is a loop stroke segment. Loop stroke segment in this character is not a sub-unit structure in the corresponding ideal character but is produced becasue of the way the character is generated.  

\begin{figure}[ht!]
\begin{center}
$\begin{array}{c}
\includegraphics[width=1\textwidth]{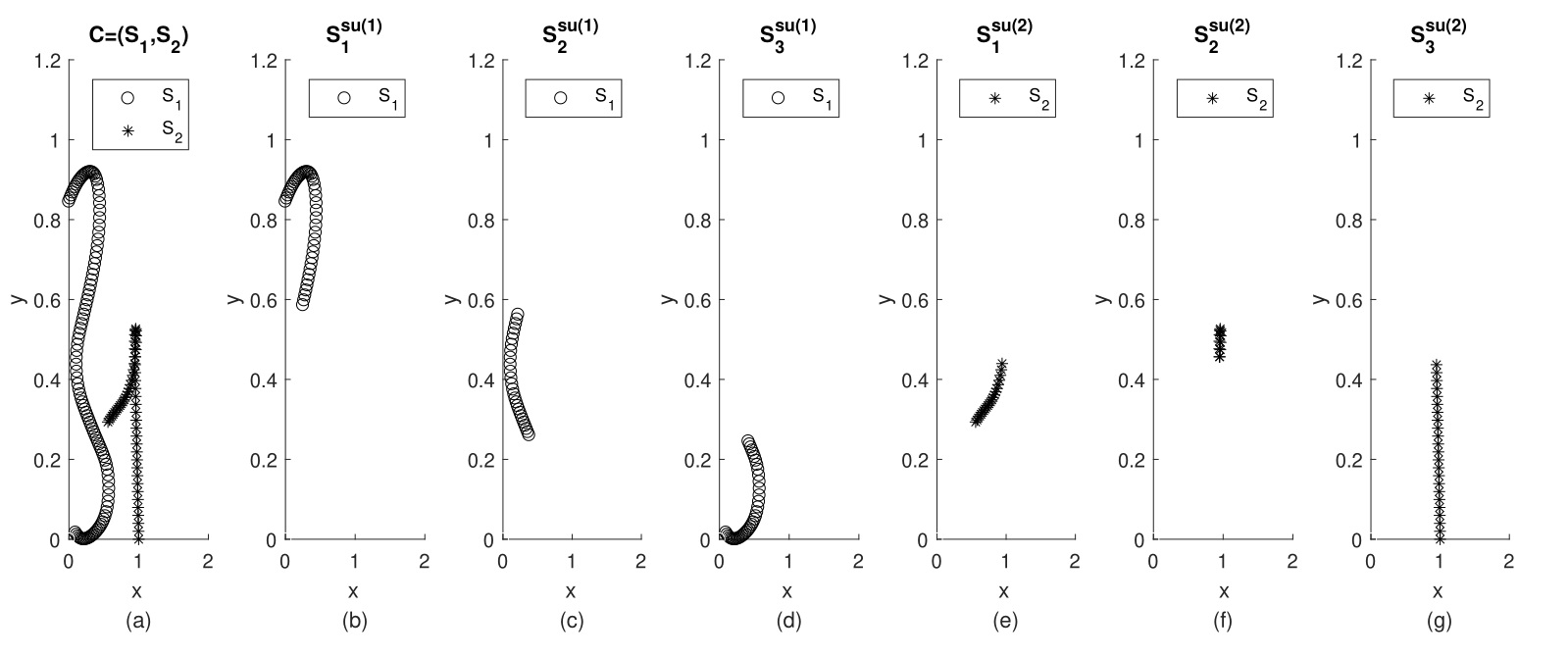}
\end{array}$
\caption{Illustration of extraction of sub-units from a handwritten character. (a) The handwritten character corresponding to the ideal online character in Fig. 17(a). This handwritten character is produced using two strokes $S_1$ and $S_2$. Both the strokes $S_1$ and $S_2$ have three sub-units. (b) $S_1^{su(1)}$ is similar to the first sub-unit in the first stroke of the corresponding ideal character. (c) $S_2^{su(1)}$ is not similar to any sub-unit in the corresponding ideal character. (d) $S_3^{su(1)}$ is similar to the second sub-unit in the corresponding ideal character. (e) $S_1^{su(2)}$ has some similarity to the sub-unit in the second stroke of the corresponding ideal character. (f) $S_2^{su(2)}$ is not similar to any sub-unit in the corresponding ideal character. (g) $S_3^{su(2)}$ is similar to the sub-unit in the third stroke of the corresponding ideal character.}
\end{center}
\end{figure}

\flushleft Figure 35 shows Hindi handwritten character corresponding to the ideal character in Fig. 20. All the sub-units extracted from this character are sub-unit structure of the corresponding ideal character.
\begin{figure}[ht!]
\begin{center}
$\begin{array}{c}
\includegraphics[width=1\textwidth]{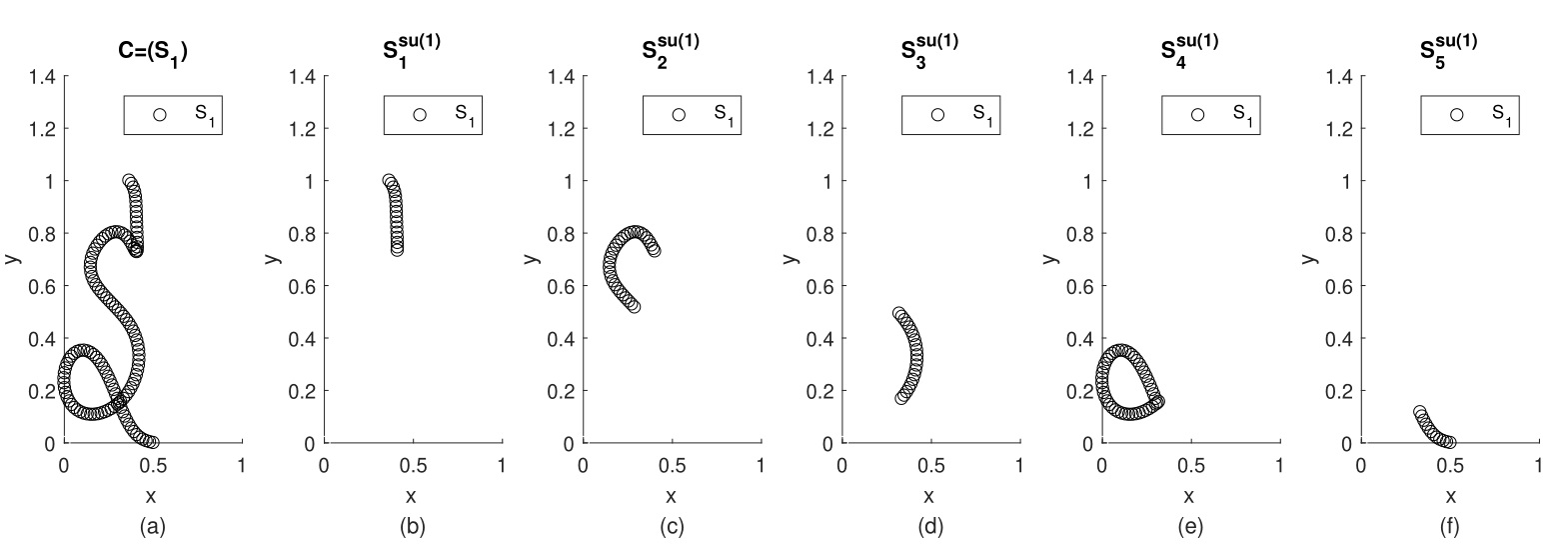}
\end{array}$
\caption{Illustration of extraction of sub-units from a handwritten character. (a) The handwritten character corresponding to the ideal online character in Fig. 20(a). This handwritten character is produced using a single stroke $S_1$. The stroke $S_1$ has five sub-units. (b)-(e) These are the sub-units extracted from the character in (a). These sub-units are similar to the corresponding sub-units in the ideal character shown in Fig. 20.}
\end{center}
\end{figure}
Figure 36 shows the same ideal character shown in Figure 20 but written differently. The first sub-unit extracted from this character is not a sub-unit structure of the corresponding ideal character. This first sub-unit region, as shown in Figure 36(b), has been extracted because of the absence of $rldc$ or pseudo sub-unit region required to split the region into two sub-unit regions.   
\begin{figure}[ht!]
\begin{center}
$\begin{array}{c}
\includegraphics[width=1\textwidth]{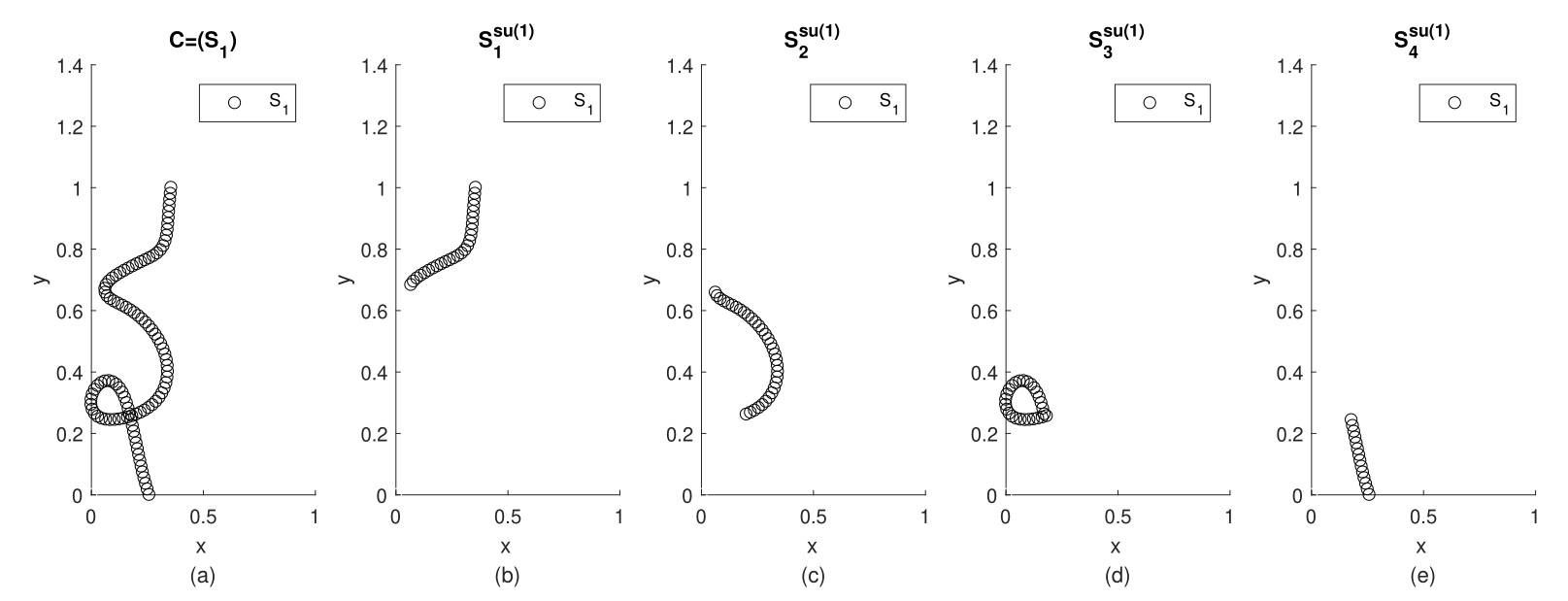}
\end{array}$
\caption{Illustration of extraction of sub-units from a handwritten character. (a) The handwritten character corresponding to the ideal online character in Fig. 20(a). This handwritten character is produced using the single stroke $S_1$. The stroke $S_1$ has four sub-units. (b) The first sub-unit $S_1^{su(1)}$ is not similar to any sub-unit in the corresponding ideal character. (c)-(e) These sub-units are similar to the third, fourth, and fifth sub-units in the corresponding ideal character.}
\end{center}
\end{figure}

\indent It is observed that the extracted sub-units from the handwritten character have the sub-unit structure of the corresponding ideal character most of the times. Extracted sub-units of the handwritten character that do not have the sub-unit structure of the corresponding ideal character are the result of variations in handwriting that produced the character.        

\subsection{Performance comparison of sub-unit based classifier with the other classifiers}
\indent To determine the importance of local representation of characters in terms of sub-units, Sharma \cite{hpod} develops a character classifier by modeling a character in terms of joint distribution of features extracted from sub-units in the character and those extracted from the same character as a whole. The performance of this classifier is compared with that of the other classifiers.\\
\indent Performance of a classifier is based on its classification accuracy on the samples of testing set, which the classifier has not been trained on. Classification performances of the Second Order Statistics (SOS), Subspace (SS), Fisher Discriminant (FD), Feedforward Neural Network (FNN), and Support Vector Machines (SVM)  classifiers trained only with global representations of characters and Sub-unit based (SUB) classifier trained with sub-unit level local and character level global representations of characters are evaluated on the testing dataset. The character recognition accuracies of different classifiers are given in table 1. The sub-unit based classifier has the highest classification accuracy of 93.5\% compared to that of the other classifiers, thus indicating the importance of local representations of characters in terms of sub-units for character recognition. 
\begin{table}[ht!]
\caption{Classification accuracies(\%) of SOS, SS, FD, FNN, SVM and SUB classifiers on the testing dataset.}
\begin{center}\begin{tabular}[ht!]{|c|c|c|c|c|c|c|c|}
\hline
Classifier&SOS&SS&FD&FNN&SVM&SUB\\
\hline
Accuracy&83.13&76.78&90.04&88.09&92.91&93.5\\
\hline
\end{tabular}\end{center}\end{table}

\section{Conclusion}
\indent Hindi online handwritten characters have been analysed in terms of local homogeneous sub-structures called the sub-units. Ideal Hindi online character can be uniquely represented in terms of sequence of these sub-units. A method for extraction of sub-units from handwritten characters has been developed. The extracted sub-units from handwritten characters have structures similar to that of the sub-units extracted from the corresponding ideal characters. 
\indent Character classifier model developed in Sharma \cite{hpod} is used to model a handwritten character locally in terms of features extracted from sub-units and globally in terms of features extracted from the character as a whole. This classifier has the highest classification accuracy among the classifiers using only the global character representation. This observation indicates the importance of local representation of characters in terms of the sub-units for building a high accuracy character classifer. 


\end{document}